\documentclass[11pt]{article}

\usepackage[final]{acl}

\usepackage{times}
\usepackage{latexsym}

\usepackage{enumitem}
\setlist[itemize]{leftmargin=*,topsep=0ex,itemsep=0.0cm,parsep=0cm}

\usepackage[T1]{fontenc}

\usepackage[utf8]{inputenc}

\usepackage{microtype}

\usepackage{inconsolata}

\usepackage{graphicx} 
\graphicspath{{plots/}}

\usepackage{amsmath}
\usepackage{amsfonts}
\usepackage{amssymb}
\usepackage{amsbsy}
\usepackage{mathtools}
\usepackage{multirow}

\usepackage{algorithm}
\usepackage{algorithmic}
\usepackage{booktabs}
\usepackage{subfig}
\usepackage{verbatim}

\usepackage{listings}
\DeclareCaptionStyle{ruled}{labelfont=normalfont,labelsep=colon,strut=off} 
\floatstyle{ruled}
\newfloat{listing}{tb}{lst}{}
\floatname{listing}{Listing}
\usepackage{cleveref}

\usepackage{xcolor}

\newcommand{\mj}[1]{\textcolor{green}{#1}}

\newcommand{\prompt}[1]{\begin{center}\fcolorbox{gray}{white}{\parbox{.95\linewidth}{\tt\scriptsize #1}}\end{center}}

\title{On the Brittleness of LLMs: A Journey around Set Membership}

\author{
Lea Hergert\textsuperscript{1} \qquad Gábor Berend\textsuperscript{1} \AND Mario Szegedy\textsuperscript{ 2} \qquad György Turán\textsuperscript{ 3,4} \qquad Márk Jelasity\textsuperscript{ 1,4}  \\\\
\textsuperscript{1 } University of Szeged, Hungary \\ \textsuperscript{2 }Rutgers University, USA \\ \textsuperscript{3 }University of Illinois at Chicago, USA \\ \textsuperscript{4 }HUN-REN--SZTE Research Group on AI, Hungary \\\\
}

\begin{document}

\maketitle

\begin{abstract}
Large language models (LLMs) achieve superhuman performance on complex reasoning tasks, yet often fail on much simpler problems, raising concerns about their reliability and interpretability. We investigate this paradox through a focused study with two key design features: simplicity, to expose basic failure modes, and scale, to enable comprehensive controlled experiments.
We focus on set membership queries---among the most fundamental forms of reasoning---using tasks like ``Is apple an element of the set \{pear, plum, apple, raspberry\}?''.
We conduct a systematic empirical evaluation across prompt phrasing, semantic structure, element ordering, and model choice. Our large-scale analysis reveals that LLM performance on this elementary task is consistently brittle, and unpredictable across all dimensions, suggesting that the models' ``understanding'' of the set concept is fragmented and
convoluted at best.
Our work demonstrates that the large-scale experiments enabled by
the simplicity of the problem allow us
to map and analyze the failure modes comprehensively, making this approach a valuable methodology
for LLM evaluation in general.

\end{abstract}

\section{Introduction}

While LLMs show impressive performance, it is important to understand their various deficiencies as well
\cite{coy24,cuskley2024limitations,pass24,shojaee2025illusion,Gyor25}.

One such potential deficiency is \emph{brittleness} that, interestingly, has often been cited as a
main criticism of \emph{classical AI} (e.g., expert systems) that is unable to handle the flexibility of human intelligence and deal with the imperfect (noisy, incomplete) nature of knowledge and reasoning ~\cite{mit21}.
Recently, \emph{LLMs have been shown to be quite brittle as well}, but from different aspects. There is a growing literature on the failure of LLMs to handle simple tasks such as
common sense inference, counting, and so on
\cite{Huck25, Zhou24,mit25}.

A particularly interesting source of such brittleness is semantics interfering with reasoning.
Human performance on simple reasoning tasks is known to be influenced by semantics, and this has been observed for LLMs as well on simple natural language inferences, syllogisms, and the Wason selection task~\cite{wason1968reasoning, Dasgu22, lampinen2024language}.
A related phenomenon called semantic leakage has also been described for instruction-tuned models~\cite{Gonen25}.

To study brittleness, we focus on a task where it is reasonable to expect perfect and robust performance: set membership in small,
explicitly listed sets.
The task is well-motivated also because the set is, perhaps, the most fundamental concept
in human thinking, both for commonsense reasoning and mathematics.
The questions we will ask will be as simple as \emph{``Is apple an element of the set \{pear, plum, apple, raspberry\}?''}.
This task is of utmost simplicity. As far as we know, it is simpler than the basic reasoning tasks studied in the related literature. Therefore, any mistakes are interesting and could reveal relevant insights into fundamental LLM behavior and failure modes.

The simplicity of the task makes it possible to perform a systematic, thorough experimental analysis. 
We explore a large space of prompts (including natural language and Python coding) using a diverse collection of sets with and without
semantic relationships.
Overall, more than 60 million set membership queries are tested using 7 instruction-tuned LLMs.

\textbf{Our contributions} are the following.
\begin{itemize}
\item We reduce the task presented to the LLM to great simplicity (set containment) to
enable a large-scale, holistic investigation.
\item We design a systematic set of experiments aimed at multiple possible sources of
brittleness and study them in conjunction; this could serve as a benchmark for
future evaluations as well.
\item We map a diverse set of error patterns, including
\emph{high sensitivity to element ordering, minor prompt
variations, and semantic relations}, none of which should play a role in our task.
\item We show that the semantic relatedness of set elements can be both an
advantage (semantic boosting) and a disadvantage (semantic leakage) in certain cases.
\item We demonstrate that different LLMs exhibit different, and often wildly dissimilar error patterns.
\end{itemize}

\section{Related Work}

In this section we give a brief overview of some of the work most relevant to our paper, from the vast related literature.

\textbf{Prompt sensitivity} is studied in~\cite{sclar2024quantifying} by formulating a grammar for prompt formatting and evaluating samples of prompt formats over a wide range of tasks. Another framework is presented in~\cite{Zhuo24}, evaluating prompts on the instance level. Our approach expands on these works in that we focus on a single basic problem, set membership,
and study prompt sensitivity along with many other dimensions relevant to the general brittleness of LLMs.

Research into the \textbf{consistency of world models} is also relevant, as our results also suggest that LLMs have \emph{no consistent models of small sets}.
\cite{wolfram2025world} consider city populations and isotope half-lives, and find that responses are relatively consistent over different prompts, but often inconsistent on related pieces of information. Other papers in the same direction are \cite{elaz21,fier22,sahu22,raj23,zheng24rel}.
Invariance can be another required property~\cite{Bent20}, with  permutation invariance being a prominent example~\cite{egr25,Rava17,Zahe17}. In our case, responses to a set membership query should be invariant under permutations of the elements in the set, but we find that they are not.

In \textbf{cognitive science}, the mental models theory of \cite{JohnL} is used in \cite{Khem14} to form a theory of human reasoning about set membership for a class of syllogisms, and present some experiments confirming the predictions of the theory. In psychology, the recognition problem\footnote{``In its simplest form, sometimes referred to as “item recognition”, subjects are first asked to study a list of items such as words, objects, or images. Then, after a delay, they are presented with a mixture of studied and nonstudied (i.e., new) items and are required to use their memory to indicate whether each item had been in the studied list or is new.''~\cite{Yonel}.} \cite{Yonel} is studied using techniques like response times~\cite{Dew06} and neuroimaging~\cite{Wais10}.
Students' acquisition of the set concept has been studied in~\cite{Razm}.




\section{Experimental Setup}
\label{sec:setup}

We study very basic set membership queries
to reveal and understand the counter-intuitive mistakes that otherwise capable LLMs make over such a simple
task.
We use simple prompts based on natural language or Python code snippets, such as the two
examples below:

\prompt{Does the set \{"Hearts", "Diamonds", "Clubs", "Spades"\} contain the element "North"?}

\prompt{Given the Python code below, what will be the value of the `result` variable?\\
\\
```python\\
S = \{"North", "South", "West", "East"\}\\\
result = "North" in S\\\
```}

Our experimental setup is based on a thorough, well-informed construction of a fixed collection of 22 sets and
a large set of prompts that we generate from templates, as described later on.

\subsection{Sets}

We decided to work with sets of exactly four elements.
Our preliminary experiments revealed that increasing the size of the set increases
the error rates.
A size of four is still very small but already allows for observing interesting patterns.

We defined 22 sets of 5 types: complete, related word, related number, unrelated word, and unrelated number.
The list of our 22 sets is given in the Supplementary material. 
Here, we give the motivation for each type.

\textbf{Complete word sets} consist of four words that exhaustively cover a category such as \{Hearts, Diamonds, Clubs, Spades\}.
We identified 10 such sets using preliminary experiments, where we tested candidate sets
for completion: we removed one element and asked a set of LLMs what the missing element is.
The selected 10 sets were completed by each LLM we tested without error, for every possible
missing element, and every ordering of the set.

\textbf{Related sets} contain words or numbers that are similar but do not form
a complete set (e.g. {\em some} set of four fruits, or {\em some} numbers divisible by 100, etc.).
We included 3 such word sets and 3 number sets.

\textbf{Unrelated sets} contain words or numbers that have no obvious commonality or relation.
We included 3 such word sets and 3 number sets.
It is not evident that there is indeed no connection among a given
set of numbers or words,
so we tested a number of LLMs asking for commonalities within our sets of unrelated elements, and
no meaningful connections were suggested.

\subsection{Queries}

\textbf{Query types.} Let us assume that we are given a set $S=\{x_1, x_2, x_3, x_4\}$.
We define the following query types.

\begin{itemize}
\item \emph{Positive:} We select an element $x_i\in S$ and ask whether $x_i\in S$. We
do this for all $i\in \{1,2,3,4\}$.
\item \emph{Negative-member:} We select an element $x_i\in S$ and ask whether $x_i\in S\setminus\{x_i\}$. We
do this for all $i\in \{1,2,3,4\}$.
\item \emph{Negative-intruder:} We select $y\not\in S$ and ask whether $y\in S$.
This query type is implemented only for related and complete sets, and  intruder $y$ is selected from a
different set of the same type (word or number) at random.
\end{itemize}

\textbf{Permutation.} Clearly, the order of the presentation of the elements of a set \emph{should not have any effect on the answer
of the LLM}, if the LLM correctly understands the concept of a set.
Yet, the elements still have to be presented in a fixed sequential order in every prompt.
To examine the effect of this ordering, we test the queries over all the possible permutations
of the set.

\textbf{Query instances.} 
A query is given by $[(z_{1},\ldots,z_{n}),z]$, where $n=3$ or $n=4$ depending
on the query type, the list $(z_{1},\ldots,z_{n})$ contains the elements of the set in the
order they are to be listed, and $z$ is the element we
query for set membership.

\subsection{Prompts}
\label{sec:prompts}

\textbf{Templates.} A prompt is created by the instantiation of a \emph{prompt template} for a given query.
For example, a prompt template could be {\tt\small 'Does the set \{ "$z_1$", "$z_2$", "$z_3$", "$z_4$" \} contain the element "$z$"?'}
Applying this template to the query instance $[$ (Hearts, Diamonds, Clubs, Spades), North $]$ results in the first prompt we
presented earlier.

\textbf{Template classes.}
We define four prompt template classes: two natural language classes (NL1 and NL2) and two Python code classes
called CS (coding simple) and CA (coding algorithmic).
Each class is defined by an abstract template in which a number of features can be selected from
a list of options.
Then, all the feature combinations are used to generate prompt instances for a given query.

The \textbf{NL1 template class} consists of two schemes:

\prompt{Does the \textbf{set|list} \textit{<set>} \textbf{contain|include element|item|character sequence|string} \textit{<element>}?}
\prompt{Is the \textbf{element|item|character sequence|string}   \textit{<element>} \textbf{included|contained} in the \textbf{set|list}  \textit{<set>}?}

The features that are defined for the NL1 class are the following:

\begin{itemize}
\item \emph{arrangement}: set-first or element-first scheme is used
\item \emph{set}: set $|$ list
\item \emph{element}: element $|$ item $|$ character sequence $|$ string
\item \emph{contain}: include $|$ contain (set-first scheme); or included $|$ contained (element-first scheme)
\item \emph{quotation}: set elements are quoted as Hearts, 'Hearts', or "Hearts", using identical settings in the set definition
\texttt{\textit{<set>}} and the element definition \texttt{\textit{<element>}}
\item \emph{listing}: defines how the set is presented. It has 5 options. The first three options
are comma-separated lists enclosed in either \{\}, (), or $[]$.
The remaining two options are bullet point listings with bullet types '-' or '*'.
\end{itemize}

The \textbf{NL2 template class} also consists of two schemes:

\prompt{A \textbf{set|list} consists of the following \textbf{elements|items|character sequences|strings}:
\textit{<set>} Does this \textbf{set|list} \textbf{include|contain} the \textbf{element|item|character sequence|string} \textit{<element>}?}

\prompt{Does a \textbf{set|list} \textbf{include|contain} the \textbf{element|item|character sequence|string} \textit{<element>}
if it consists of the following \textbf{elements|items|character sequences|strings}:
\textit{<set>?}}

The features of the NL2 class are identical to those of the NL1 class, noting
that the \emph{set} and \emph{element} features assign identical values to each occurrence of the given feature
within the prompt.

The \textbf{CS and CA Python template classes} contain more variation, the
full description is given in the supplementary material.
The main difference between the two classes is that in the case of CS the
set membership is decided by Python's \texttt{in} operator, while in the case
of CA a \texttt{for} loop is used to iterate through the data structure
to determine membership.

For illustration, an example CS template is

\prompt{Given the Python code below, what will be the value of the `result` variable?\\
\\
```python\\
my\_set = \textit{<set>}\\\
result = \textit{<element>} in my\_set\\\
```}

Replacing the computation of the \texttt{result} variable with

\prompt{
result = False\\
for set\_item in my\_set:\\
\phantom{xxxx}if set\_item == \textit{<element>}:\\
\phantom{xxxxxxxx}result = True}

\noindent results in a CA template.
To generate such templates, similarly to the NL schemes, we define Python schemes
with features such as whether we have a
function named \texttt{contains} to test for set membership, whether we have a docstring documenting the
Python code, as well as several initialization methods for the data structure
containing the set, and so on.
Both the CS and CA template class contains 960 templates.

\textbf{Notes on prompt engineering.}
We simply defined a large and diverse set of prompts that we considered unambiguous and intuitive, without
using any prompt engineering techniques explicitly.
The reason is that our goal was not to achieve perfect accuracy, but to investigate the error patterns of LLMs under extremely simple and clear
task definitions.
Still, some of our prompt schemes 
produced perfect accuracy (see \cref{sec:llmsensitivity}) justifying our
prompting approach.

\subsection{Practical Notes}
\label{sec:setupnotes}

The \textbf{number of prompt instances} is given by the number of queries multiplied by
the number of prompt schemes.
The number of queries is given by
\begin{equation*}
\begin{split}
3024 = \overbrace{16\cdot(\color{red}{4!\cdot 4}+\color{blue}{3!\cdot 4}+\color{cyan}{3!\cdot 4})}^{\text{semantically related queries}}+\\\underbrace{6\cdot(\color{red}{4!\cdot 4}+\color{blue}{3!\cdot 4})}_{\text{semantically unrelated queries}},
\end{split}
\end{equation*}
where we have 16 semantically related and 6 unrelated sets, 
we considered \textcolor{red}{positive}, \textcolor{blue}{negative-intruder}, and \textcolor{cyan}{negative-member} types, and $x!$ is the number of permutations
of $x$ elements.
The number of prompt instances is thus
\begin{equation*}
8,709,120= 3024 \cdot (480\cdot 2+960\cdot 2),
\end{equation*}
where 480 is the number of prompt schemes in the NL1 and NL2 classes and we have 960 in the CS and CA classes.

\textbf{Evaluation.} We insert the prefix {\tt `Answer with a single word.'} in front of every prompt.
We then search the LLM's answer for negative (false, no) and positive (true, yes) words.
If both or none of these answers are found, then the answer is undecided; these
represent about 0.14\% of the answers and were treated as incorrect.
Almost every such undecided answer was found in the NL1 and NL2 prompt
categories, there were none in the CA category and 11 cases (out of more than 20 million) in the CS category.

\begin{figure*}
\centering

\setlength{\tabcolsep}{2pt}
\begin{tabular}{ccccccc}
\scriptsize Llama3.3-70B &
\scriptsize Llama3.1-70B & 
\scriptsize Phi-3.5-MoE &
\scriptsize Qwen2.5-32B &
\scriptsize Mistral-24B &
\scriptsize Llama-3.1-8B &
\scriptsize Llama-3.2-3B \\
\includegraphics[width=0.134\textwidth]{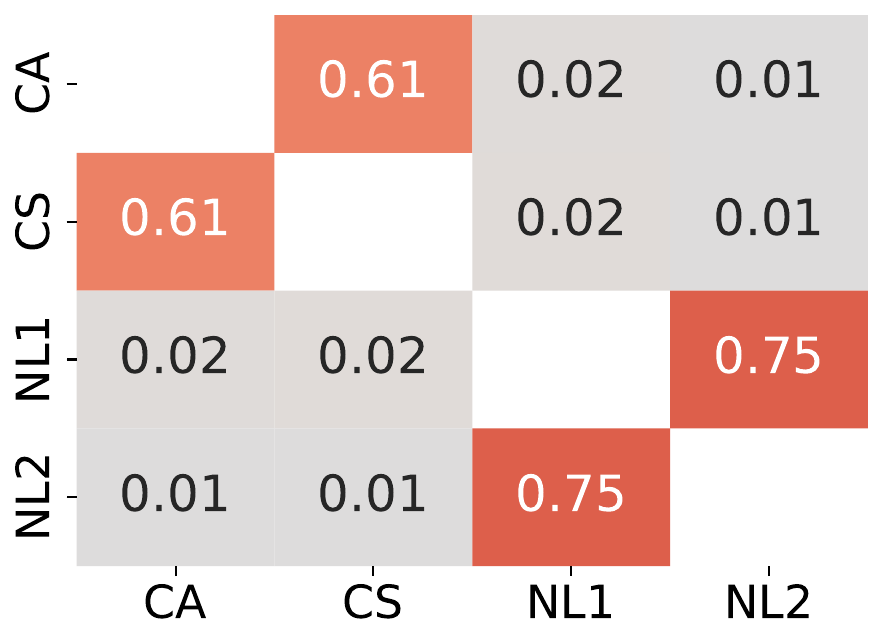} &
\includegraphics[width=0.134\textwidth]{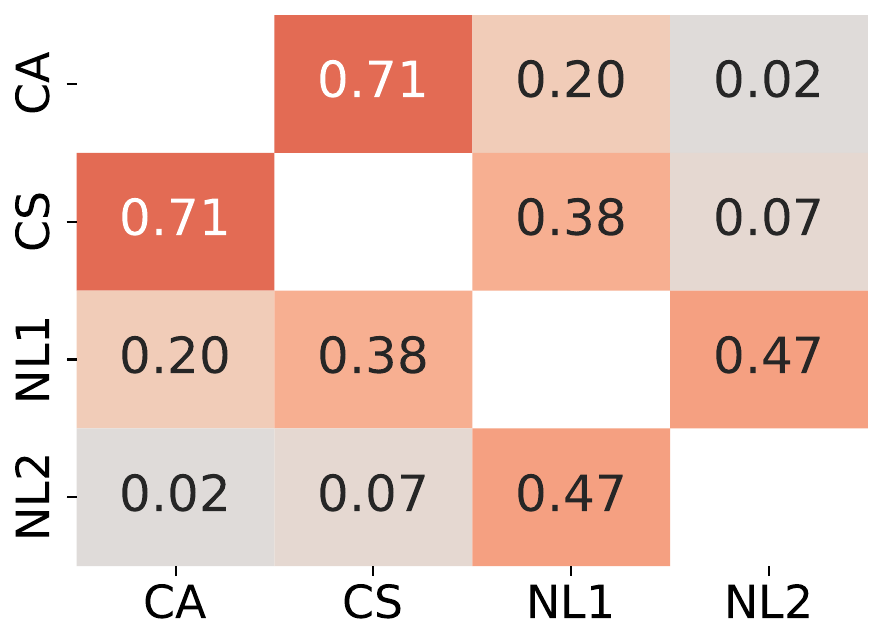} &
\includegraphics[width=0.134\textwidth]{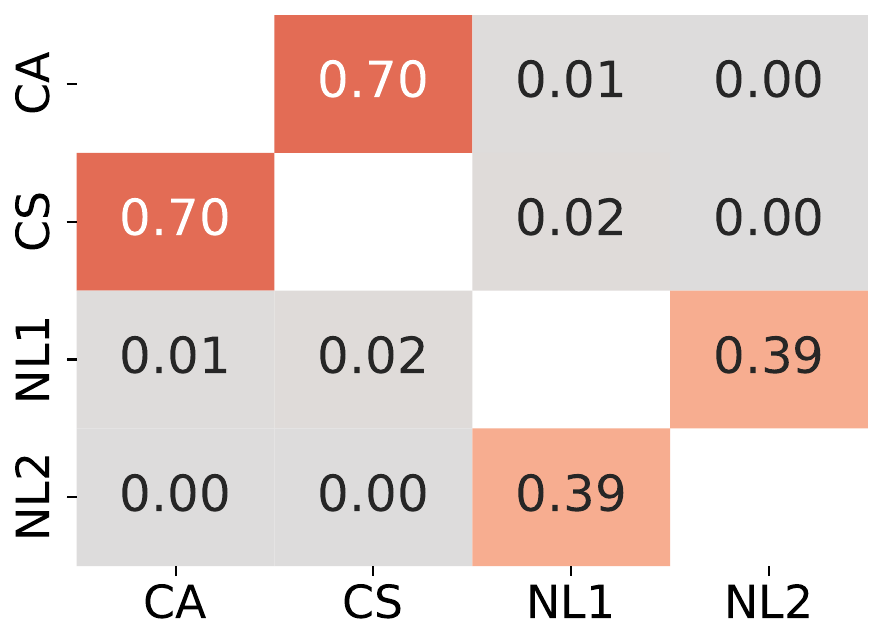} &
\includegraphics[width=0.134\textwidth]{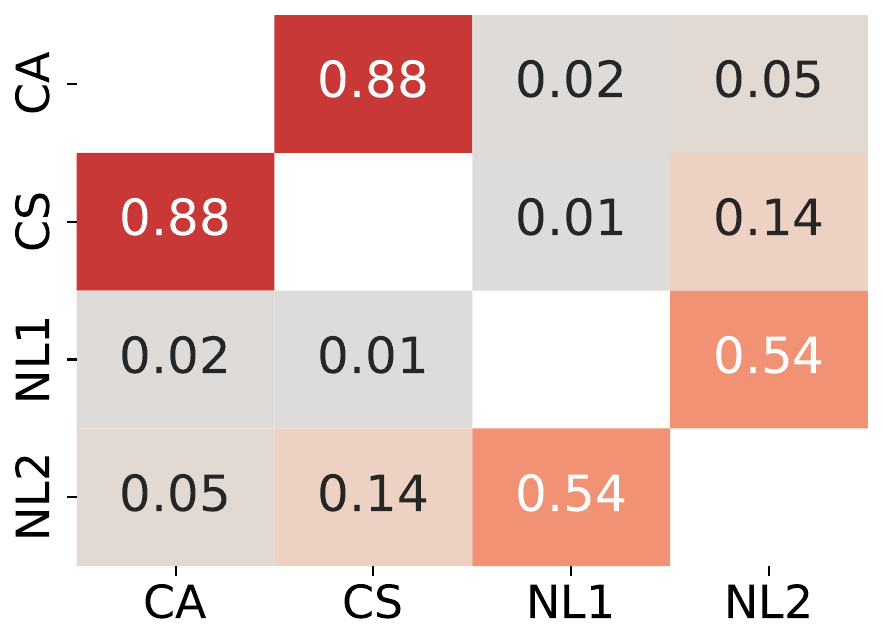} &
\includegraphics[width=0.134\textwidth]{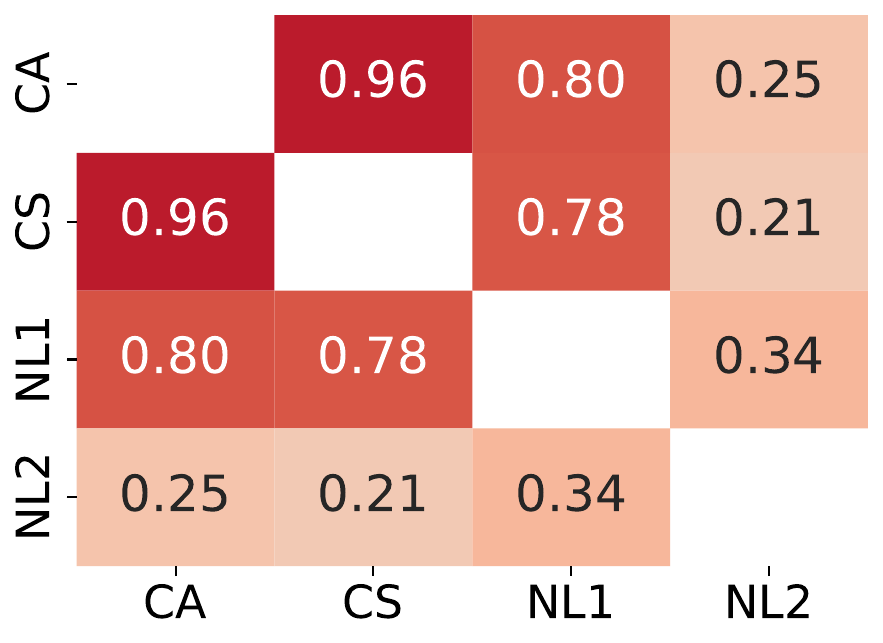} &
\includegraphics[width=0.134\textwidth]{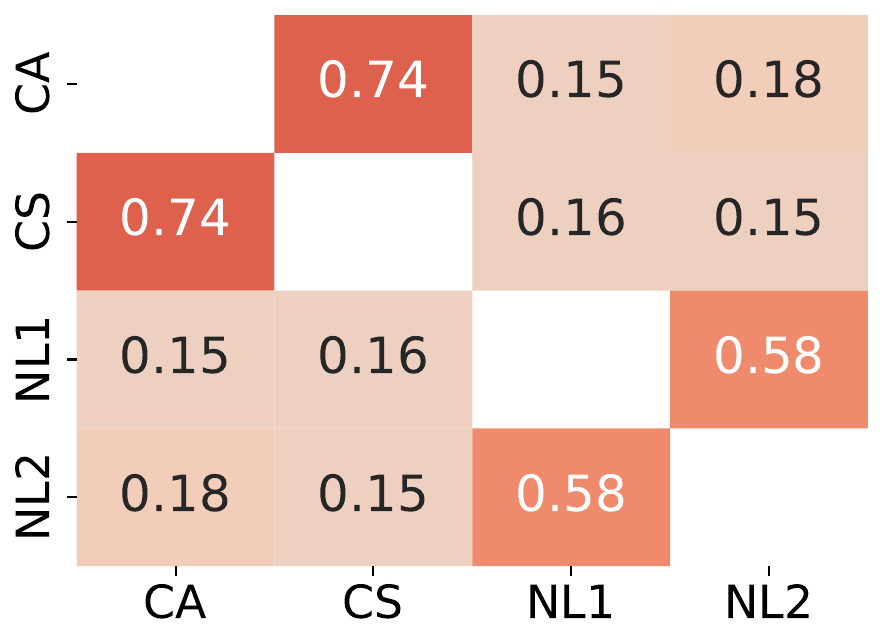} &
\includegraphics[width=0.134\textwidth]{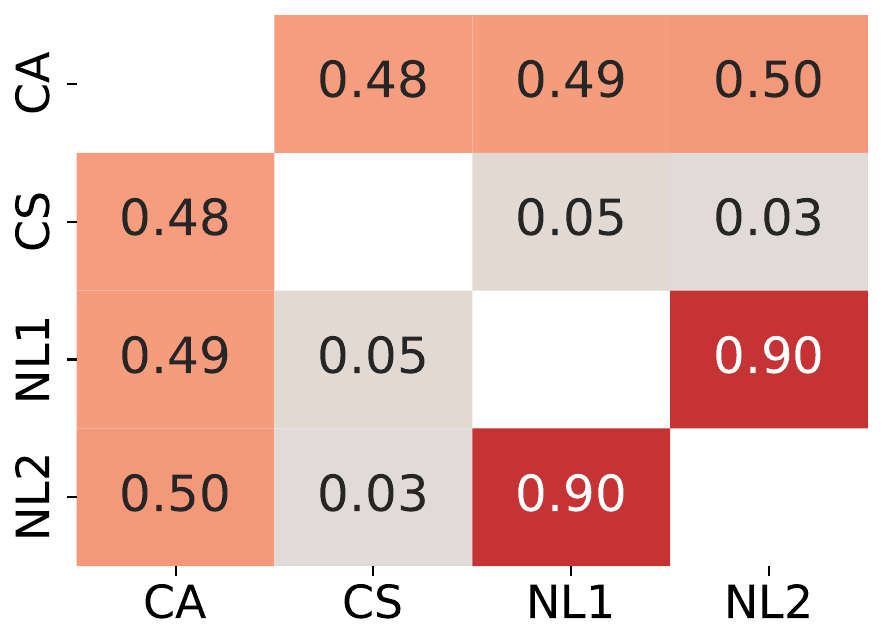} \\
\includegraphics[width=0.134\textwidth]{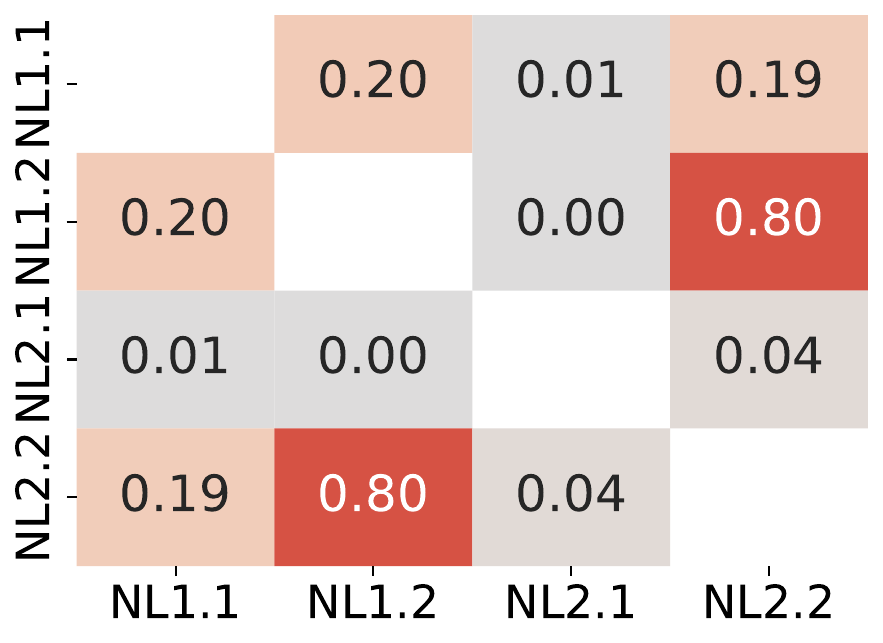} &
\includegraphics[width=0.134\textwidth]{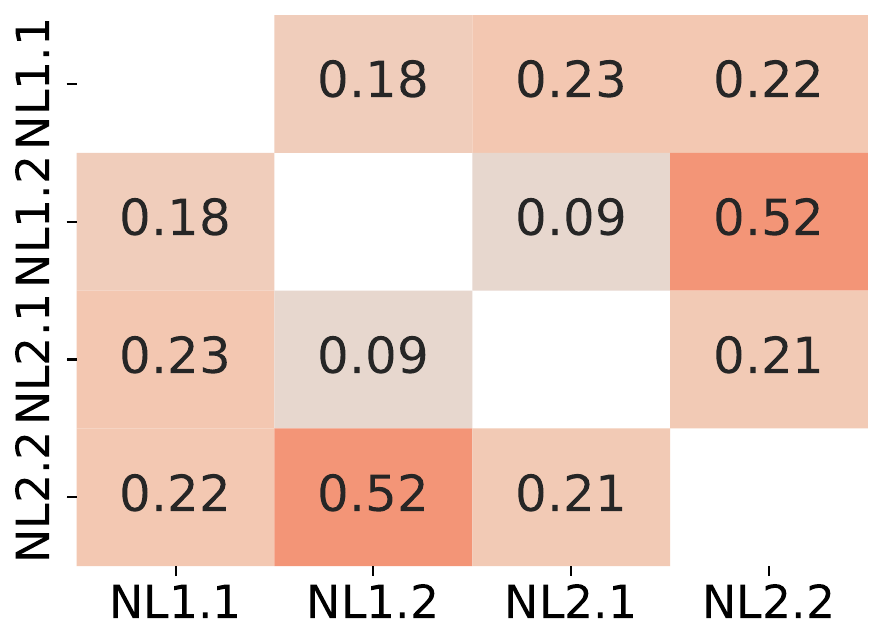} &
\includegraphics[width=0.134\textwidth]{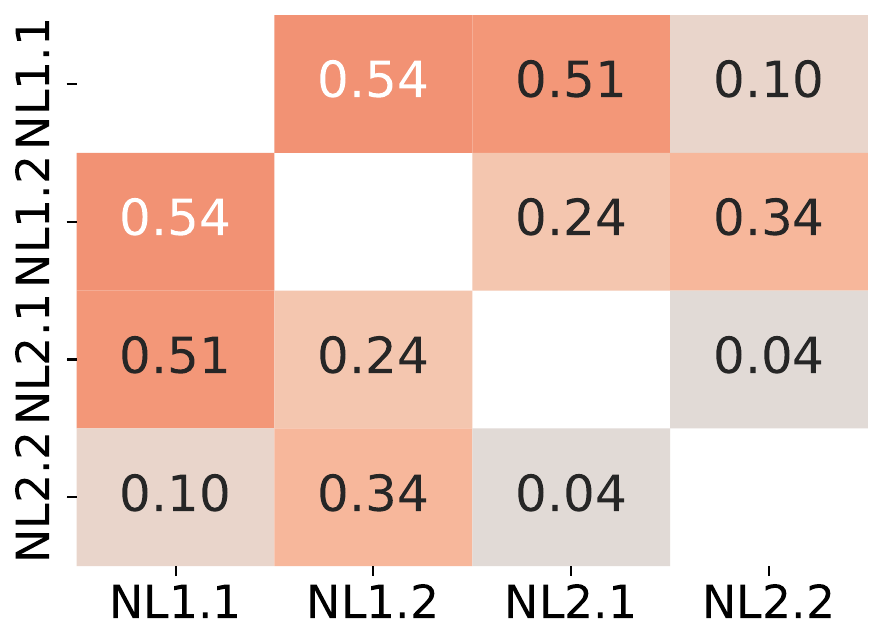} &
\includegraphics[width=0.134\textwidth]{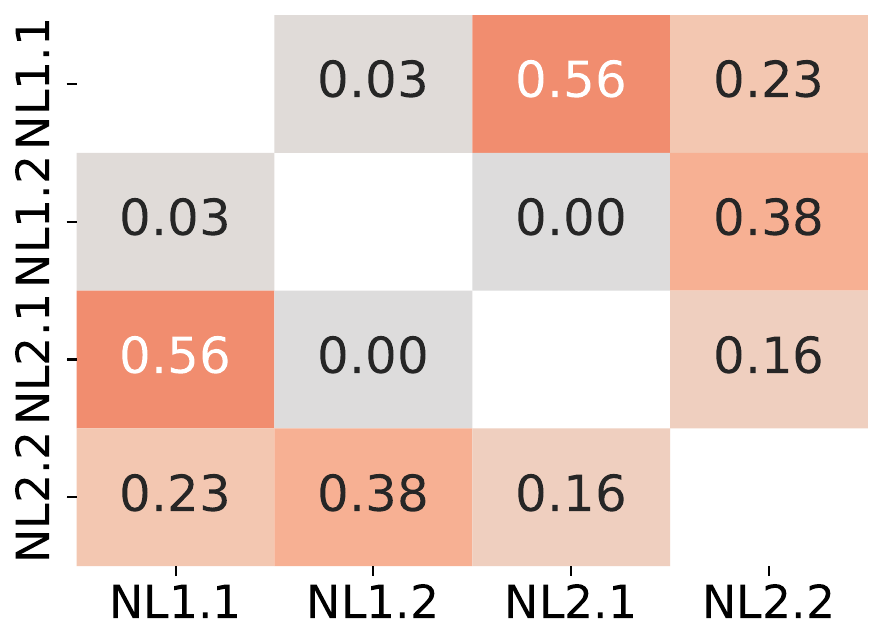} &
\includegraphics[width=0.134\textwidth]{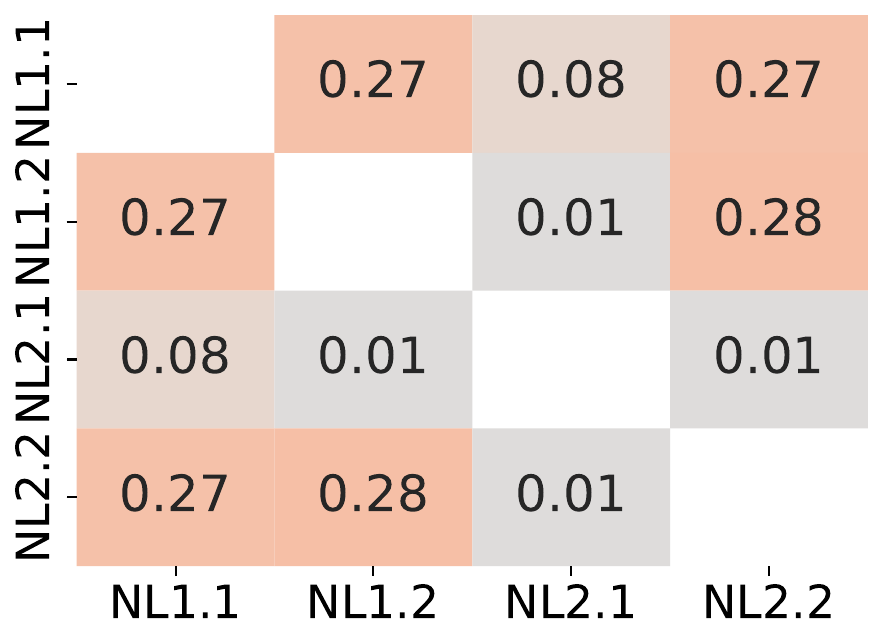} &
\includegraphics[width=0.134\textwidth]{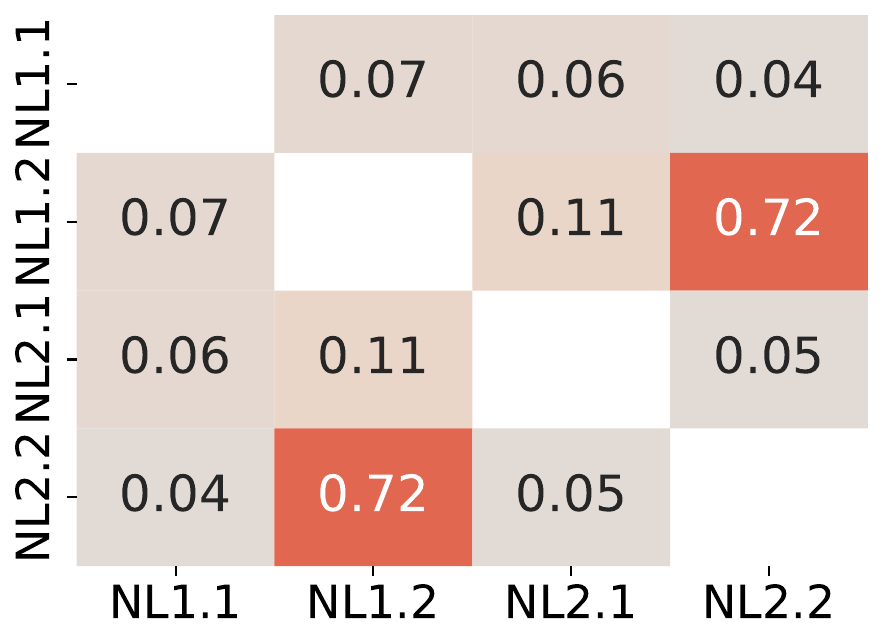} &
\includegraphics[width=0.134\textwidth]{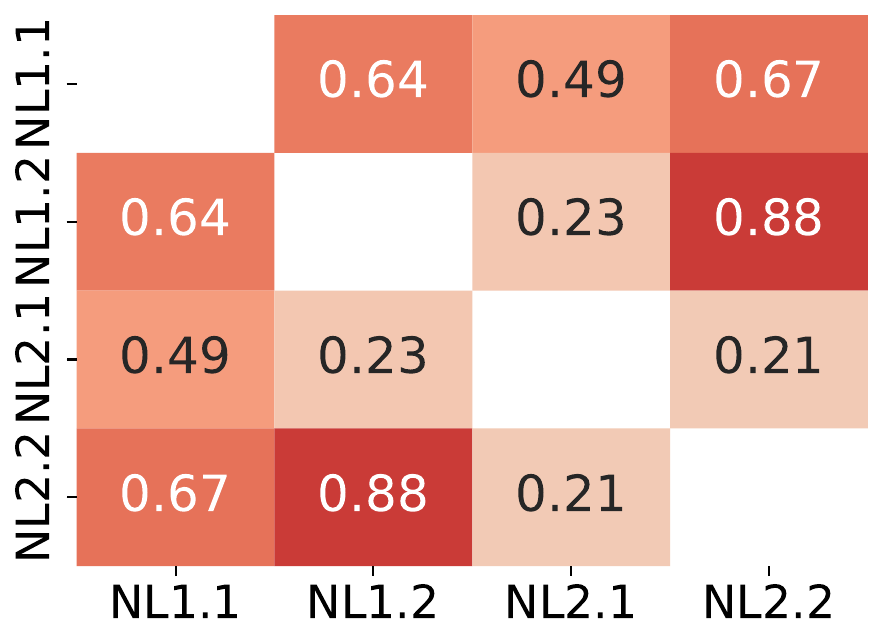} \\
\end{tabular}
\caption{Cosine similarity between the average query error-rates of prompt categories for all the LLMs (top row).
The bottom row shows the cosine similarity between sub-classes of NL1 and NL2, where the division is based on
the \emph{arrangement} feature (see \cref{sec:prompts}). NL1.1 and NL2.1 use the element-first, the others use the set-first arrangement.}
\label{fig:promptcos}
\end{figure*}

\begin{table}
\centering
\small
\begin{tabular}{lccc}
      &           & Member & Intruder \\
Model &  Accuracy &  FPR &    FPR \\ \midrule
Llama-3.2-3B  & 94.670\% & 5.716\% & 4.363\% \\
Llama-3.1-8B  & 99.470\% & 1.541\% & 0.594\% \\
Mistral-24B   & 98.512\% & 0.019\% & 0.000\% \\
Qwen2.5-32B   & 99.892\% & 0.134\% & 0.004\% \\
Phi-3.5-MoE   & 98.832\% & 0.587\% & 0.191\% \\
Llama-3.1-70B & 99.467\% & 0.030\% & 0.036\% \\
Llama-3.3-70B & 99.462\% & 0.073\% & 0.076\% \\
\midrule
Average & 98.615\% & 1.157\% & 0.752\% \\ 
\end{tabular}
\caption{Statistics 
of LLM performance.}
\label{table:llmstats}
\end{table}

\begin{figure}[t]
\centering
\setlength{\tabcolsep}{1pt}
\begin{tabular}{cc}
\rotatebox{90}{\scriptsize \phantom{XXXXXX}Llama3.3-70B, NL1} &
\includegraphics[width=.45\textwidth]{shap/shap-llama3370b-nl1} \\
\rotatebox{90}{\scriptsize \phantom{XXXXXX}Llama3.1-70B, NL1} &
\includegraphics[width=.45\textwidth]{shap/shap-llama3170b-nl1} \\
\rotatebox{90}{\scriptsize \phantom{XXXXXXX}Phi3.5-MoE, NL1} &
\includegraphics[width=.45\textwidth]{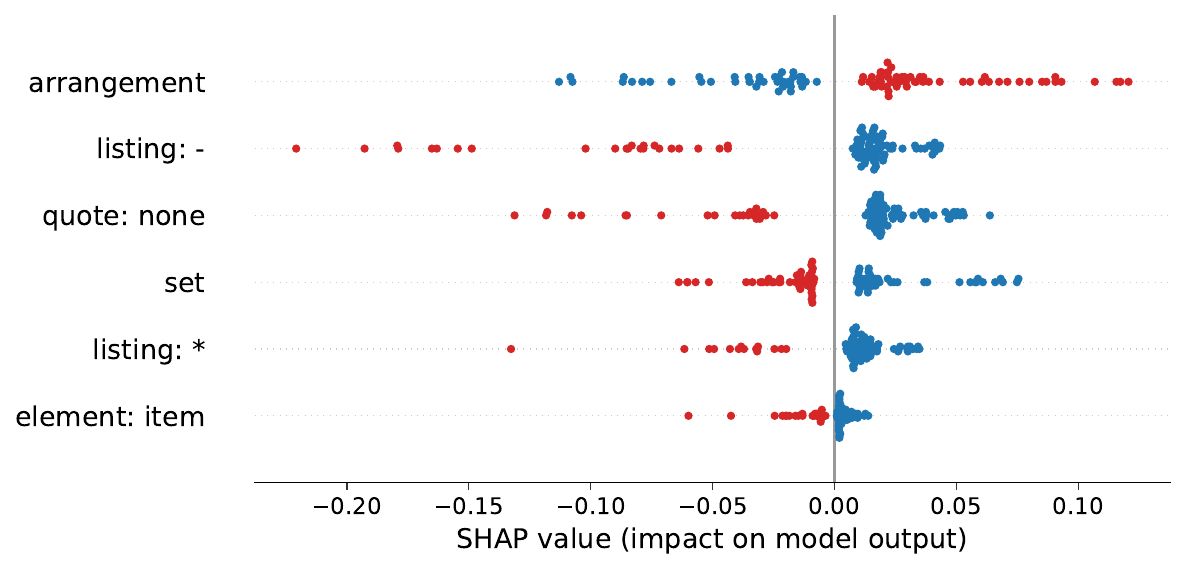}\\
\end{tabular}
\caption{Top 6 Shapley values among binary features of the NL1 prompt category.
The features are defined in \cref{sec:prompts}, non-binary features are converted to one-hot representation.
A dot corresponds to a prompt template and the color red indicates that the feature is present.}
\label{fig:shap}
\end{figure}

\begin{figure*}[tb]
\centering
\setlength{\tabcolsep}{2pt}
\begin{tabular}{@{}cccc@{}}
\includegraphics[width=0.26\linewidth]{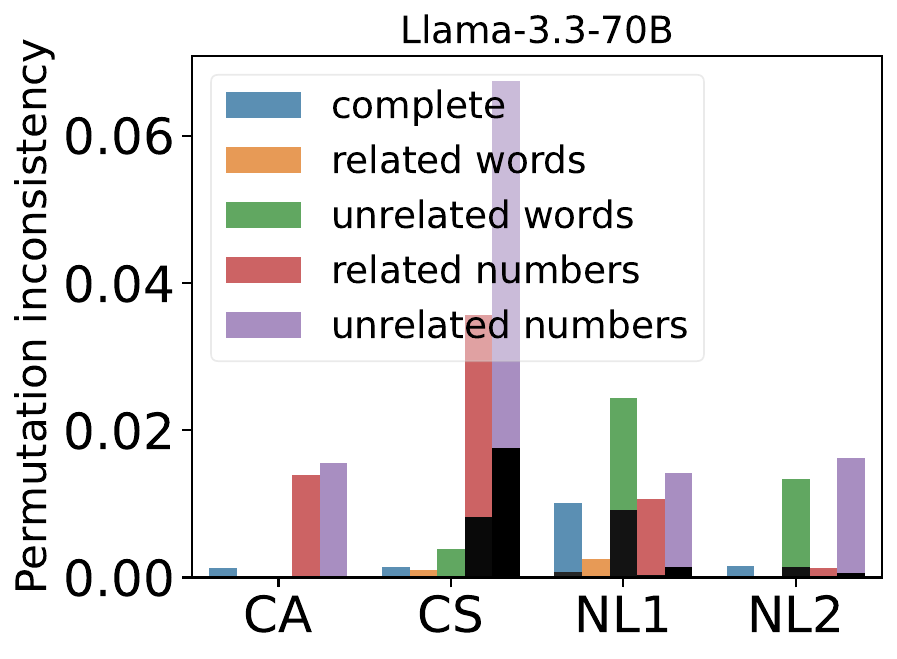} &
\includegraphics[trim={1cm 0 0 0},clip,width=0.24\linewidth]{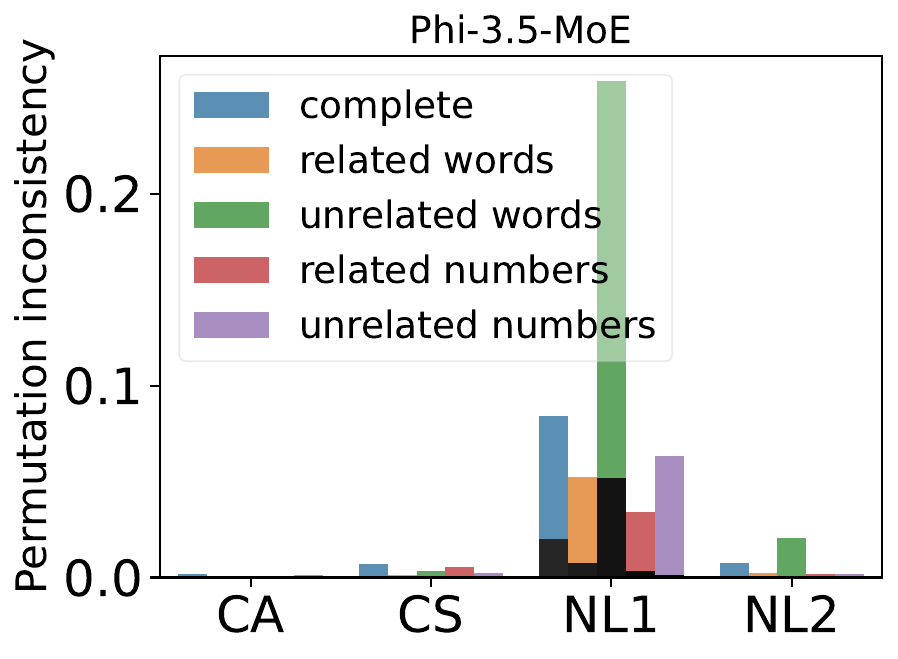} &
\includegraphics[trim={1cm 0 0 0},clip,width=0.24\linewidth]{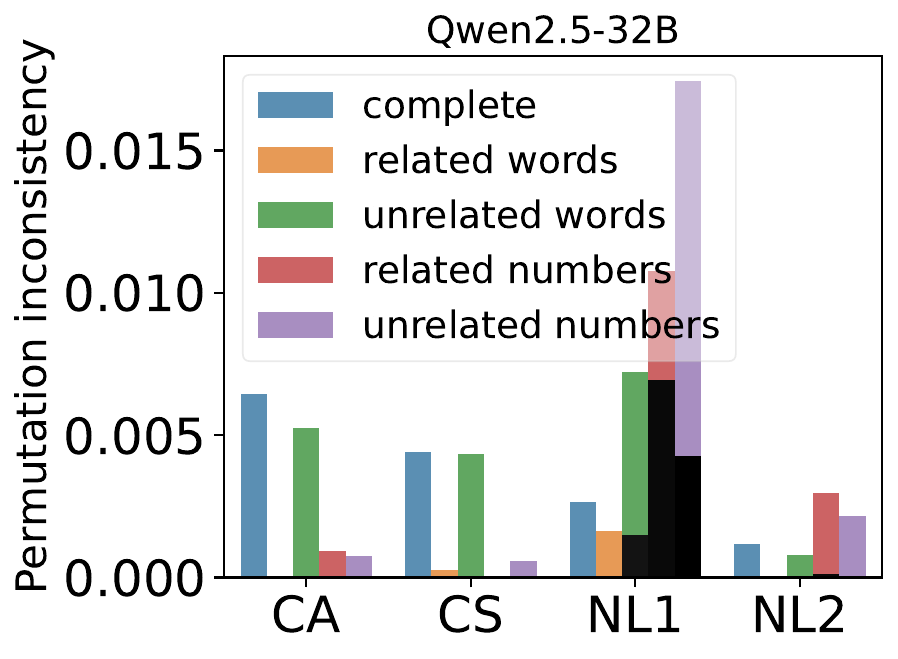} &
\includegraphics[trim={1cm 0 0 0},clip,width=0.24\linewidth]{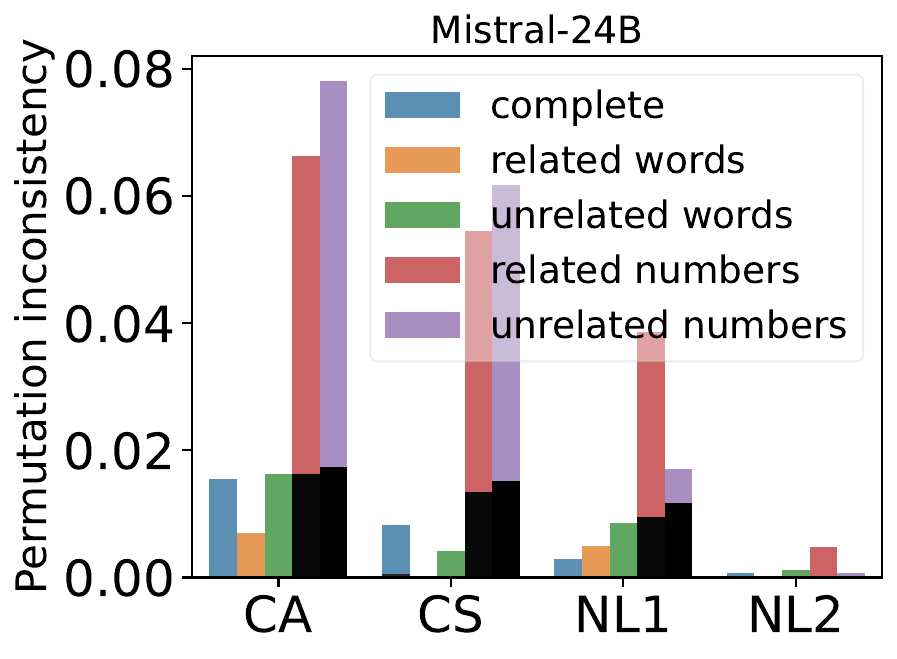} \\
\end{tabular}
\caption{The fraction of order-independent queries where the answer of the LLM is not consistently correct for all the permutations of the set, by
prompt type and set type. 
Values higher than 0.0 indicate the presence of sensitivity to ordering. The black region of a bar corresponds to the fraction of consistently incorrect order-independent queries.}
\label{fig:consistency}
\end{figure*}

\begin{figure*}[tb]
\centering
\includegraphics[width=\textwidth]{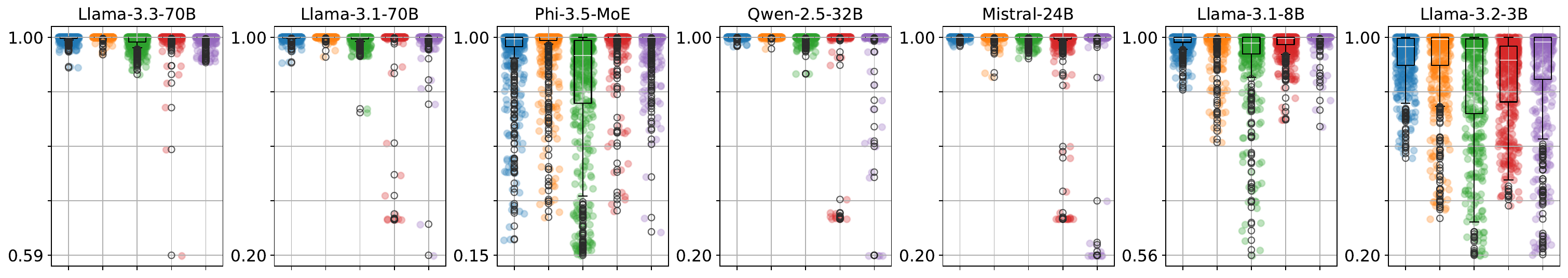}
\includegraphics[trim={0 0 0 0.8cm},clip,width=\textwidth]{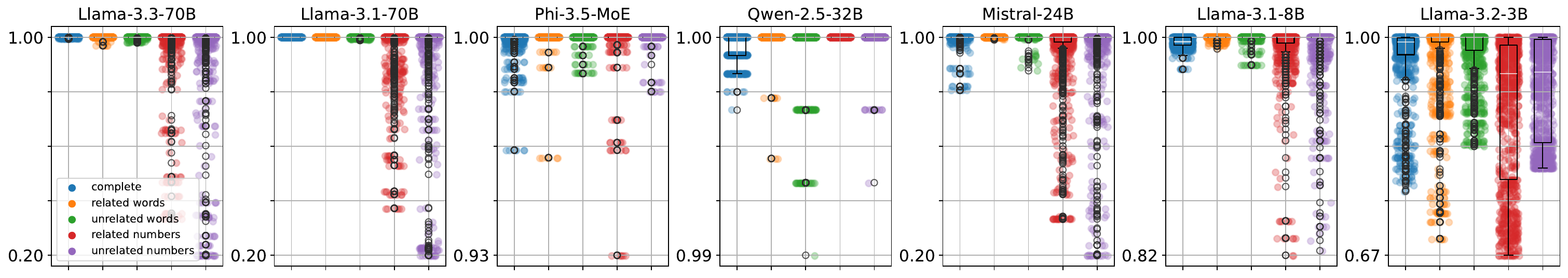}
\caption{The average accuracy of queries with different set-types, for all the LLMs, in the NL1 (top) and CS (bottom) prompt category.
Every point belongs to a fixed prompt template and represents the average of the queries that belong to the given set type.
For example, for the complete set type, every point is the average accuracy of $10\cdot (24\cdot 4+6\cdot 4+6\cdot 4)$ prompts.}
\label{fig:settype}
\end{figure*}

\begin{figure*}
\centering
\setlength{\tabcolsep}{-2pt}
\begin{tabular}{ccccccc}
\scriptsize Llama3.3-70B &
\scriptsize Llama3.1-70B & 
\scriptsize Phi-3.5-MoE &
\scriptsize Qwen2.5-32B &
\scriptsize Mistral-24B &
\scriptsize Llama-3.1-8B &
\scriptsize Llama-3.2-3B \\
\includegraphics[width=0.15\textwidth]{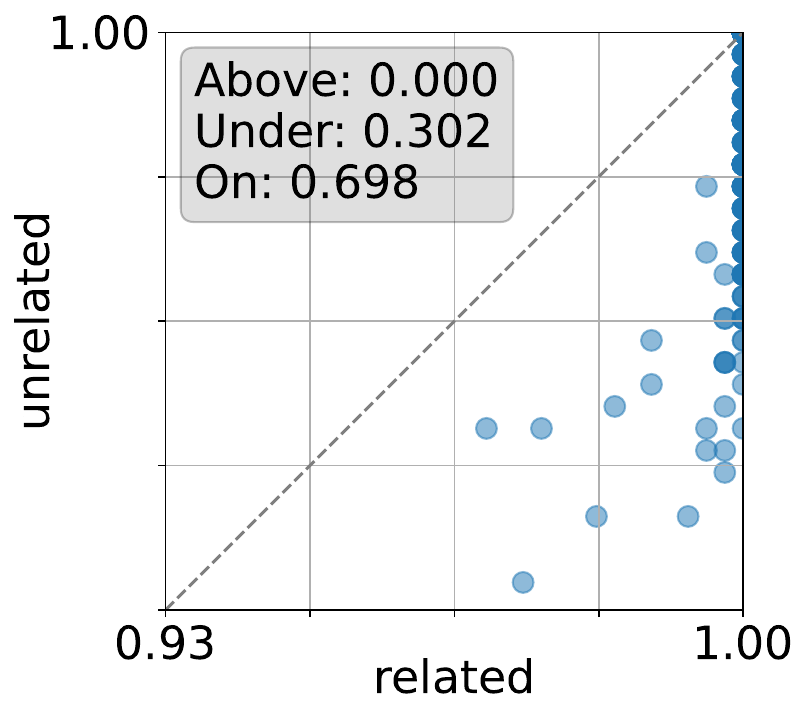} &
\includegraphics[width=0.15\textwidth]{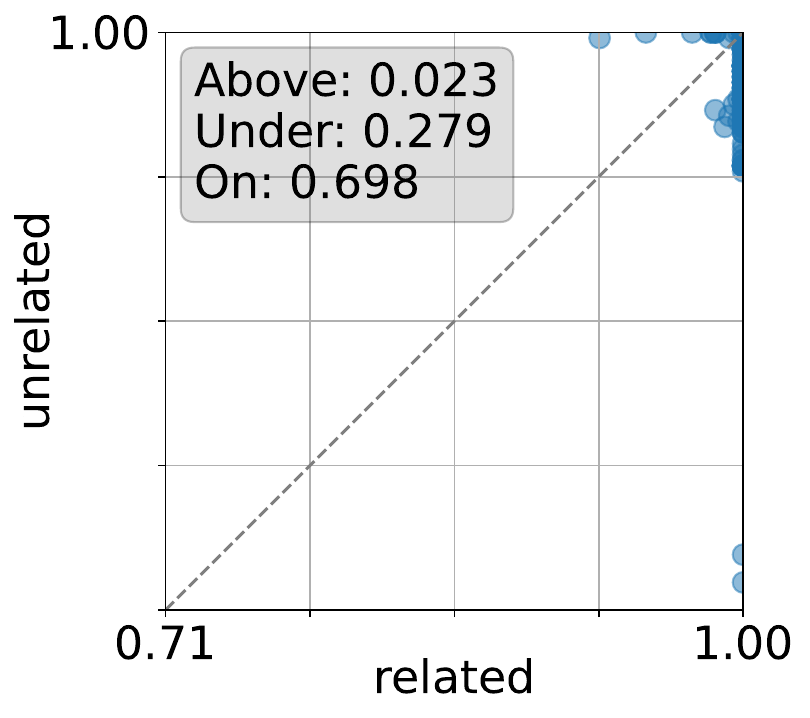} &
\includegraphics[width=0.15\textwidth]{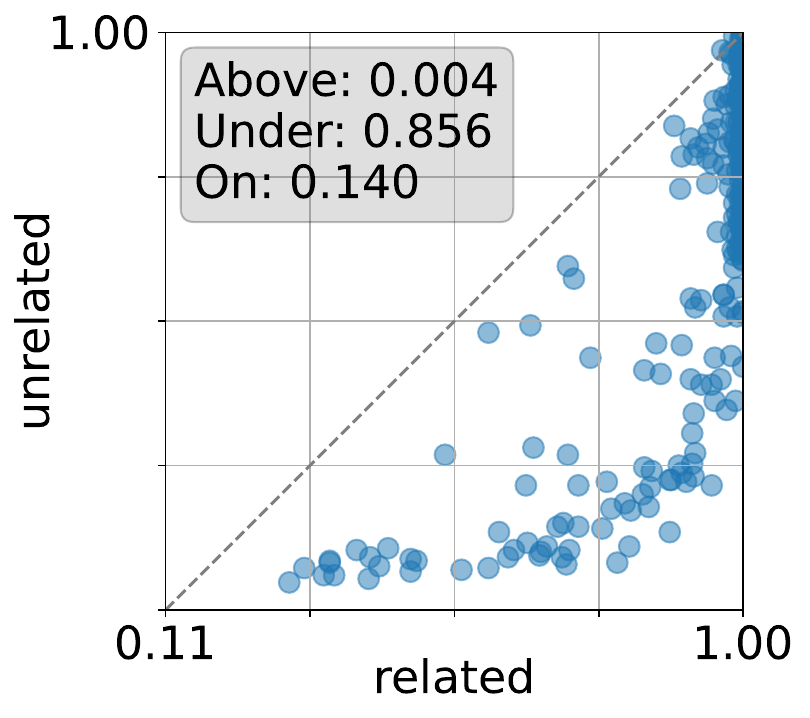} &
\includegraphics[width=0.15\textwidth]{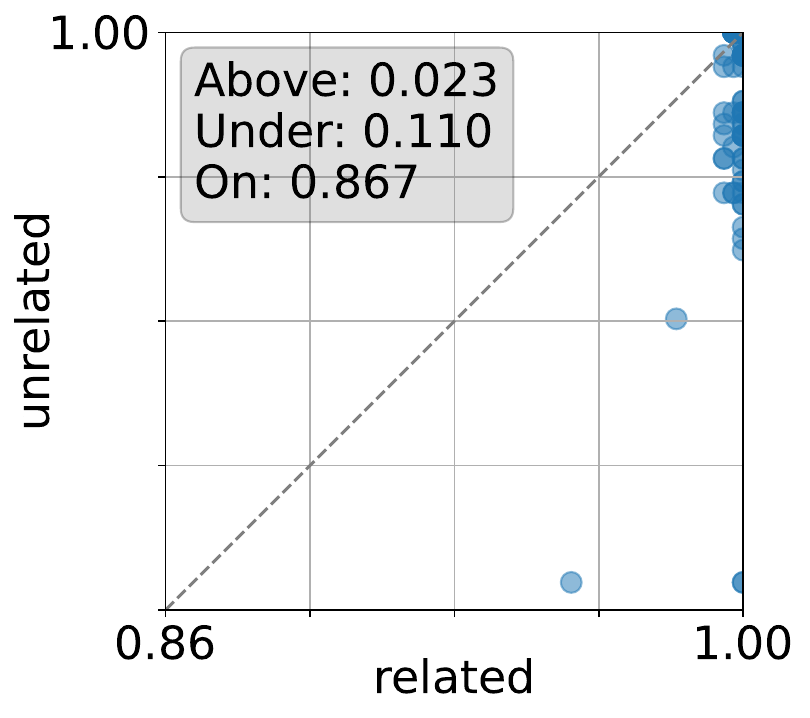} &
\includegraphics[width=0.15\textwidth]{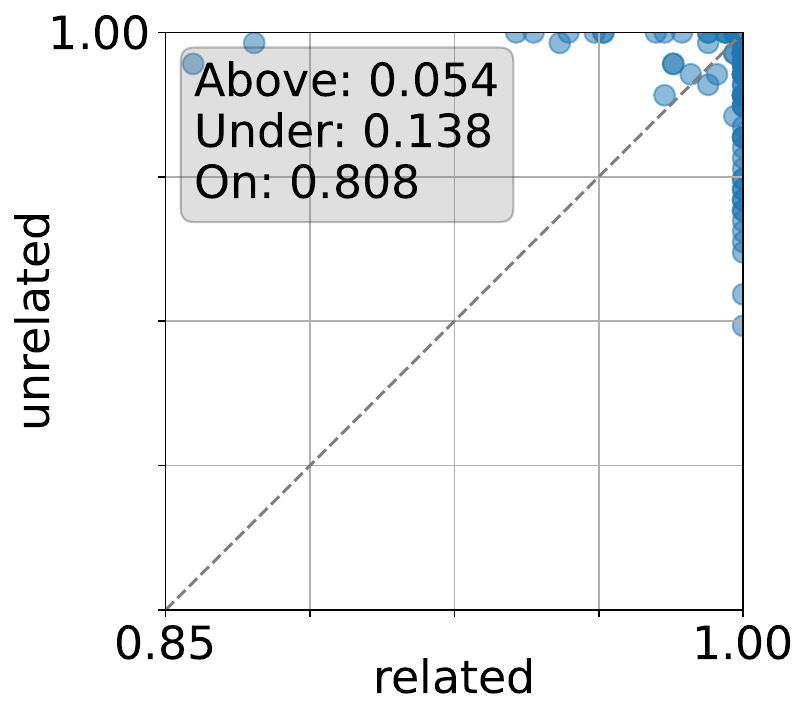} &
\includegraphics[width=0.15\textwidth]{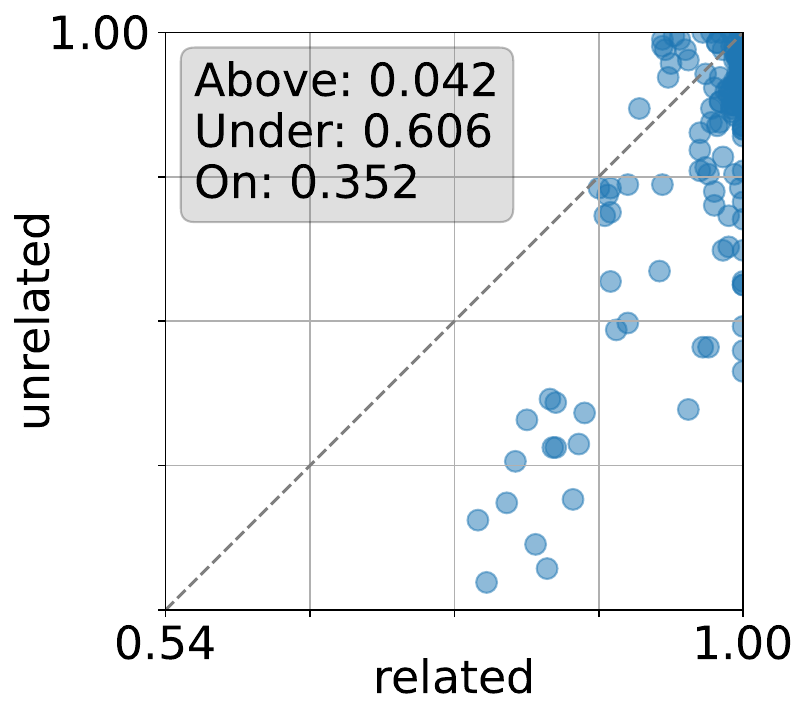} &
\includegraphics[width=0.15\textwidth]{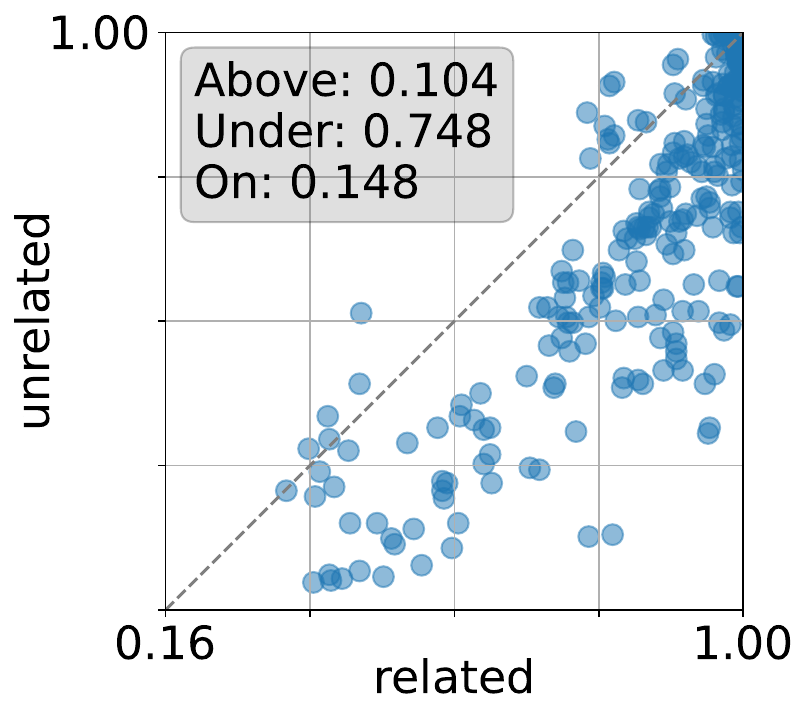} \\
\includegraphics[width=0.15\textwidth]{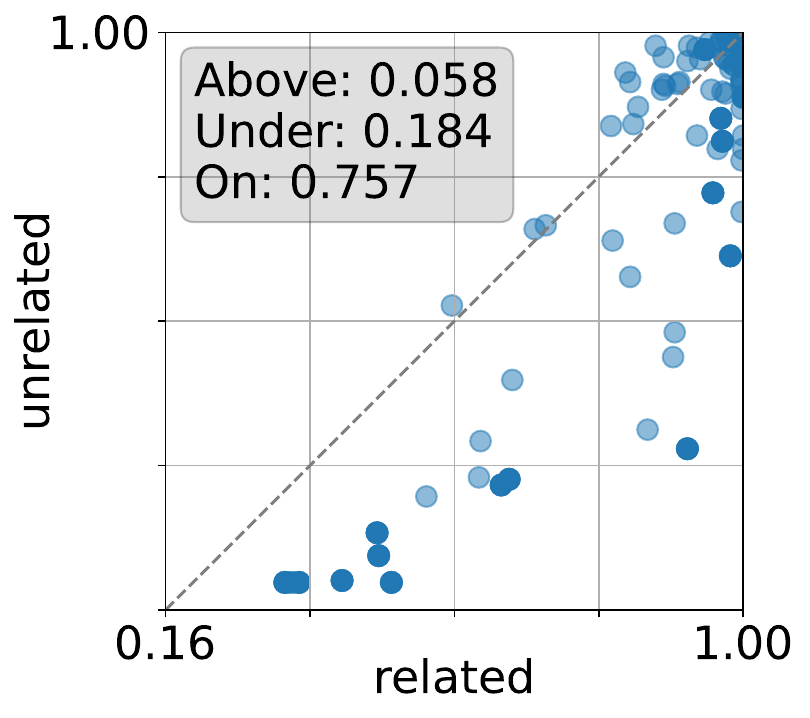} &
\includegraphics[width=0.15\textwidth]{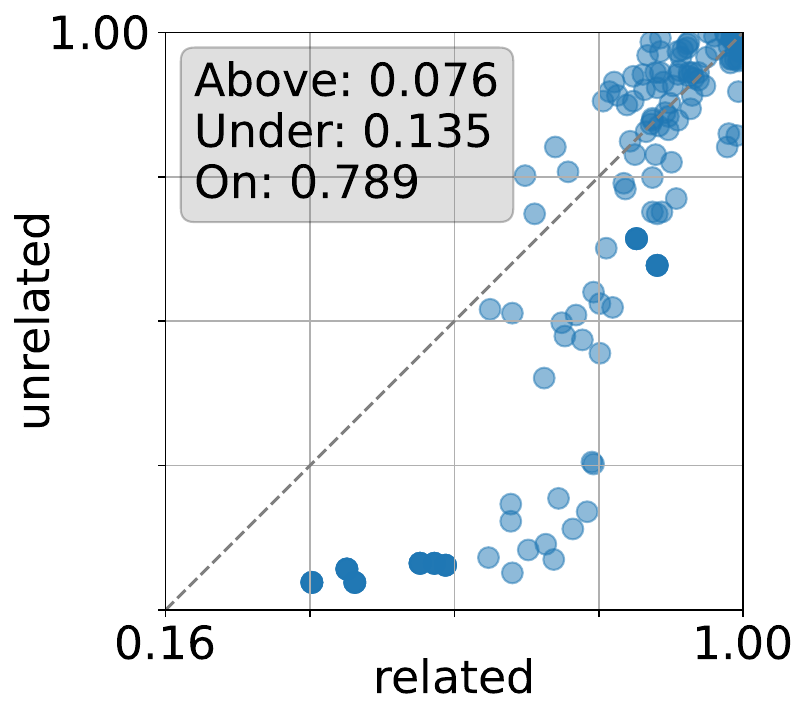} &
\includegraphics[width=0.15\textwidth]{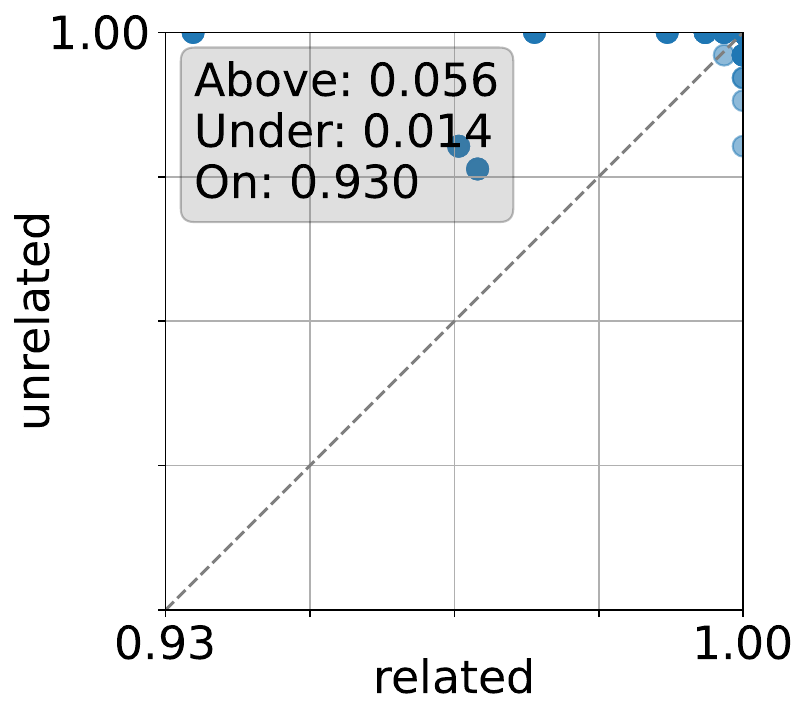} &
\includegraphics[width=0.15\textwidth]{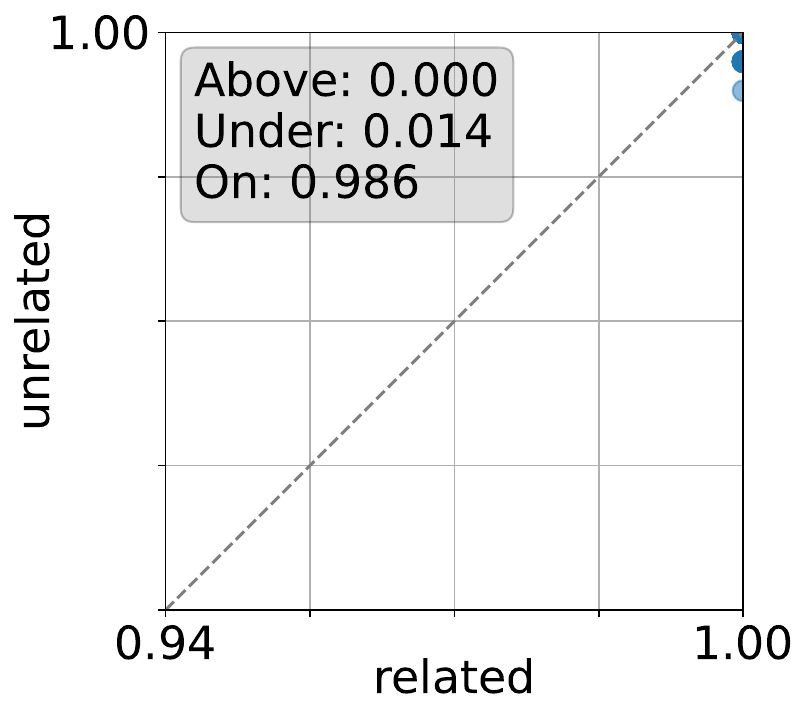} &
\includegraphics[width=0.15\textwidth]{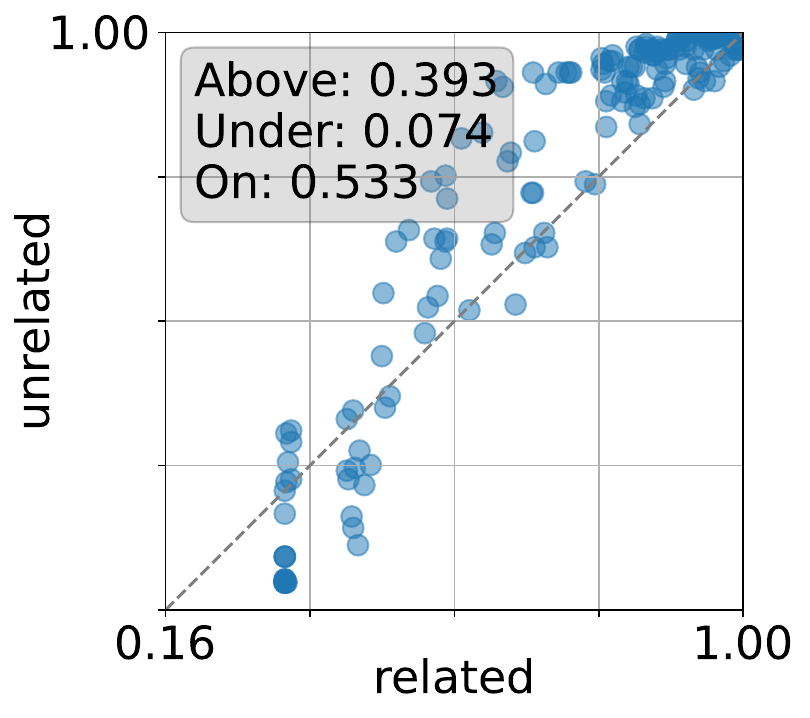} &
\includegraphics[width=0.15\textwidth]{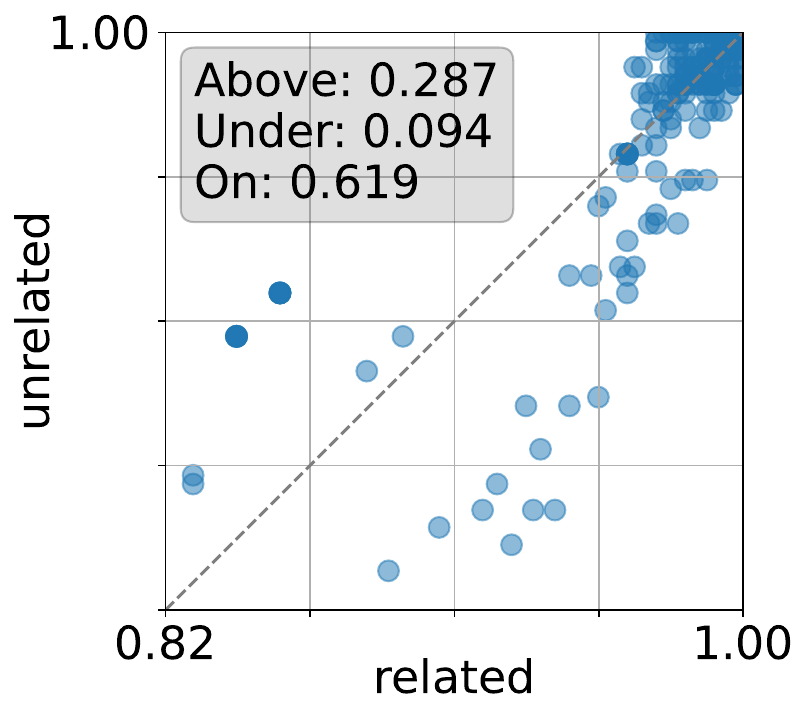} &
\includegraphics[width=0.15\textwidth]{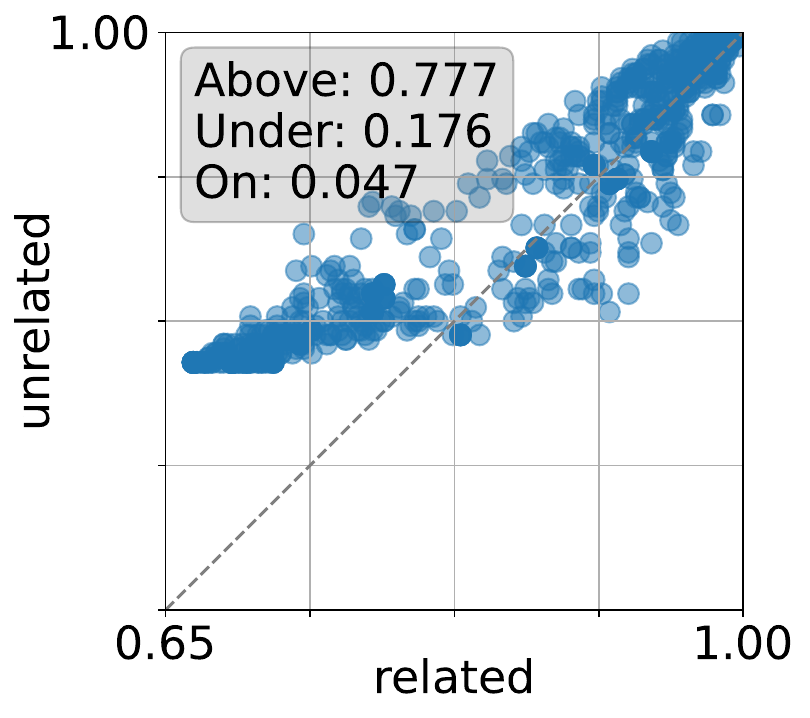} \\
\end{tabular}
\caption{Scatter plots illustrating the relationship of unrelated and related word sets in the NL1 prompt category
(top) and number sets in the CS prompt category (bottom).
A point belongs to the same prompt template, the two coordinates are average accuracies over queries with related and unrelated
sets of the given type.}
\label{fig:scatter}
\end{figure*}

\section{Experimental Results}

The experiment design described in \cref{sec:setup} was executed on
Llama-3.2-3B, Llama-3.1-8B, Llama-3.1-70B, and Llama-3.3-70B \cite{llamacitation},
Mistral-24B \cite{mistralcitation},
Qwen2.5-32B \cite{qwencitation}, and
Phi-3.5-MoE \cite{phicitation}.
Each model is instruction-tuned.

Overall, we executed more than 60 million prompts forming a complete grid covering every possible
combination of sets, prompts, query types, and LLMs (see \cref{sec:setupnotes}).
The overall accuracy of each LLM is shown in \cref{table:llmstats}. 
As we can see, performance is reasonably good overall, but since
each LLM was tested 8,709,120 times, the small error rate still
indicates a large number of errors (about 790,000 altogether).
In the following, we will focus mostly on these errors.

\textbf{Motivation.} We will examine in more detail three dimensions, to which \emph{the set
membership problem is supposed to be invariant}: (1) variations of
the prompts that formulate the same query (2) the order in which we present the members
of the set in the prompt and (3) the semantic features and relations of the members of the set.
We will demonstrate that along all these dimensions \emph{there is significant variance} in
the error patterns.
We also demonstrate that the different LLMs show substantial 
differences in their error patterns as well.

\subsection{Prompt sensitivity}

All our queries were executed using a large number of prompt templates.
These templates are differentiated by major features---natural language (NL1, NL2) or Python code (CS, CA)---as well as minor features such as the exact selection of wording and punctuation.

\textbf{Prompt categories.} 
\Cref{fig:promptcos} illustrates the similarity between prompt categories.
We computed cosine similarities between 3024 dimensional vectors
that represent the set of queries we defined (see \cref{sec:setupnotes}).
The vector of a prompt category contains, for each query, the average
error-rate of those prompts from the given prompt category that implement the given query.

The conclusion we can draw is that the Python code prompt types CA and CS are similar, also
the natural language prompts that use the same \emph{arrangement} feature (that is, they present the
set and the membership question in the same order) are somewhat similar as well, although
less consistently so.
Otherwise, the error pattern is not similar between different prompt categories, despite
the fact the very same set membership queries are instantiated over the same LLM,
and the different LLMs also show a large variability.

\textbf{Prompt features.}
We also tested the effect of minor features such as wording and punctuation.
While different explanation methods can give different results~\cite{Kris24}, here, we
opted for Shapley values~\cite{Lund17} that describe the impact of
individual features on the accuracy (\cref{fig:shap}).
For a given LLM, we performed the analysis over the set of all the prompts that
belong to a certain prompt category. 
For example, in the NL1 category, this means $3024\cdot 480$ prompts (see \cref{sec:setupnotes}).

Note that the most important feature is different in every LLM shown, and in general,
the importance of the same feature can vary radically across LLMs. 
Still, some features have a consistent effect on the success of the models, despite being
designed to be indifferent to the formulation of the query.
One such feature is the \emph{arrangement} feature (set-first or element-first prompt).
Interestingly, \emph{different LLMs might strongly prefer opposite arrangements}.

\subsection{Permutation sensitivity}

Another interesting question is whether the order in which the set elements are listed has an effect
on the answer.
Clearly, the concept of the set suggests that the element ordering should not matter.
Yet, we see many counterexamples, like the one below, where
Llama3.3-70B answered the first question correctly, but the
second one incorrectly:

\prompt{Does the set (Hearts, Clubs, Spades) contain the character sequence East?}
\prompt{Does the set (Hearts, Spades, Clubs) contain the character sequence East?}

\Cref{fig:consistency} offers a more detailed insight, visualizing the sensitivity to ordering.
To understand the plots, recall, that we execute separate queries for each different ordering of the same set.
Now, if we group those queries together that are the same except the ordering of the set (thereby defining order-independent queries), we can compute
the fraction of such order-independent queries that receive a consistent answer (i.e., the same answer for every permutation of the set elements).
The figure shows statistics of the consistency of order-independent queries using a bar plot as a function of set-type and prompt-type.

Inconsistency seems to be relatively low at first sight (a few percent of the order-independent queries),
but that is because accuracy is high in general.
In fact, inconsistency is much higher than the overall error rate (see \cref{table:llmstats}), indicating that many sets
have ``adversarial'' orderings, in which the query fails.
Interestingly, this is true even for the Python code prompts, to some extent, which is perhaps more
surprising than the natural language case, given the formal nature of the domain.
Note also the diversity among LLMs.

\subsection{Semantic sensitivity}

The semantic features of set elements should also be ignored by the design of
the set membership task.
Yet, we observe a strong semantic interference.

\textbf{Semantic leakage.} \Cref{table:llmstats} presents the false positive rate (FPR) percentages for each investigated model,
aggregated over every prompt type, for negative-member and negative-intruder query types.
Apart from Llama-3.1-70B, FPRs are substantially larger for  member words than for intruder words.
In other words, when the correct answer is ``not a member'', most LLMs are more likely to say ``member'' when
the word tested for membership is semantically related to the set elements.
This is a clear case of semantic leakage.

\textbf{Semantic boosting.} \Cref{fig:scatter} illustrates a phenomenon that can be considered the opposite
to semantic leakage, namely when semantic relatedness increases accuracy, as opposed to decreasing it.
The scatter plots visualize the comparison of fixed prompt templates when the same template is
applied to sets with and without semantically related elements.
When points are under the diagonal, the errors tend to be smaller for semantically related sets.
In the case of the NL1 prompt family, this is the case in every LLM.

However, for number sets with the CS prompt family, we can observe an almost opposite trend,
where sets of related numbers seem to generate more errors.
It is interesting, though, that the 70B Llama models are an exception, where the bias still
favors the related sets.

\textbf{Semantic preference.} \Cref{fig:settype} shows how the accuracy of different LLMs depends on
the semantic type of the set in the query, for the case of the NL1 and CS prompt categories.
The dependence on set-type is the clearest in the case of CS prompts, where number sets perform
worse than word sets (but note that some LLMs do not show this bias).
In the case of the NL1 prompts, there is a slight bias against sets of unrelated words.

\subsection{LLM Sensitivity}
\label{sec:llmsensitivity}

\begin{figure}
\centering
\begin{tabular}{@{}c@{}c@{}}
\small NL1 & \small NL2 \\
\includegraphics[width=.5\columnwidth]{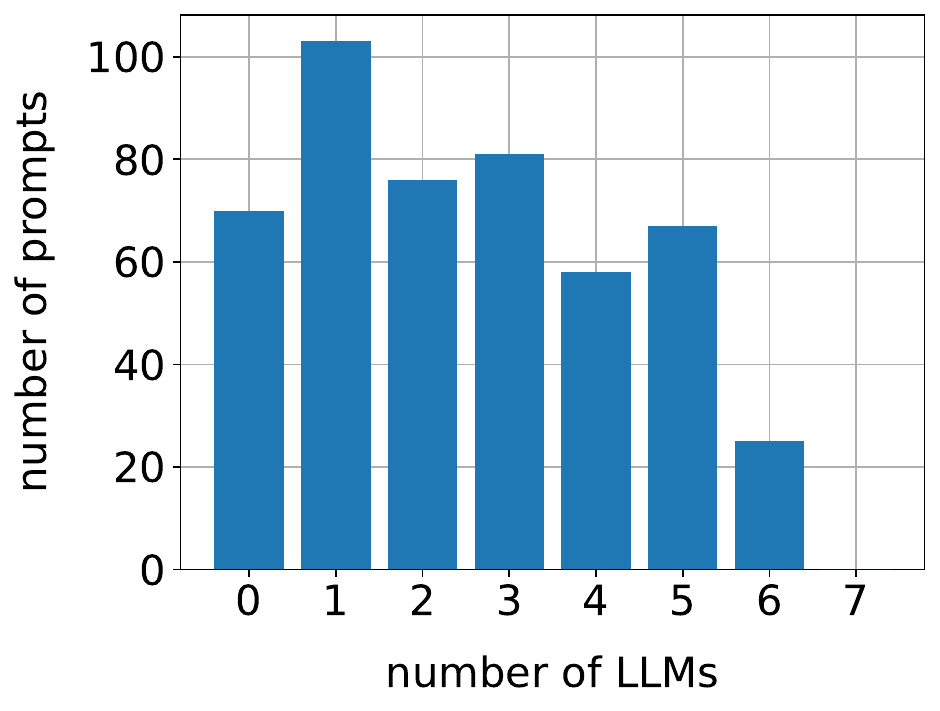} &
\includegraphics[width=.5\columnwidth]{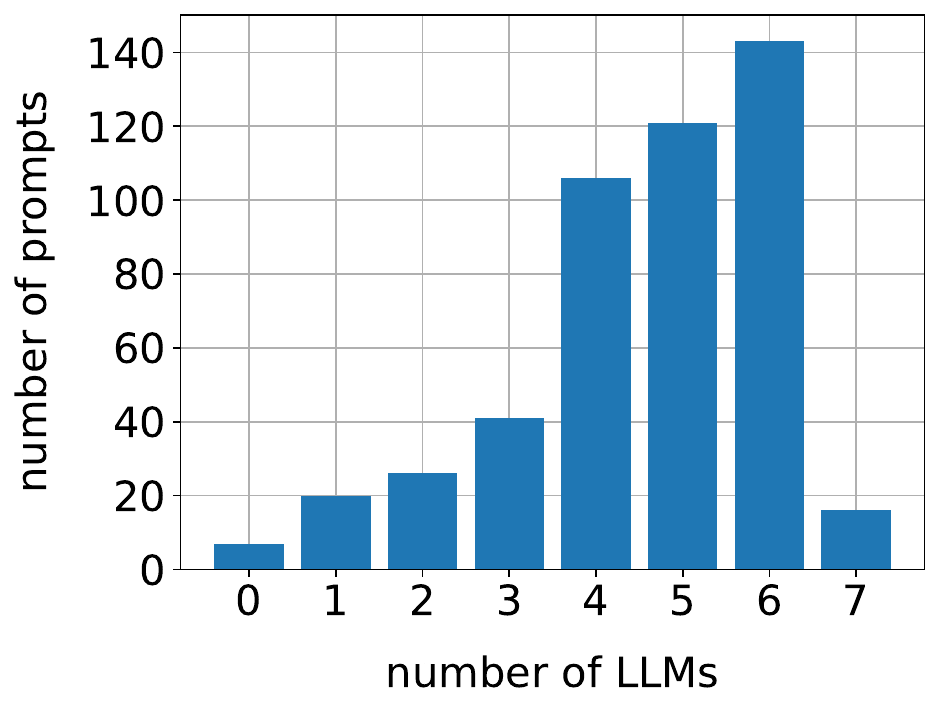} \\
\\ \small CS & \small CA \\
\includegraphics[width=.5\columnwidth]{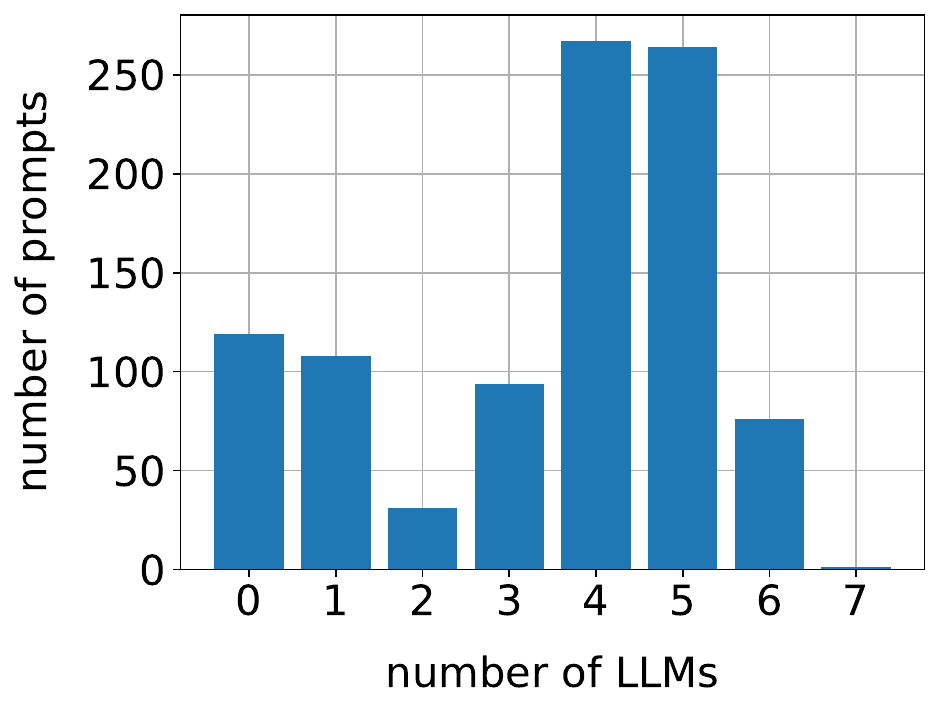} &
\includegraphics[width=.5\columnwidth]{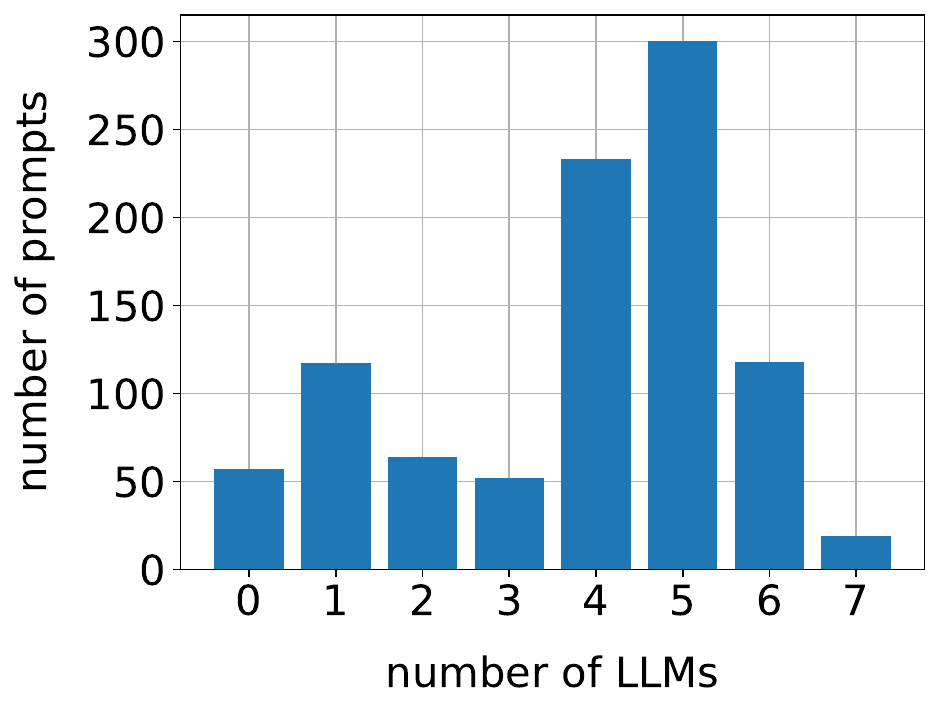} \\
\end{tabular}
\caption{Column $i$ shows the number of prompt templates that were perfect (resulted in a correct answer for every query) on exactly $i$ LLMs.}
\label{fig:perfectprompt}
\end{figure}

Initially, we expected to see similar error patterns over the different LLMs, but our results contradict this
hypothesis.
Consider, for example, \cref{fig:perfectprompt} that illustrates the success of the different prompt templates
over our set of 7 LLMs.
If the LLMs had very similar error patterns then we would see most of the prompt templates in either column 0 or 7, but
this is not the case.
For our four classes of prompts we observe slightly different distributions, but in each case, there are
only a few templates that are perfect---result in a correct answer in every single query over all the LLMs---but there
are many templates that are specific to one or two LLMs, especially in the case of NL1 prompts.
This is despite the fact that the overall accuracy is rather high for each LLM.
In the NL1 class, there is in fact no perfect template and \emph{the highest number of templates are perfect only
on a single LLM}.

Let us illustrate some prompt templates, instantiated with the query instance $[$(North, South, West, East), North$]$.
Examples for perfect templates in NL2 and CS include

\prompt{A set consists of the following character sequences: \{'North', 'South', 'West', 'East'\}\\
Does this set include the character sequence 'North'?}

\prompt{Given the Python code below, what will be the value of the `result` variable?\\
\\
```python\\
def contains(s, item):\\
\hphantom{xxxx}return item in s\\
\\
my\_set = set(('North', 'South', 'West', 'East'))\\
result = contains(my\_set, 'North')\\
```}

Examples for CS and NL1 templates that are never perfect include

\prompt{Given the Python code below, what will be the output?\\
\\
```python\\
def contains(s, item):\\
\hphantom{xxxx}return item in s\\
\\
S = set("North South West East".split())\\
print(contains(S, "North"))\\
```}

\prompt{Does the set $[$North, South, West, East$]$ include the string North?}


\begin{figure}
\centering
\includegraphics[width=\columnwidth]{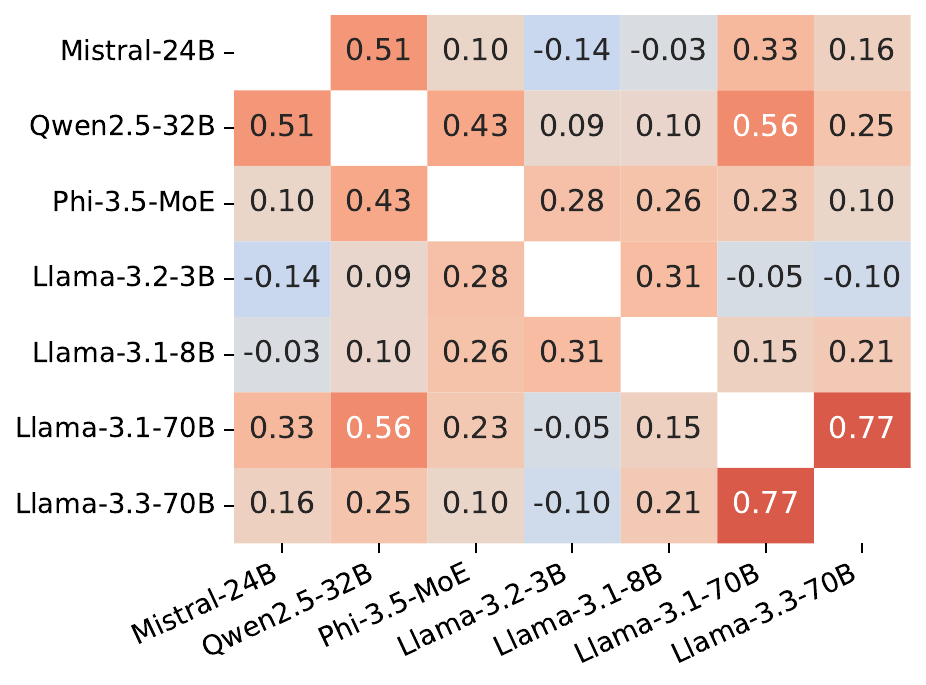}
\caption{Normalized pointwise mutual information between the error patterns of the LLMs.}
\label{fig:pmi}
\end{figure}

To investigate LLM similarity further, we computed the normalized pointwise mutual information
between the error patterns of the LLMs, as shown in \cref{fig:pmi}.
In more detail, for a pair of LLMs $a$ and $b$ we define
the random variables $x_a$ and $x_b$ as indicator variables of giving an incorrect answer
for the same random prompt by LLM $a$ and $b$, respectively.
We assume a uniform distribution over the possible 8,709,120 prompts.
We then compute the normalized pointwise mutual information for the point $(x_a=1,x_b=1)$, i.e., the case
when both LLMs are incorrect, according to
\begin{equation*}
\mbox{npmi}(1,1) = \left[\log\frac{p_{a,b}(1,1)}{p_a(1)p_b(1)}\right]/-\log p_{a,b}(1,1).
\end{equation*}
The maximal value is 1, and 0 indicates independence.

Overall, in case of many pairs, the similarity is above the baseline level of 0, yet fairly low,
except between the two 70B Llama models.
It is interesting that the Qwen model is somewhat similar to three models from different families that are not necessarily similar to
each other.

Finally, the diversity of LLM behavior is also well illustrated by some of the previously discussed
results, when focusing on the differences between LLMs.
The markedly different LLM behaviors include the different (sometimes opposite) effects of both minor (\cref{fig:shap})
and major (\cref{fig:promptcos,fig:settype}) prompt features, the different (in some cases, opposite) effect of semantic
relatedness of the set elements (\cref{fig:scatter}), especially in the case of number sets, and the different sensitivity to the ordering of the set
elements (\cref{fig:consistency}).

\section{Conclusion}

In this paper, we formulated the simplest possible problem
we could think of: set membership on very small, explicitly given sets.
Our goal was to examine the brittleness of instruction-tuned LLMs over this
very basic task.

Our work uncovers a rather comprehensive set of brittleness issues. 
Most importantly, we found a strong sensitivity
to three aspects the LLMs should be robust to in the context of set membership:
prompt features, element ordering, and semantic relatedness.
We also found that different LLMs often behave very differently.
These results provide multi-faceted evidence for the \emph{lack of a crisp and robust
set concept in LLMs}.

We also believe that \emph{our experimental design and analysis
could be used as a benchmark}, beside the usual standard benchmarks based only on a few
performance metrics that do not offer a lot of insight into potential design flaws.
In a more general sense, we propose a valuable methodology
for LLM evaluation that can be applied using other extremely simple problems as well.

\section*{Limitations}
A limitation of our work is that---due to attempting to provide a comprehensive map
of issues---we have not explored in depth the several directions we uncovered that look
interesting and promising for understanding the set concept of LLMs better. 
This would involve looking ``under the hood'' using,
e.g., mechanistic interpretability techniques~\cite{Shar25}, or even human experiments to
understand whether certain brittleness properties (e.g., semantic leakage or boosting) are 
specific to LLMs.
The approach of~\cite{Yiu25} using the performance of children on benchmarks
is a possible starting point. 

\section*{Acknowledgements}

Support from Project 2024-1.2.3-HU-RIZONT-2024-00017 is acknowledged,
financed by the Ministry of Culture and Innovation of Hungary from the National Research,
Development and Innovation Fund, under the a 2024-1.2.3-HU-RIZONT funding
scheme.
Support from grants NSF 2217023 and NSF 2240532 is acknowledged.

\bibliography{refs}

@article{coy24,
author = {R. T. McCoy  and S. Yao  and D. Friedman  and M. D. Hardy  and T. L. Griffiths },
title = {Embers of autoregression show how large language models are shaped by the problem they are trained to solve},
journal = {Proceedings of the National Academy of Sciences},
volume = {121},
number = {41},
pages = {e2322420121},
year = {2024}}

@misc{llamacitation,
  title={The Llama 3 Herd of Models},
  author={Dubey, Abhimanyu and Jauhri, Abhinav and Pandey, Abhinav and Kadian, Abhishek and Al-Dahle, Ahmad and Letman, Aiesha and Mathur, Akhil and Schelten, Alan and Yang, Amy and Fan, Angela and others},
  year={2024},
  eprint={2407.21783},
  archivePrefix={arXiv},
  primaryClass={cs.AI},
  url={https://arxiv.org/abs/2407.21783}
}

@misc{mistralcitation,
  title={Mistral 7B},
  author={Jiang, Albert Q. and Sablayrolles, Alexandre and Mensch, Arthur and Bamford, Chris and Chaplot, Devendra Singh and Casas, Diego de las and Bressand, Florian and Lengyel, Gianna and Lample, Guillaume and Saulnier, Lucile and others},
  year={2023},
  eprint={2310.06825},
  archivePrefix={arXiv},
  primaryClass={cs.CL},
  url={https://arxiv.org/abs/2310.06825}
}

@misc{qwencitation,
  title={Qwen2.5: A Party of Foundation Models},
  author={Qwen Team},
  year={2024},
  howpublished={Alibaba Cloud},
  url={https://qwenlm.github.io/blog/qwen2.5/}
}

@misc{phicitation,
  title={Phi-3 Technical Report: A Highly Capable Language Model Locally on Your Phone},
  author={Abdin, Marah and Jacobs, Sam Ade and Awan, Ammar Ahmad and Aneja, Jyoti and Awadallah, Ahmed and Awadalla, Hany and Bach, Nguyen and Bahree, Amit and Bakhtiari, Arash and Behl, Harkirat and others},
  year={2024},
  eprint={2404.14219},
  archivePrefix={arXiv},
  primaryClass={cs.CL},
  url={https://arxiv.org/abs/2404.14219}
}

@article{lampinen2024language,
  title={Language models, like humans, show content effects on reasoning tasks},
  author={Lampinen, A K and Dasgupta, I and Chan, S C Y and others},
  journal={PNAS Nexus},
  volume={3},
  pages={233},
  year={2024},
  publisher={Oxford University Press}
}

@article{mit21,
  author       = {M. Mitchell},
  title        = {Why {AI} is Harder Than We Think},
  journal      = {CoRR},
  volume       = {abs/2104.12871},
  year         = {2021}
}

@article{mit25,
  author       = {M. Lewis and
                  M. Mitchell},
  title        = {Evaluating the Robustness of Analogical Reasoning in Large Language
                  Models},
  journal      = {Trans. Mach. Learn. Res.},
  year         = {2025}
}

@article{Kris24,
  author       = {S. Krishna and
                  T. Han and
                  A. Gu and
                  S. Wu and
                  S. Jabbari and
                  H. Lakkaraju},
  title        = {The Disagreement Problem in Explainable Machine Learning: {A} Practitioner's
                  Perspective},
  journal      = {Trans. Mach. Learn. Res.},
  year         = {2024}
}

@inproceedings{Lund17,
  author       = {S. M. Lundberg and
                  S.{-}I. Lee},
  title        = {A Unified Approach to Interpreting Model Predictions},
  booktitle    = {Advances in Neural Information Processing Systems 30: Annual Conference
                  on Neural Information Processing Systems 2017},
  pages        = {4765--4774},
  year         = {2017}
}

@misc{shojaee2025illusion,
  title         = {The Illusion of Thinking: Understanding the Strengths and Limitations of Reasoning Models via the Lens of Problem Complexity},
  author        = {Parshin Shojaee and Iman Mirzadeh and Keivan Alizadeh and Maxwell Horton and Samy Bengio and Mehrdad Farajtabar},
  year          = {2025},
  eprint        = {2506.06941},
  archivePrefix = {arXiv},
  primaryClass  = {cs.AI},
  url           = {https://arxiv.org/abs/2506.06941}
}

@article{cuskley2024limitations,
  title={The Limitations of Large Language Models for Understanding Human Language and Cognition},
  author={Cuskley, Christine and Woods, Rebecca and Flaherty, Molly},
  journal={Open Mind},
  volume={8},
  pages={1058--1083},
  year={2024},
  publisher={MIT Press}
}

@article{elaz21,
    title = "Measuring and Improving Consistency in Pretrained Language Models",
    author = {Elazar, Y.  and
      Kassner, N.  and
      Ravfogel, S.  and
      Ravichander, A.  and
      Hovy, E.  and
      Sch{\"u}tze, H.  and
      Goldberg, Y.},
    journal = "Transactions of the Association for Computational Linguistics",
    volume = "9",
    year = "2021",
    pages = "1012--1031",
}

@inproceedings{fier22,
    title = "Factual Consistency of Multilingual Pretrained Language Models",
    author = "Fierro, C.  and
      S{\o}gaard, A.",
    year = "2022",
    address = "Dublin, Ireland",
    publisher = "Association for Computational Linguistics",
    pages = "3046--3052"
}

@misc{raj23,
      title={Measuring Reliability of Large Language Models through Semantic Consistency}, 
      author={H. Raj and D. Rosati and S. Majumdar},
      year={2023}
}

@misc{sahu22,
      title={Unpacking Large Language Models with Conceptual Consistency}, 
      author={P. Sahu and M. Cogswell and Y. Gong and A. Divakaran},
      year={2022},
      eprint={2209.15093},
      archivePrefix={arXiv}
}

@misc{zheng24rel,
      title={How Reliable are LLMs as Knowledge Bases? Re-thinking Factuality and Consistency}, 
      author={D. Zheng and M. Lapata and J. Z. Pan},
      year={2024},
      eprint={2407.13578},
      archivePrefix={arXiv}
}

@article{Pass24,
  author       = {A. Passerini and
                  A. Gema and
                  P. Minervini and
                  B. Sayin and
                  K. Tentori},
  title        = {Fostering effective hybrid human-LLM reasoning and decision making},
  journal      = {Frontiers Artif. Intell.},
  volume       = {7},
  year         = {2024}
}

@inbook{Yonel,
  title =	{Recognition memory: The role of recollection and familiarity},
  booktitle =	 {Oxford Handbook of Human Memory},
  publisher =	 {Oxford Handbook of Human Memory, Oxford Univ. Press},
  author =	 {Yonelinas, A. P. and Ramey, M. M. and Riddell, C.},
  editor =	 {Kahane, M. J. and Wagner, A. D.},
  year =	 2022
  }

@inproceedings{wolfram2025world,
  title={World Models and Consistent Mistakes in LLMs},
  author={Wolfram, Christopher and Schein, Aaron},
  booktitle={Proceedings of the 42nd International Conference on Machine Learning},
  series={ICML '25},
  year={2025},
  organization={PMLR}
}

@book{JohnL,
  title={Mental Models. Towards a Cognitive Science of Language, Inference and Consciousness},
  author={Johnson-Laird, P. N.},
  year={1983},
  publisher={Cambridge Univ. Press}
}

@inproceedings{Bent20,
  author       = {G. W. Benton and
                  M. Finzi and
                  P. Izmailov and
                  A. G. Wilson},
  title        = {Learning Invariances in Neural Networks from Training Data},
  booktitle    = {Advances in Neural Information Processing Systems 33: Annual Conference
                  on Neural Information Processing Systems 2020, NeurIPS 2020},
  year         = {2020}
}

@article{Dew06,
  title={Measuring the speed of the conscious components of recognition memory: Remembering is faster than knowing},
  author={Dewhurst, S. A. and Holmes, S. J. and Brandt, K. R. and Dean, G. M.},
  journal={Consciousness and Cognition},
  volume={15},
  pages={147-162},
  year={2006}
}

@article{Wais10,
  author       = {P. E. Wais and
                  L. R. Squire and
                  J. T. Wixted},
  title        = {In Search of Recollection and Familiarity Signals in the Hippocampus},
  journal      = {J. Cogn. Neurosci.},
  volume       = {22},
  pages        = {109--123},
  year         = {2010}
}

@inproceedings{Khem14,
  author       = {S. Khemlani and
                  M. Lotstein and
                  P. Johnson{-}Laird},
  title        = {A mental model theory of set membership},
  booktitle    = {Proceedings of the 36th Annual Meeting of the Cognitive Science Society, CogSci 2014},
  year         = {2014}
}

@article{wason1968reasoning,
  title={Reasoning about a rule},
  author={Wason, Peter C},
  journal={Quarterly journal of experimental psychology},
  volume={20},
  number={3},
  pages={273--281},
  year={1968}
}

@inproceedings{Gonen25,
title = "Does Liking Yellow Imply Driving a School Bus? Semantic Leakage in Language Models",
    author = "Gonen, Hila  and
      Blevins, Terra  and
      Liu, Alisa  and
      Zettlemoyer, Luke  and
      Smith, Noah A.",
    editor = "Chiruzzo, Luis  and
      Ritter, Alan  and
      Wang, Lu",
    booktitle = "Proceedings of the 2025 Conference of the Nations of the Americas Chapter of the Association for Computational Linguistics: Human Language Technologies (Volume 1: Long Papers)",
    month = apr,
    year = "2025",
    address = "Albuquerque, New Mexico",
    publisher = "Association for Computational Linguistics",
    url = "https://aclanthology.org/2025.naacl-long.35/",
    doi = "10.18653/v1/2025.naacl-long.35",
    pages = "785--798",
    ISBN = "979-8-89176-189-6",
}

@article{Dasgu22,
  author       = {I. Dasgupta and
                  A. K. Lampinen and
                  S. C. Y. Chan and
                  A. Creswell and
                  D. Kumaran and
                  J. L. McClelland and
                  F. Hill},
  title        = {Language models show human-like content effects on reasoning},
  journal      = {CoRR},
  volume       = {abs/2207.07051},
  year         = {2022}
}

@article{Gyor25,
  author       = {A. Gy{\"{o}}rgy and
                  T. Lattimore and
                  N. Lazic and
                  Cs. Szepesv{\'{a}}ri},
  title        = {Beyond Statistical Learning: Exact Learning Is Essential for General Intelligence},
  journal      = {CoRR},
  volume       = {abs/2506.23908},
  year         = {2025}
}

@inproceedings{Huck25,
  author       = {J. Huckle and
                  S. Williams},
  title        = {Easy Problems that LLMs Get Wrong},
  booktitle    = {Future of Information and Communication Conference (FICC 2025)},
  pages        = {313--332},
  year         = {2025}
}

@misc{egr25,
      title={Set-LLM: A Permutation-Invariant LLM}, 
      author={B. Egressy and J. Stühmer},
      year={2025},
      eprint={2505.15433},
      archivePrefix={arXiv}
}

@inproceedings{Zahe17,
  author       = {M. Zaheer and
                  S. Kottur and
                  S. Ravanbakhsh and
                  B. P{\'{o}}czos and
                  R. Salakhutdinov and
                  A. J. Smola},
  title        = {Deep Sets},
  booktitle    = {Advances in Neural Information Processing Systems 30: Annual Conference
                  on Neural Information Processing Systems 2017},
  pages        = {3391--3401},
  year         = {2017}
}

@techreport{Razm,
  author      = "A. Razmjooei",
  title       = "Investigation of Some Cognitive Difficulties in Set Theory",
  institution = "University of Stockholm",
  year        = "2013"
}

@inproceedings{Yiu25,
  author       = {E. Yiu and
                  M. Qraitem and
                  A. Noor Majhi and
                  C. Wong and
                  Y. Bai and
                  S. Ginosar and
                  A. Gopnik and
                  K. Saenko},
  title        = {KiVA: Kid-inspired Visual Analogies for Testing Large Multimodal Models},
  booktitle    = {The Thirteenth International Conference on Learning Representations,
                  {ICLR} 2025},
  year         = {2025}
}

@inproceedings{Rava17,
  author       = {S. Ravanbakhsh and
                  J. G. Schneider and
                  B. P{\'{o}}czos},
  title        = {Equivariance Through Parameter-Sharing},
  booktitle    = {Proceedings of the 34th International Conference on Machine Learning,
                  {ICML} 2017},
  volume       = {70},
  pages        = {2892--2901},
  publisher    = {{PMLR}},
  year         = {2017}
}

@article{Shar25,
title={Open Problems in Mechanistic Interpretability},
author={Lee Sharkey and Bilal Chughtai and Joshua Batson and Jack Lindsey and Jeffrey Wu and Lucius Bushnaq and Nicholas Goldowsky-Dill and Stefan Heimersheim and Alejandro Ortega and Joseph Isaac Bloom and Stella Biderman and Adri{\`a} Garriga-Alonso and Arthur Conmy and Neel Nanda and Jessica Mary Rumbelow and Martin Wattenberg and Nandi Schoots and Joseph Miller and William Saunders and Eric J Michaud and Stephen Casper and Max Tegmark and David Bau and Eric Todd and Atticus Geiger and Mor Geva and Jesse Hoogland and Daniel Murfet and Thomas McGrath},
journal={Transactions on Machine Learning Research},
issn={2835-8856},
year={2025},
url={https://openreview.net/forum?id=91H76m9Z94},
note={Survey Certification}
}

@article{Zhou24,
  author       = {L. Zhou and
                  W. Schellaert and
                  F. Mart{\'{\i}}nez{-}Plumed and
                  Y. Moros{-}Daval and
                  C. Ferri and
                  J. Hern{\'{a}}ndez{-}Orallo},
  title        = {Larger and more instructable language models become less reliable},
  journal      = {Nat.},
  volume       = {634},
  number       = {8032},
  pages        = {61--68},
  year         = {2024}
}

@inproceedings{Zhuo24,
  author       = {J. Zhuo and
                  S. Zhang and
                  X. Fang and
                  H. Duan and
                  D. Lin and
                  K. Chen},
  title        = {Pro{SA}: Assessing and Understanding the Prompt Sensitivity of LLMs},
  booktitle    = {Findings of the ACL: {EMNLP}
                  2024},
  pages        = {1950--1976},
  year         = {2024}
}

@inproceedings{sclar2024quantifying,
title={Quantifying Language Models' Sensitivity to Spurious Features in Prompt Design or: How I learned to start worrying about prompt formatting},
author={Melanie Sclar and Yejin Choi and Yulia Tsvetkov and Alane Suhr},
booktitle={The Twelfth International Conference on Learning Representations},
year={2024},
url={https://openreview.net/forum?id=RIu5lyNXjT}
}

@inproceedings{wolf-etal-2020-transformers,
    title = "Transformers: State-of-the-Art Natural Language Processing",
    author = "Thomas Wolf and Lysandre Debut and Victor Sanh and Julien Chaumond and Clement Delangue and Anthony Moi and Pierric Cistac and Tim Rault and Rémi Louf and Morgan Funtowicz and Joe Davison and Sam Shleifer and Patrick von Platen and Clara Ma and Yacine Jernite and Julien Plu and Canwen Xu and Teven Le Scao and Sylvain Gugger and Mariama Drame and Quentin Lhoest and Alexander M. Rush",
    booktitle = "Proceedings of the 2020 Conference on Empirical Methods in Natural Language Processing: System Demonstrations",
    month = oct,
    year = "2020",
    address = "Online",
    publisher = "Association for Computational Linguistics",
    url = "https://www.aclweb.org/anthology/2020.emnlp-demos.6",
    pages = "38--45"
}

\section*{Appendix}
\begin{table*}[]
\centering
\setlength{\tabcolsep}{4pt}
\begin{tabular}{lll}
\textbf{Set Type} & \textbf{Intruder Element} & \textbf{Semantic Connection} \\ \midrule\midrule
\textit{Complete Sets} \\ \midrule
 North, West, East, South & incisors & cardinal directions \\
NE, NW, SE, SW & plasma & ordinal directions \\ 
freshman, sophomore, junior, senior & Thymine & years in high school \\
Gryffindor, Hufflepuff, Ravenclaw, Slytherin & Spades & houses of Hogwarts \\ 
earth, fire, air, water & molars & classical elements \\
solid, liquid, gas, plasma & freshman & fundamental states of matter \\
incisors, canines, premolars, molars & multiplication & types of human teeth \\
addition, subtraction, multiplication, division & Hufflepuff & basic arithmetic operations \\ 
Hearts, Diamonds, Clubs, Spades & East & French card suits \\
Adenine, Thymine, Cytosine, Guanine & canines & bases of DNA\\
%
\midrule\midrule
\textit{Related Word Sets}\\ \midrule
cat, horse, deer, bear & amethyst & animals \\
cherry, pear, melon, banana & attack & fruits \\
shirt, dress, skirt, sweater & candy & clothes \\
\midrule\midrule
\textit{Unrelated Word Sets} \\ \midrule
candy, amethyst, belt, pond & - & - \\
gas, bee, cheese, actor & - & - \\
attack, daisy, attic, commerce & - & - \\
\midrule\midrule
\textit{Related Number Sets} \\ \midrule
100, 102, 104, 106 & 257 & arithmetic sequence with d=2 \\
100, 175, 250, 325 & 276 & arithmetic sequence with d=75 \\
100, 200, 300, 400 & 399 & arithmetic sequence with d=100 \\
\midrule\midrule
\textit{Unrelated Number Sets}\\ \midrule
399, 690, 734, 847 & - & - \\
276, 508, 661, 863 & - & - \\
257, 356, 650, 935 & - & - \\
\end{tabular}
\caption{The 22 sets used in our experiments.}
\label{table:sets}
\end{table*}

\begin{figure*}
\centering
\setlength{\tabcolsep}{2pt}
\begin{tabular}{ccc}
\includegraphics[trim={0cm 0 0 0},clip,width=0.32\linewidth]{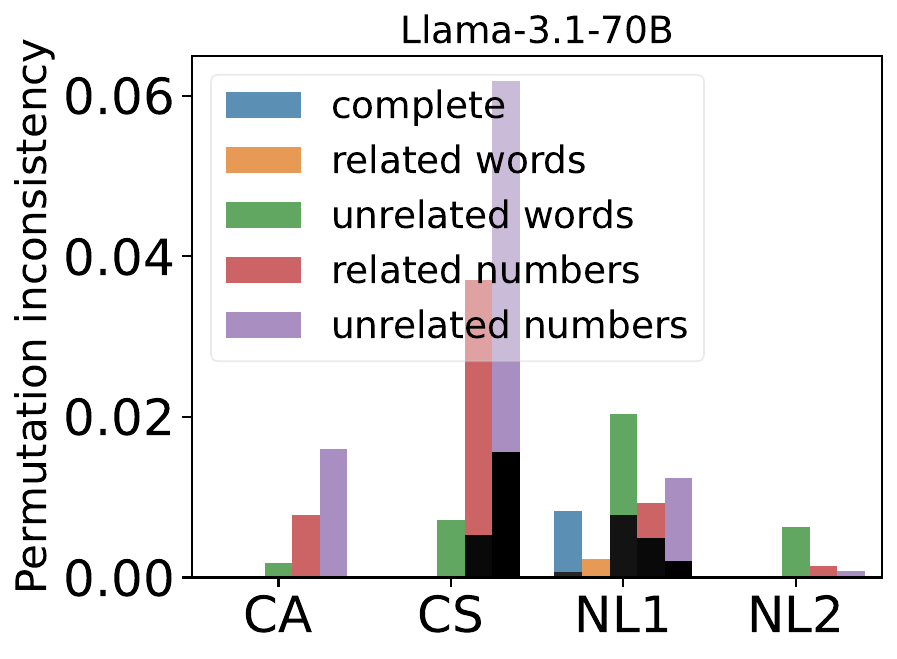} &
\includegraphics[trim={1cm 0 0 0},clip,width=0.3\linewidth]{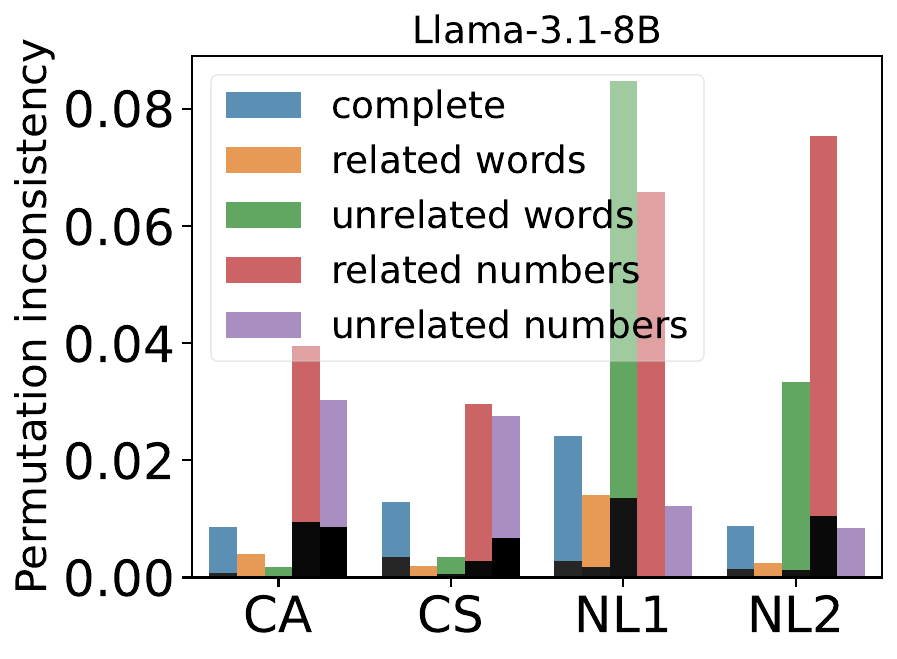} &
\includegraphics[trim={1cm 0 0 0},clip,width=0.3\linewidth]{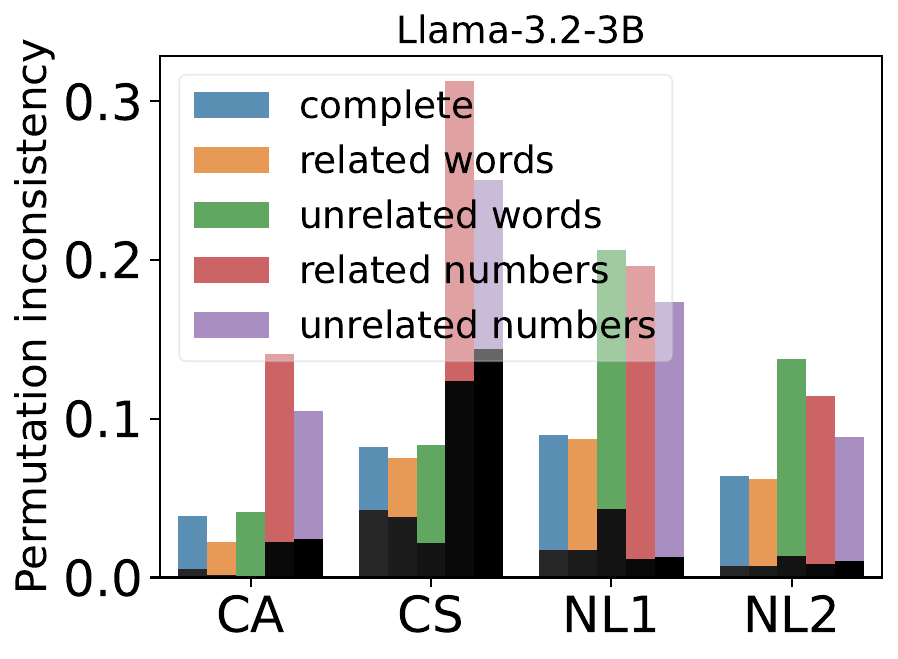} \\
\end{tabular}
\caption{The fraction of order-independent queries where the answer of the LLM is not consistently correct for all the permutations of the set, by
prompt type and set type for LLMs not included in the main article.
Values higher than 0.0 indicate the presence of sensitivity to ordering. The black region of a bar corresponds to the fraction of consistently incorrect order-independent queries.}
\label{fig:consistency_sup}
\end{figure*}

\appendix
\noindent In this Appendix, we present our experimental setup in full detail to allow for reproducibility.
We also present all the plots that were not included in the main paper.

\section{Setup}

We performed our empirical evaluations with Huggingface’s \texttt{transformers} library \cite{wolf-etal-2020-transformers}. For every query, we used the greedy search generation strategy, i.e. during generation the most likely token was selected. 

We used Nvidia H100 GPUs in our experiments. The evaluation took around 140 GPU-days to complete on all models, prompt schemas and sets with all permutations (60 963 840 queries all together).

The HuggingFace IDs of the LLMs we used are the following:

\begin{itemize}
\item meta-llama/Llama-3.2-3B-Instruct
\item meta-llama/Llama-3.1-8B-Instruct
\item mistralai/Mistral-Small-24B-Instruct-2501
\item Qwen/Qwen2.5-32B-Instruct
\item microsoft/Phi-3.5-MoE-instruct
\item meta-llama/Llama-3.1-70B-Instruct
\item meta-llama/Llama-3.3-70B-Instruct
\end{itemize}

\section{List of sets}

We list the 22 sets used in our experiments in \cref{table:sets}.
The construction of the sets was done based on human judgement, that is,
based on the judgement of the authors.
This involved consulting with LLMs as well, but not in a structured manner.

The \textbf{complete sets of words} are an exception, where 
we first came up with a pool of candidate quadruples. We then tested the candidate sets
for completion: we removed one element and asked the LLMs (the ones used in the evaluation in the main paper)
what the missing element is.
We regarded a set to be complete if all the LLMs proposed the removed element as the missing one for every query.
Queries tested removing all the elements of a given quadruple and presenting the remaining three elements in every possible order.

An example prompt to test what element the LLM considers to be missing from the set would be of the form
\prompt{
Answer with the most likely missing element from the below set:\\North, East, South
}
In this case, we require the response of all the investigated LLMs to be \texttt{West}.
A single incorrect answer discarded the set from the candidate quadruples.

\section{Python prompt schemes in CS and CA}

The features defined for the CS and CA schemas are the following:
\begin{itemize}
\item \emph{init}: initialization of the set
    \begin{itemize}
    \item shorthand
    \item constructor with list
    \item constructor with tuple
    \item constructor with string and split
    \end{itemize}
\end{itemize}

Examples for the different init features:
\prompt{
S = {"North", "South", "West", "East"}\\
S = set(["North", "South", "West", "East"])\\
S = set(("North", "South", "West", "East")\\
S = set("North South West East".split())
}

\begin{itemize}
\item \emph{variable}: \texttt{S $|$ my\_set} (variable name for the set)
\item \emph{quotation}: \texttt{' $|$ "} (quotation mark for strings)
\item \emph{tab}: how the code was tabulated
    \begin{itemize}
    \item 2 spaces
    \item 4 spaces
    \item tab
    \end{itemize}
\item \emph{result}: representation of the result
    \begin{itemize}
    \item stored in a variable
    \item printed out
    \end{itemize}
\end{itemize}

Example CS prompts for result features. Note that the given question changes accordingly.

\prompt{
Answer with a single word. Given the Python code below, what will be the value of the `result` variable?\\
\\
```python\\
S = {"North", "South", "West", "East"}\\
result = "North" in S\\
```}

\prompt{
Answer with a single word. Given the Python code below, what will be the output?\\
\\
```python\\
S = {"North", "South", "West", "East"}\\
print("North" in S)\\
```}

\begin{itemize}
\item \emph{logic}: structure of the code
    \begin{itemize}
    \item \textit{main}: all operations are executed in the global scope
    \item \textit{function}: the set containment operation is defined in a function
    \item \textit{function with typehints}: the function has typehints in the function header
    \item \textit{function with docstring}: the function has a docstring
    \item \textit{function with typehints and docstring}
    \end{itemize}
\item \emph{argument order}: order of arguments if a function is present
    \begin{itemize}
    \item set-first
    \item element-first
    \end{itemize}
\end{itemize}

Example CS prompts for logic features, i.e. everything is executed in the main scope or a function was defined with both typehinting and docstring.

\prompt{
Answer with a single word. Given the Python code below, what will be the value of the `result` variable?\\
\\
```python\\
S = {"North", "South", "West", "East"}\\
result = "North" in S\\
```}

\prompt{
Answer with a single word. Given the Python code below, what will be the value of the `result` variable?\\
\\
```python\\
from typing import Any, Set\\
\\
def contains(s: Set[Any], item: Any) -> bool:\\
\phantom{xxxx}"""\\
\phantom{xxxx}Checks if an item is in a set.\\
\\
\phantom{xxxx}Args:\\
\phantom{xxxxxxxx}s (set): The set to check in.\\
\phantom{xxxxxxxx}item (any): The item to check for.\\
\phantom{xxxx}Returns:\\
\phantom{xxxxxxxx}bool: True if the item is in the set, False otherwise.\\
\phantom{xxxx}"""\\
\phantom{xxxx}return item in s\\
\\
S = {"North", "South", "West", "East"}\\
result = contains(S, "North")\\
```}

\section{Additional plots}

\Cref{fig:consistency_sup} shows the permutation inconsistency of the remaining LLMs (those that were left out from the main paper). 
Here, again, we see that the LLMs show a significant inconsistency to permutation, as well as among each other.
Llama-3.2-3B and 3.1-70B, for example, are surprisingly sensitive to the order of number-sets in the CS prompt type, while the 8B version
is not.

\begin{figure*}[tb]
\centering
\includegraphics[width=\textwidth]{settype-error/settype-errors-nl1}
\includegraphics[trim={0 0 0 0.8cm},clip,width=\textwidth]{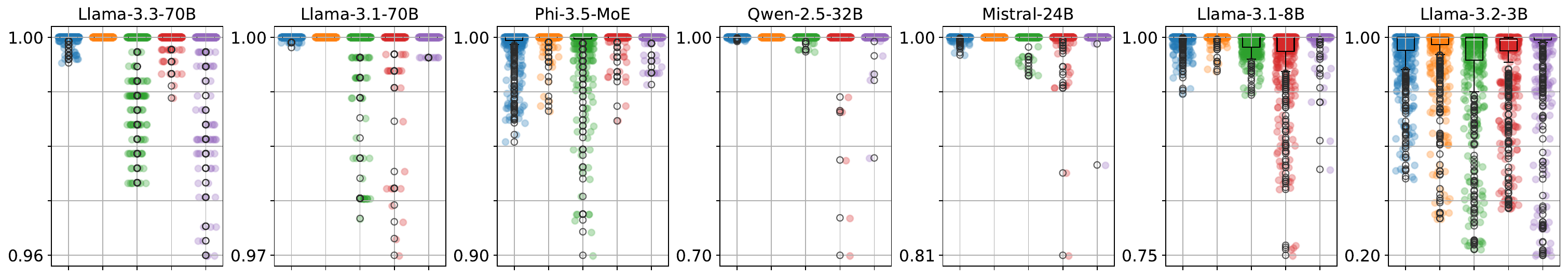}
\includegraphics[trim={0 0 0 0.8cm},clip,width=\textwidth]{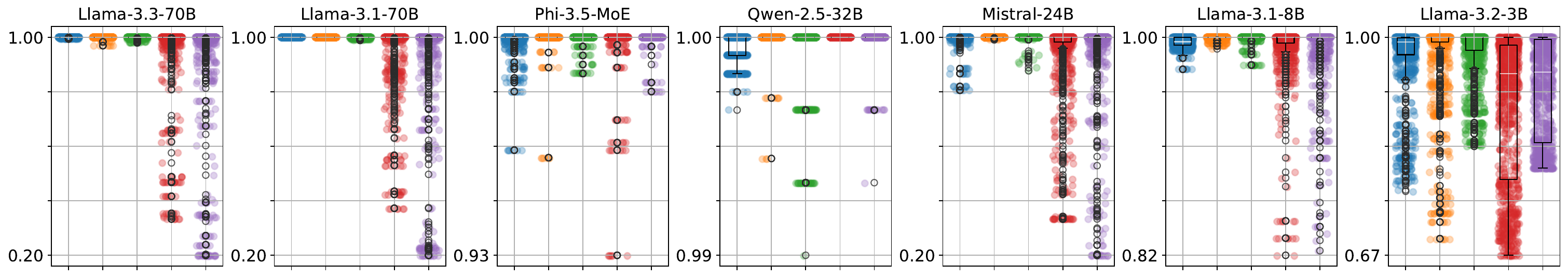}
\includegraphics[trim={0 0 0 0.8cm},clip,width=\textwidth]{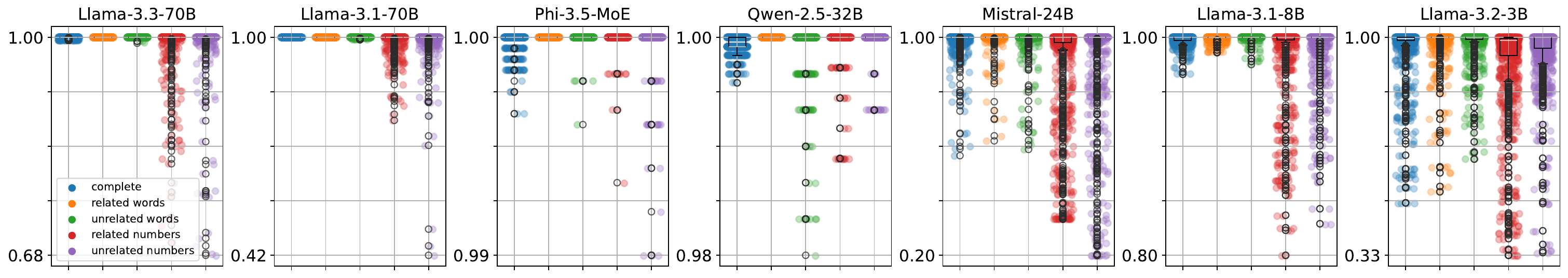}
\caption{The average accuracy of queries with different set-types, for all the LLMs, in the NL1, NL2, CS and CA prompt categories (rows from top to bottom).
Every point belongs to a fixed prompt template and represents the average of the queries that belong to the given set type.
For example, for the complete set type, every point is the average accuracy of $10\cdot (24\cdot 4+6\cdot 4+6\cdot 4)$ prompts.}
\label{fig:supp_settype}
\end{figure*}

\begin{figure*}
\centering
\setlength{\tabcolsep}{-2pt}
\begin{tabular}{ccccccc}
\scriptsize Llama3.3-70B &
\scriptsize Llama3.1-70B & 
\scriptsize Phi-3.5-MoE &
\scriptsize Qwen2.5-32B &
\scriptsize Mistral-24B &
\scriptsize Llama-3.1-8B &
\scriptsize Llama-3.2-3B \\
\includegraphics[width=0.15\textwidth]{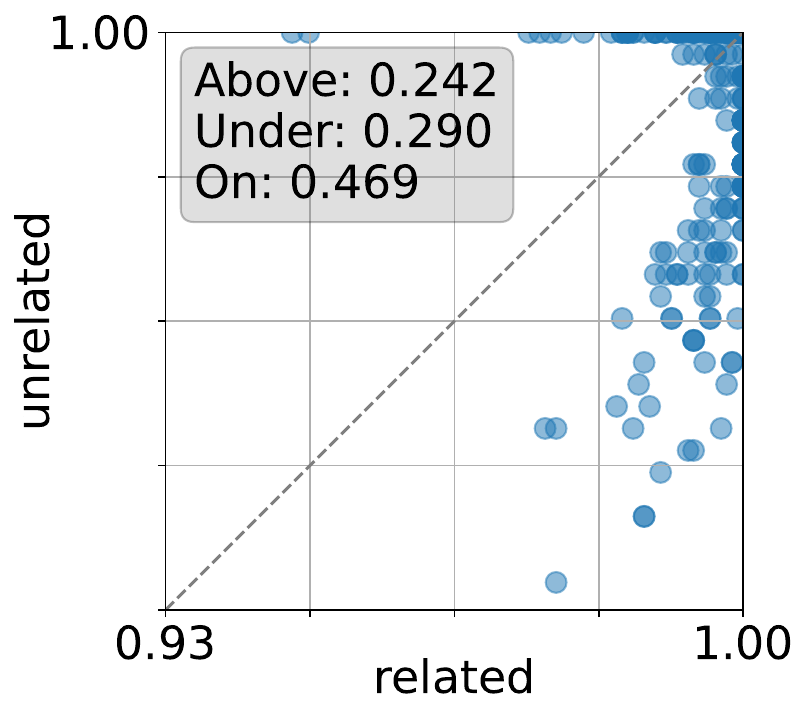} &
\includegraphics[width=0.15\textwidth]{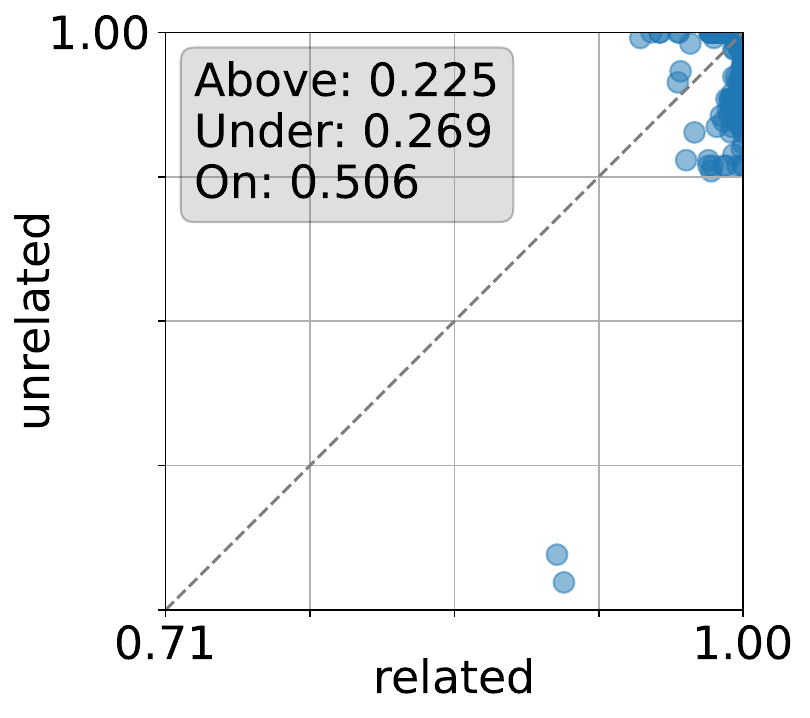} &
\includegraphics[width=0.15\textwidth]{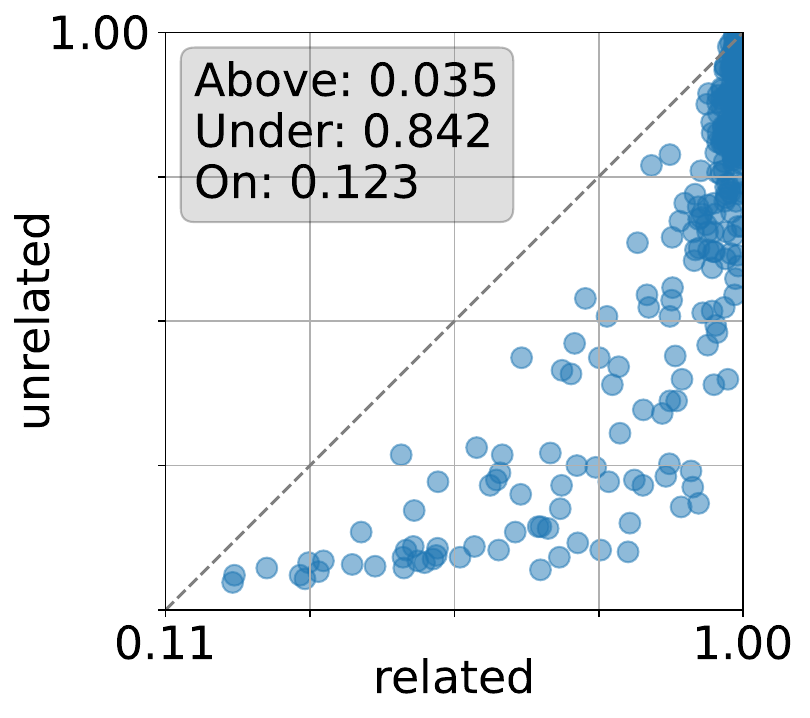} &
\includegraphics[width=0.15\textwidth]{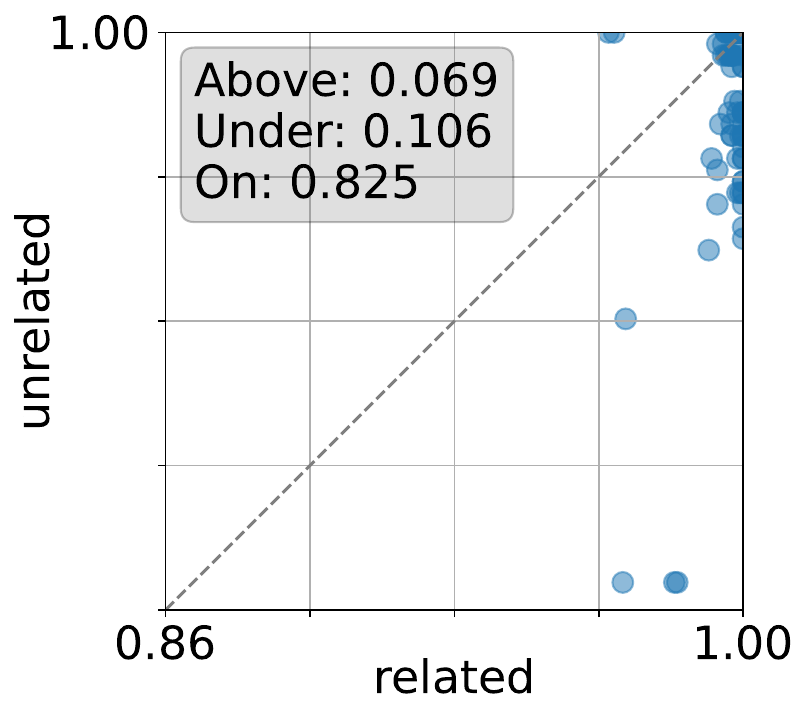} &
\includegraphics[width=0.15\textwidth]{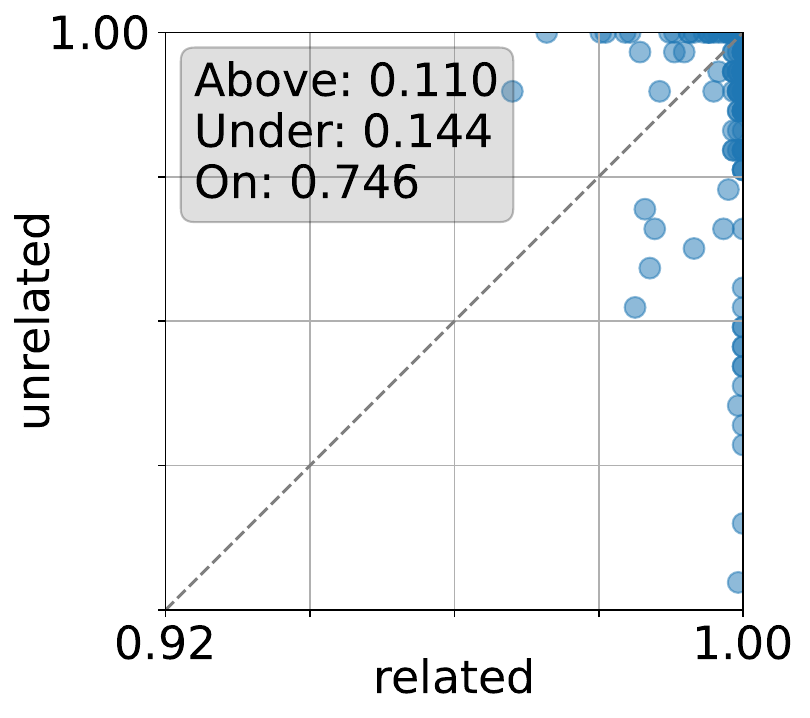} &
\includegraphics[width=0.15\textwidth]{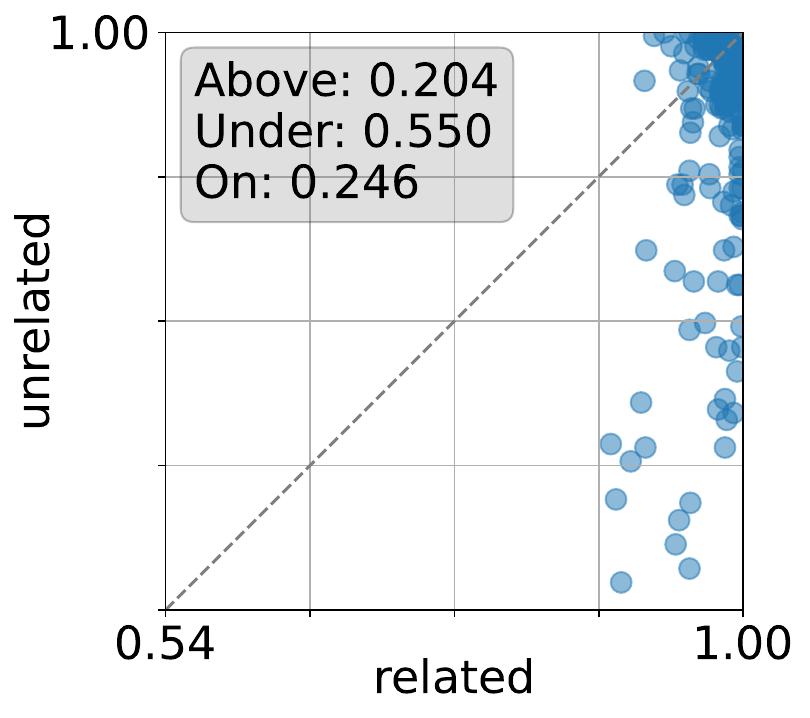} &
\includegraphics[width=0.15\textwidth]{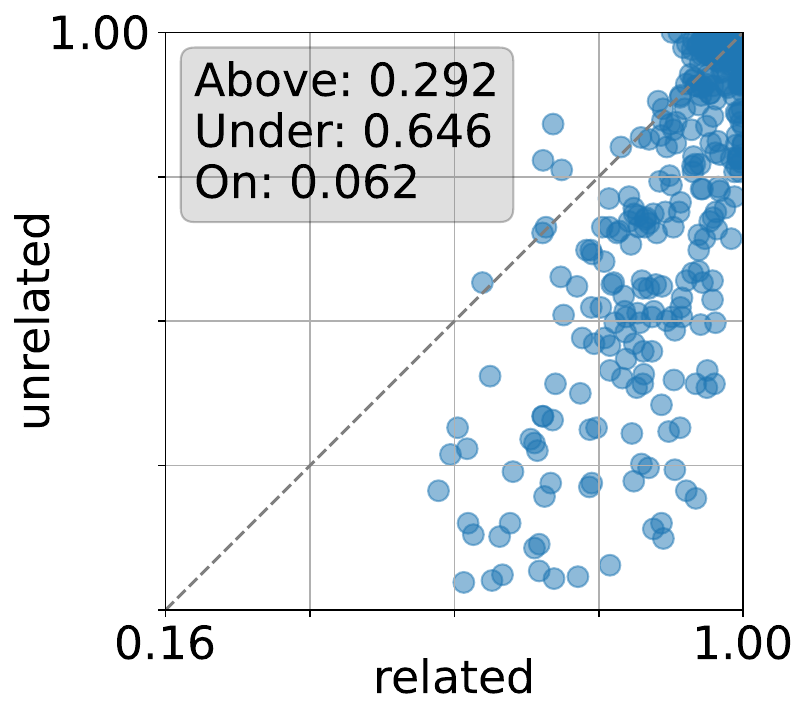} \\
\includegraphics[width=0.15\textwidth]{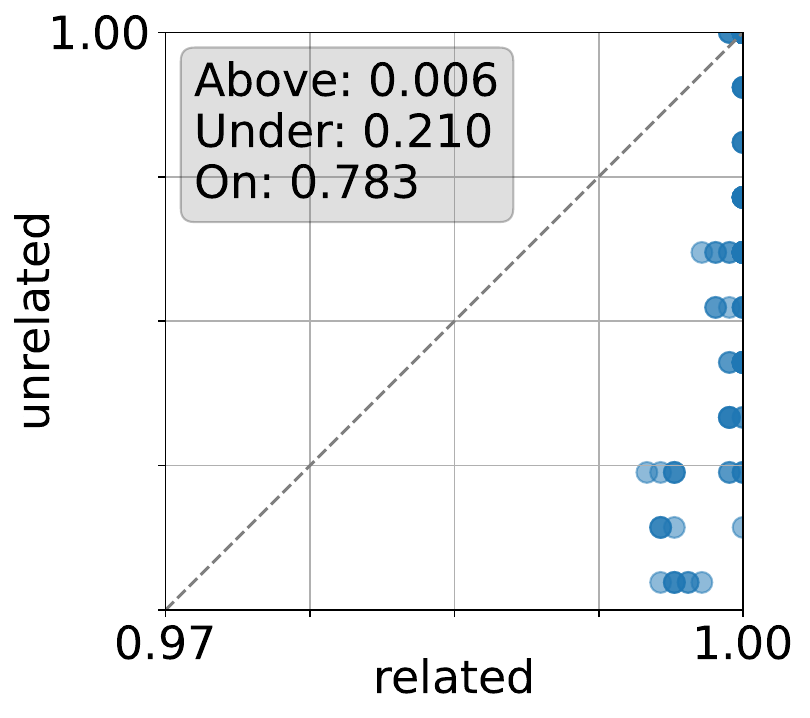} &
\includegraphics[width=0.15\textwidth]{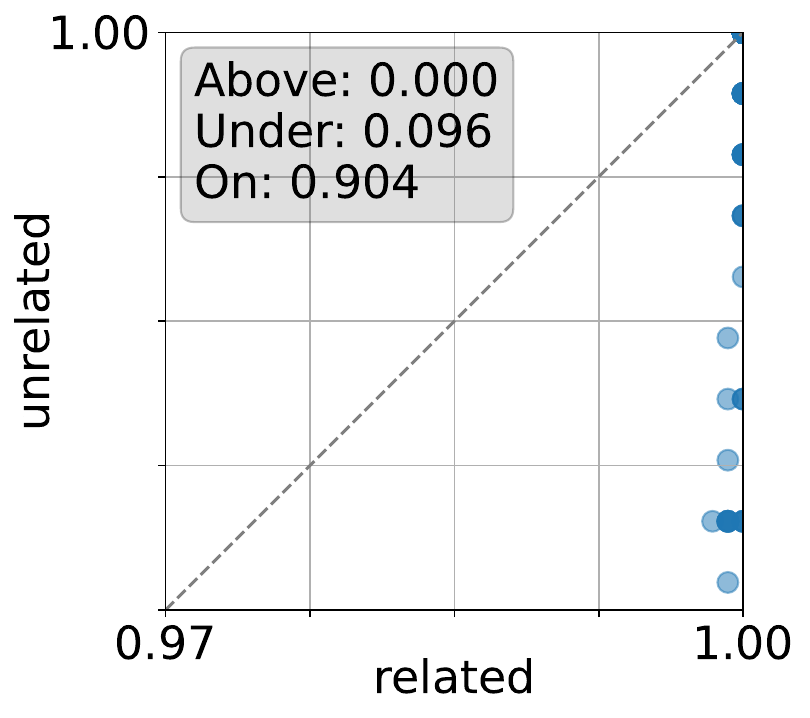} &
\includegraphics[width=0.15\textwidth]{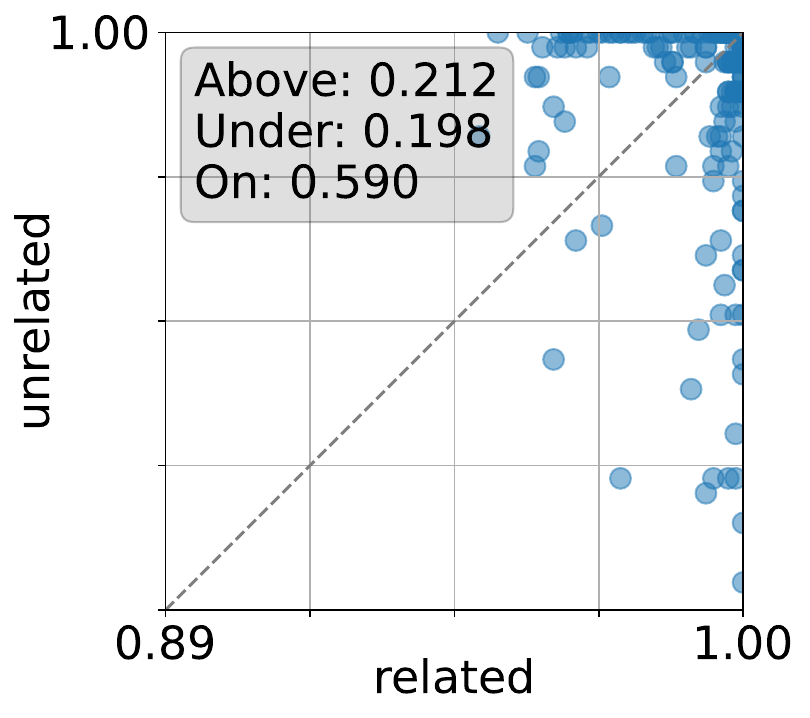} &
\includegraphics[width=0.15\textwidth]{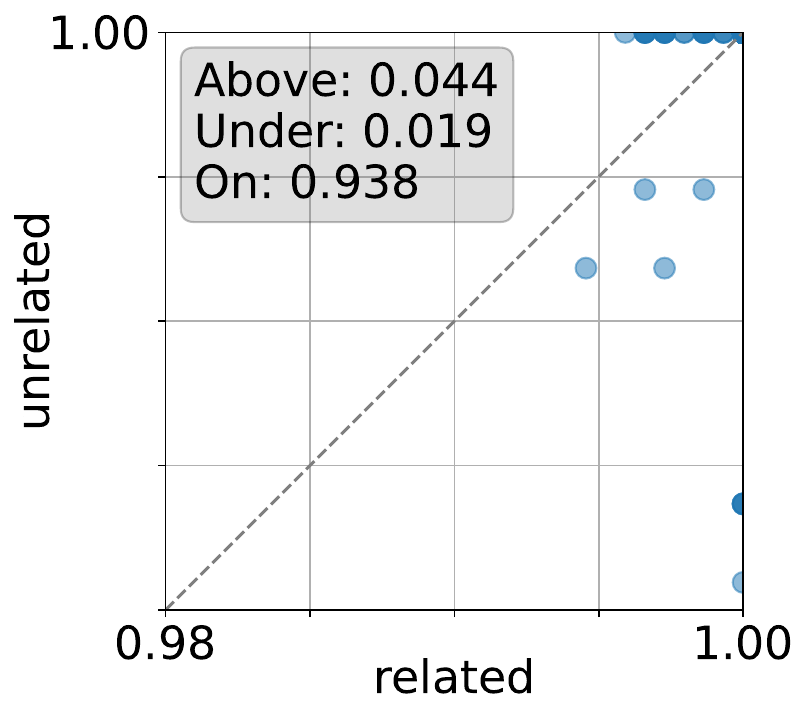} &
\includegraphics[width=0.15\textwidth]{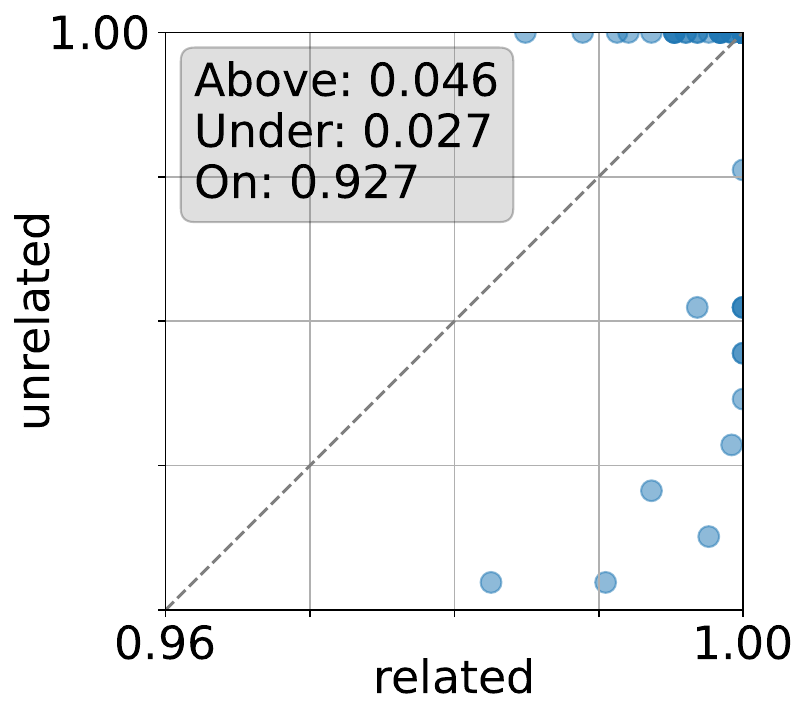} &
\includegraphics[width=0.15\textwidth]{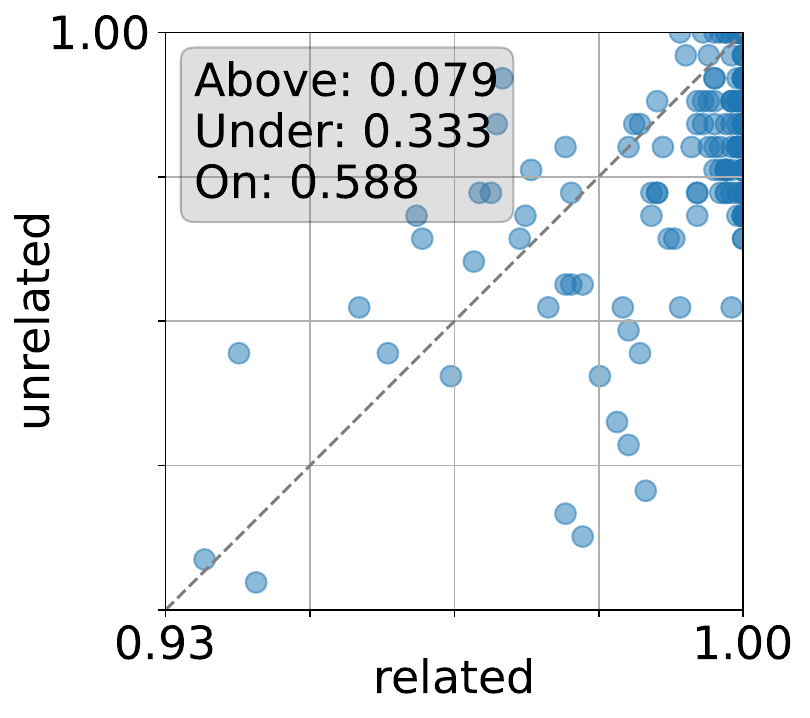} &
\includegraphics[width=0.15\textwidth]{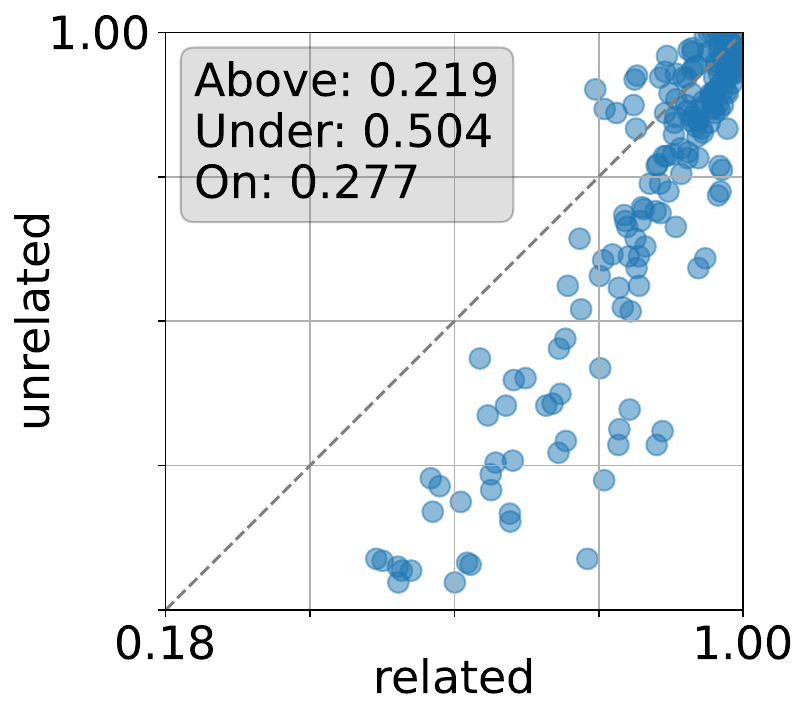} \\
\includegraphics[width=0.15\textwidth]{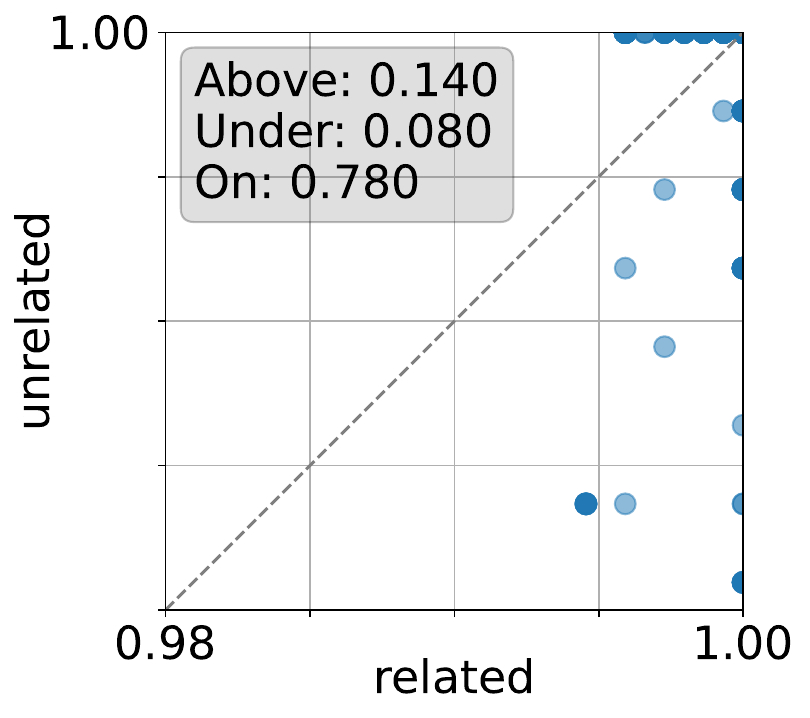} &
\includegraphics[width=0.15\textwidth]{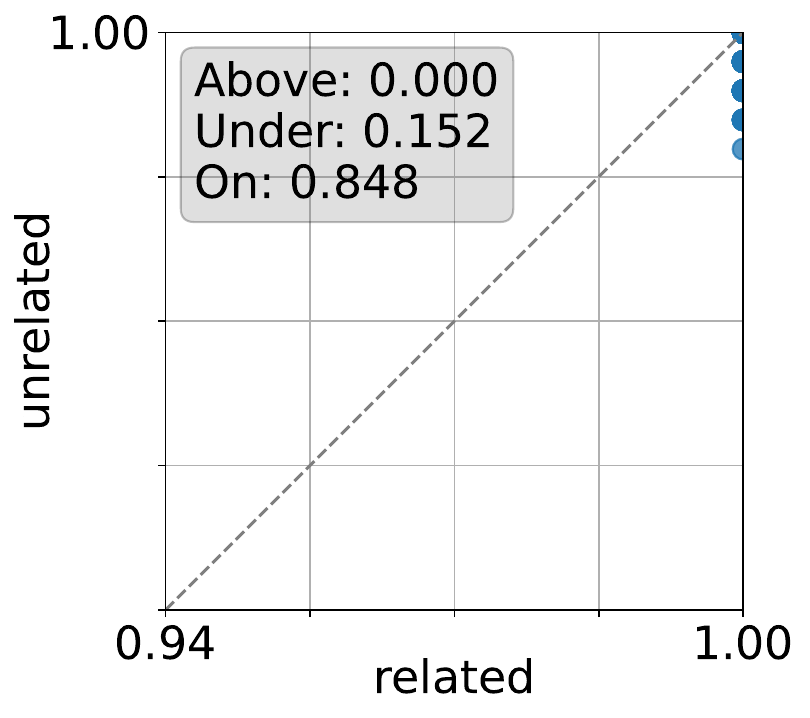} &
\includegraphics[width=0.15\textwidth]{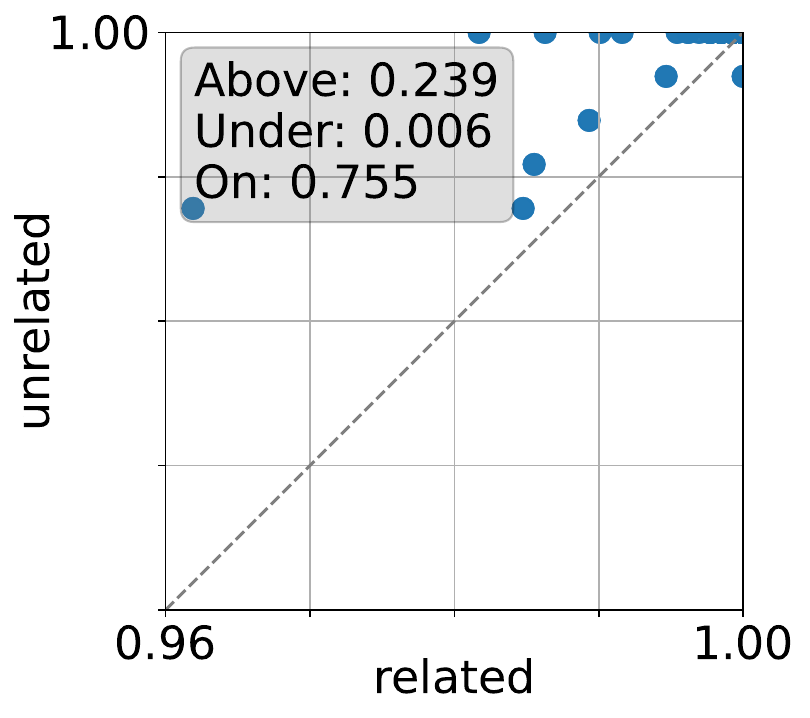} &
\includegraphics[width=0.15\textwidth]{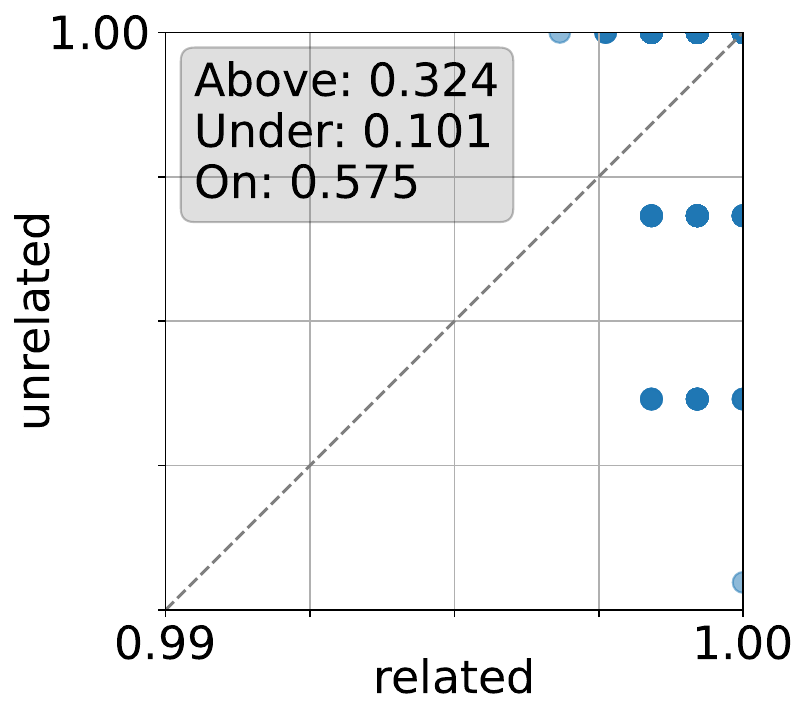} &
\includegraphics[width=0.15\textwidth]{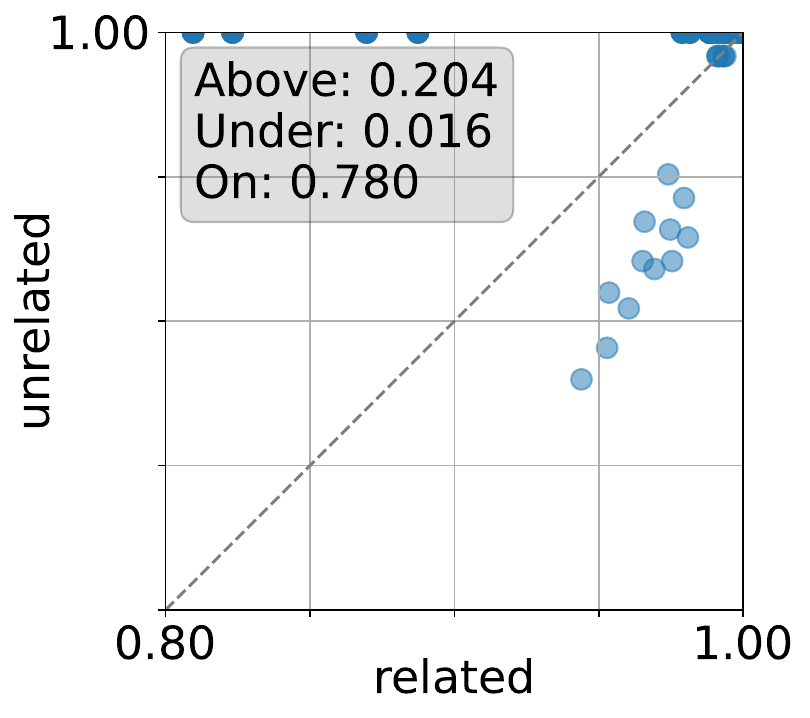} &
\includegraphics[width=0.15\textwidth]{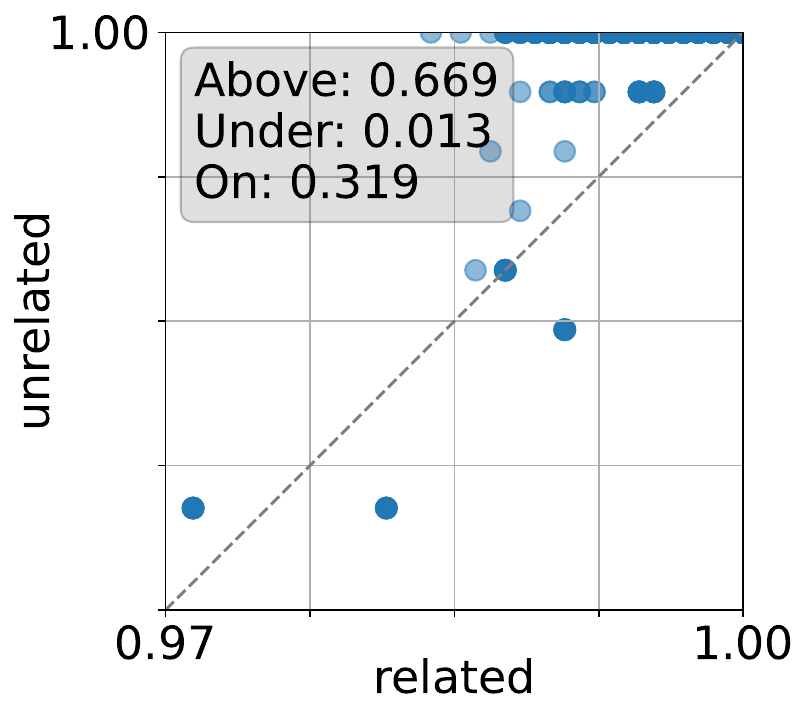} &
\includegraphics[width=0.15\textwidth]{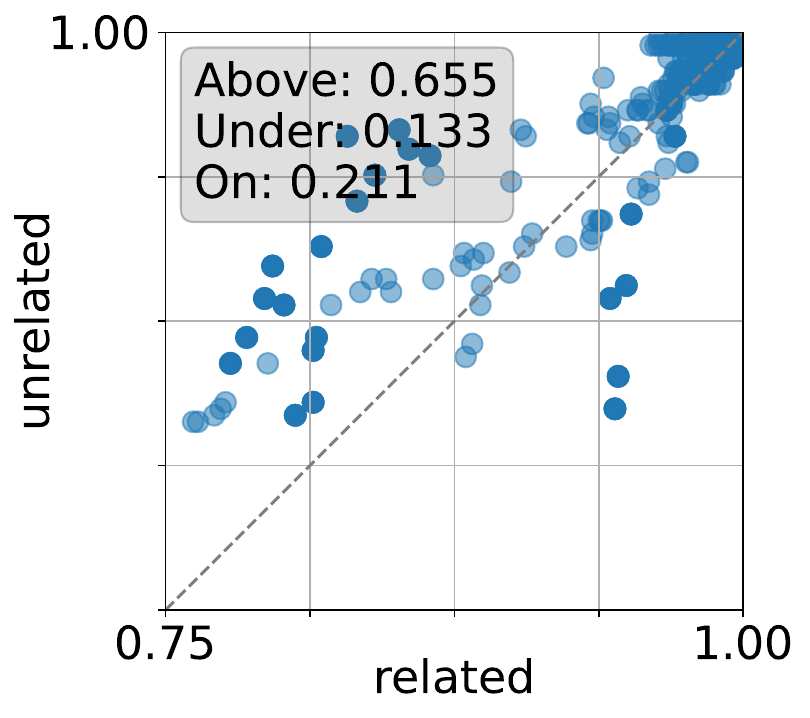} \\
\includegraphics[width=0.15\textwidth]{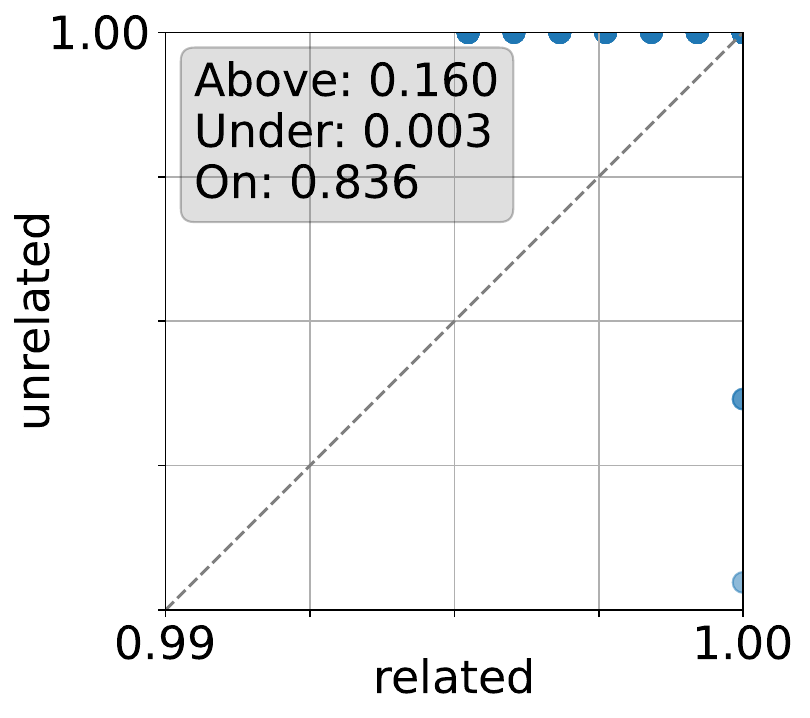} &
\includegraphics[width=0.15\textwidth]{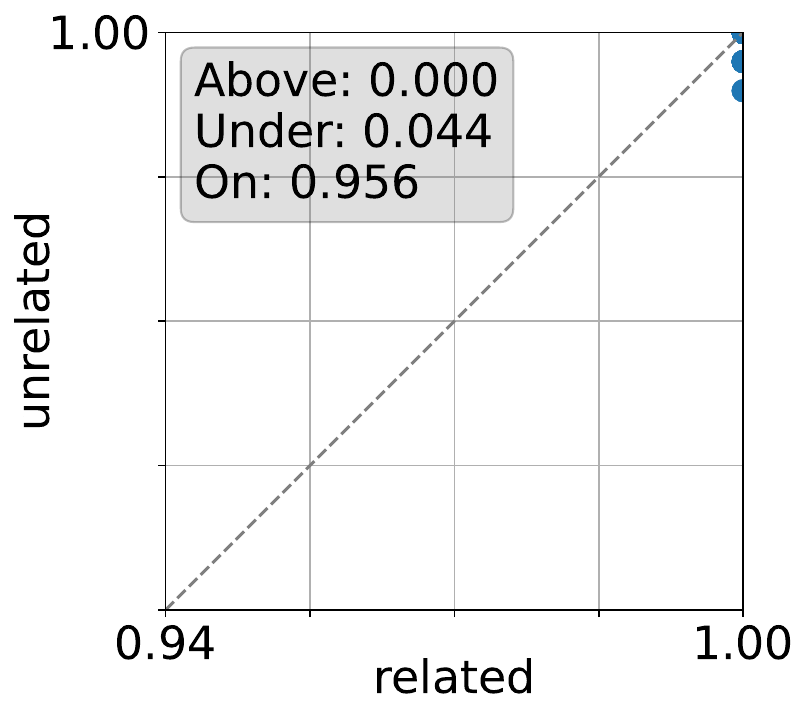} &
\includegraphics[width=0.15\textwidth]{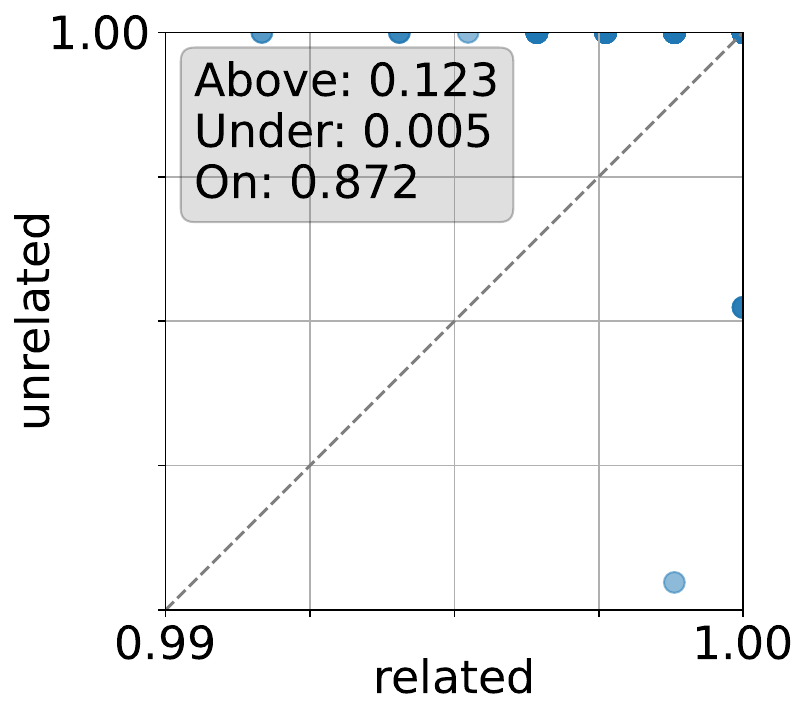} &
\includegraphics[width=0.15\textwidth]{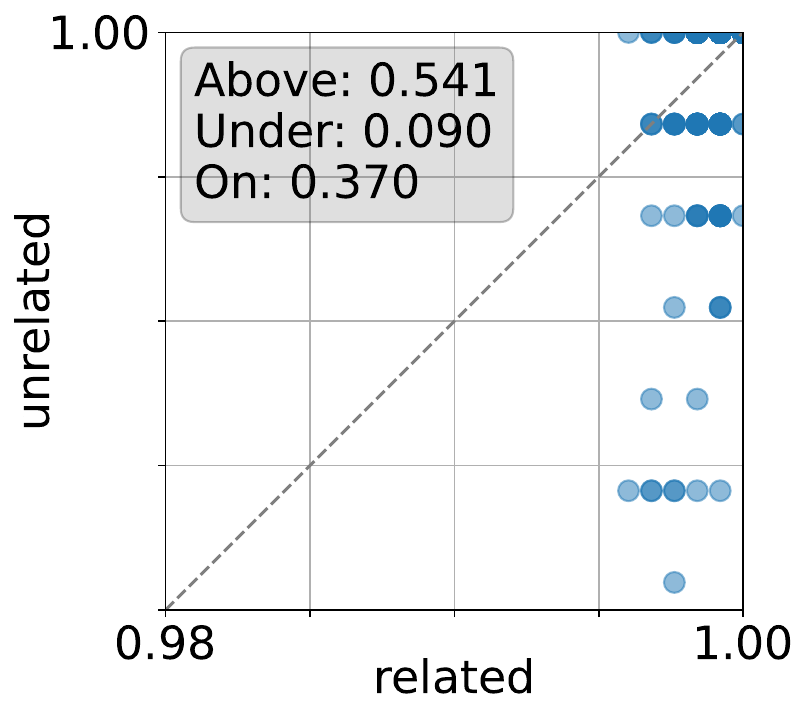} &
\includegraphics[width=0.15\textwidth]{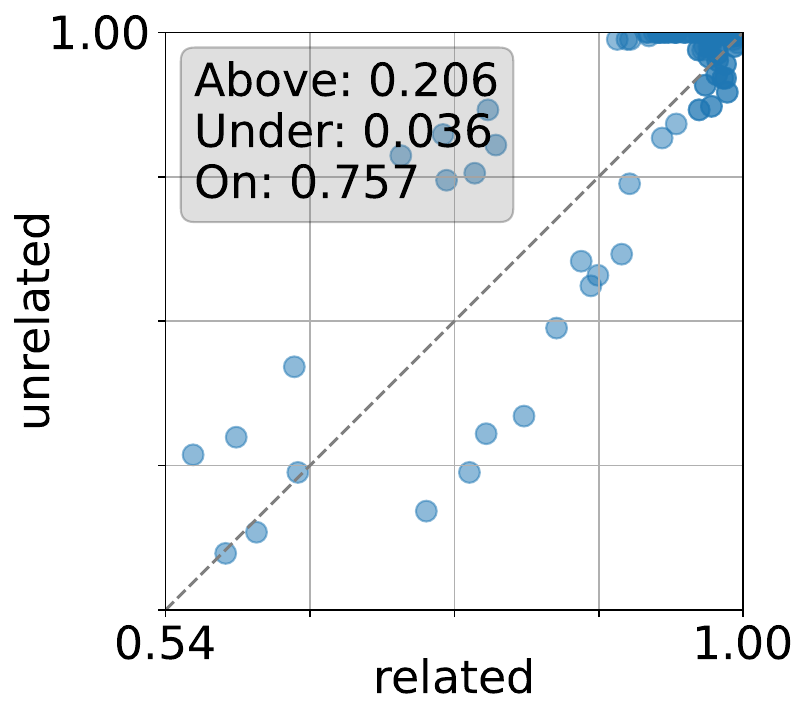} &
\includegraphics[width=0.15\textwidth]{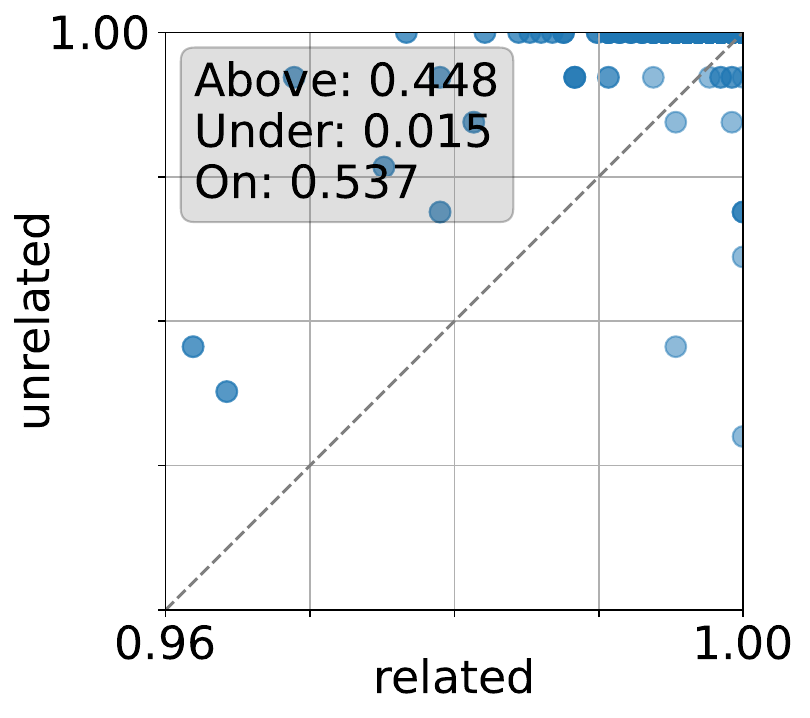} &
\includegraphics[width=0.15\textwidth]{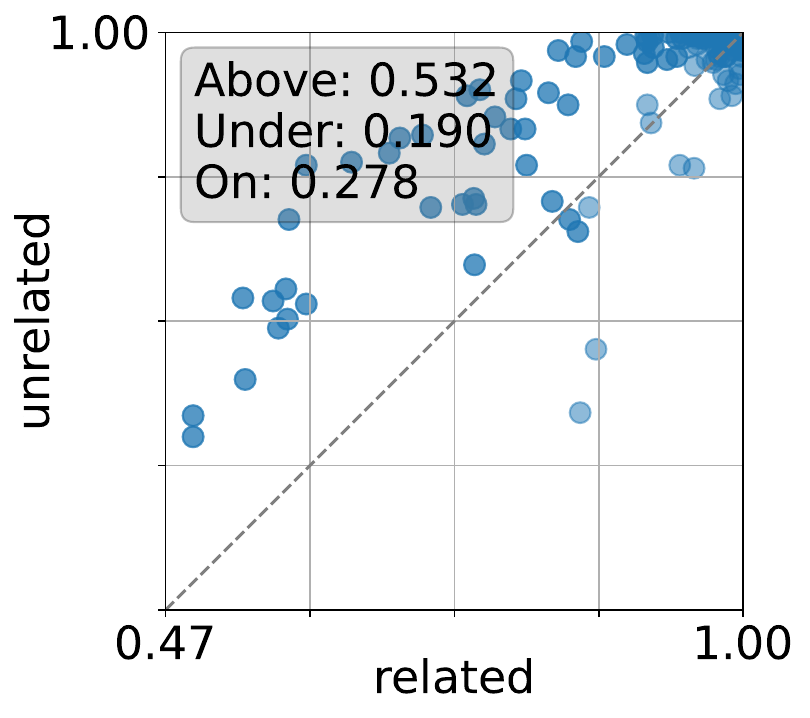} \\
\end{tabular}
\caption{Scatter plots illustrating the relationship of \textbf{unrelated word sets and complete sets} in the NL1, NL2, CS and CA prompt categories (rows from top to bottom).
A point belongs to the same prompt template, the two coordinates are average accuracies over queries with related and unrelated
sets of the given type.}
\label{fig:supp_scatter_complete}
\end{figure*}

\begin{figure*}
\centering
\setlength{\tabcolsep}{-2pt}
\begin{tabular}{ccccccc}
\scriptsize Llama3.3-70B &
\scriptsize Llama3.1-70B & 
\scriptsize Phi-3.5-MoE &
\scriptsize Qwen2.5-32B &
\scriptsize Mistral-24B &
\scriptsize Llama-3.1-8B &
\scriptsize Llama-3.2-3B \\
\includegraphics[width=0.15\textwidth]{scatter/NL1_llama_70b_3.3_related_accuracy} &
\includegraphics[width=0.15\textwidth]{scatter/NL1_llama_70b_3.1_related_accuracy} &
\includegraphics[width=0.15\textwidth]{scatter/NL1_phi_moe_related_accuracy} &
\includegraphics[width=0.15\textwidth]{scatter/NL1_qwen_32b_related_accuracy} &
\includegraphics[width=0.15\textwidth]{scatter/NL1_mistral_24b_related_accuracy} &
\includegraphics[width=0.15\textwidth]{scatter/NL1_llama_8b_related_accuracy} &
\includegraphics[width=0.15\textwidth]{scatter/NL1_llama_3b_related_accuracy} \\
\includegraphics[width=0.15\textwidth]{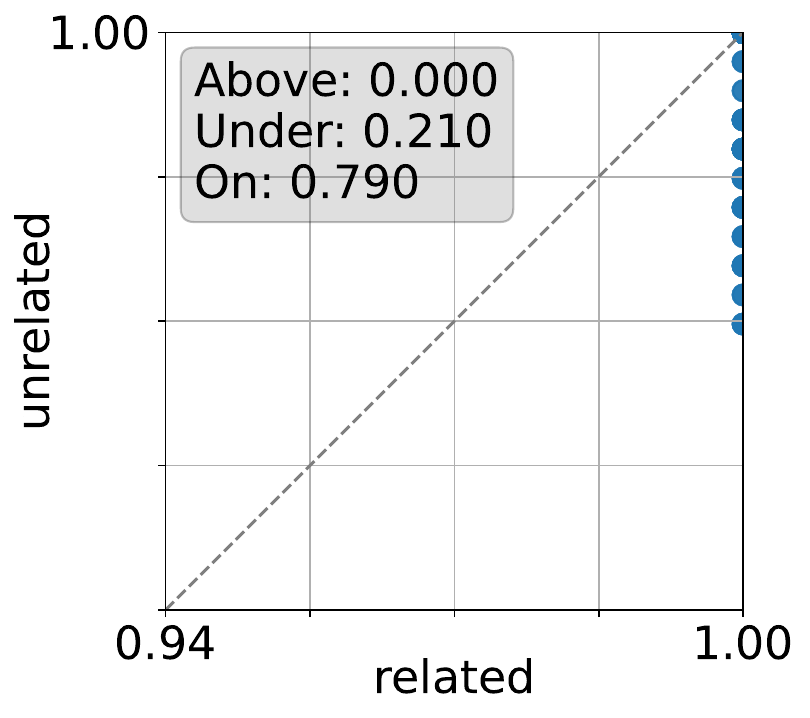} &
\includegraphics[width=0.15\textwidth]{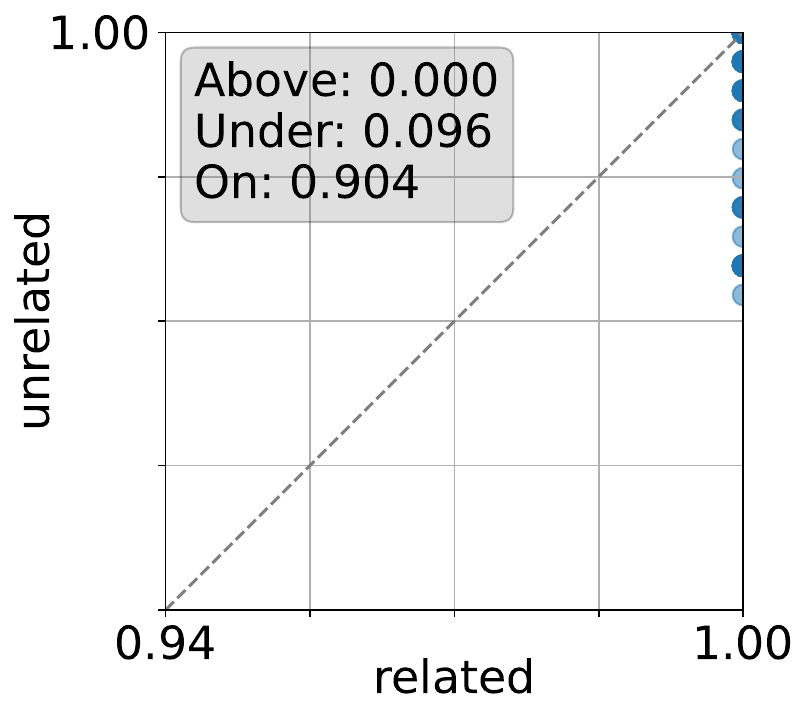} &
\includegraphics[width=0.15\textwidth]{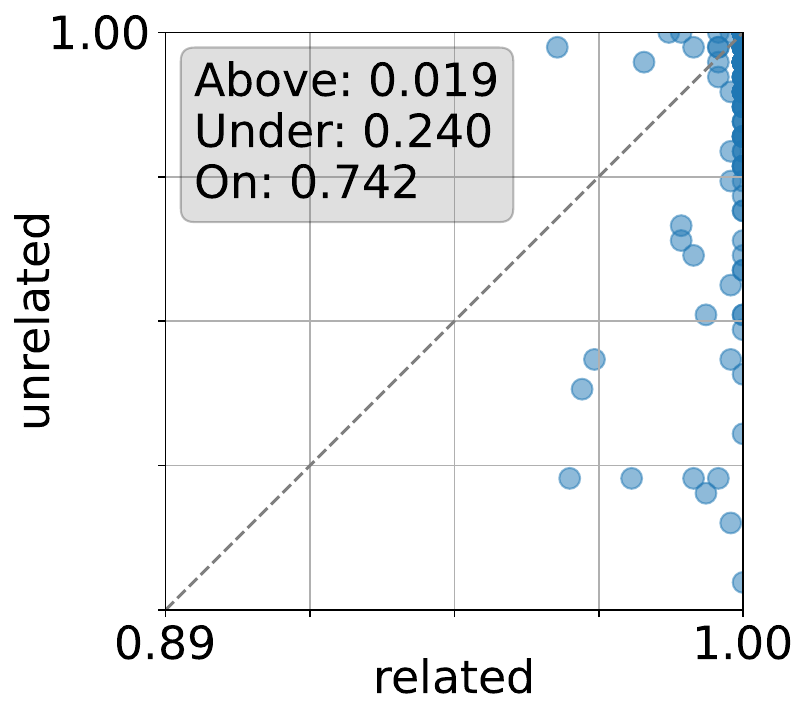} &
\includegraphics[width=0.15\textwidth]{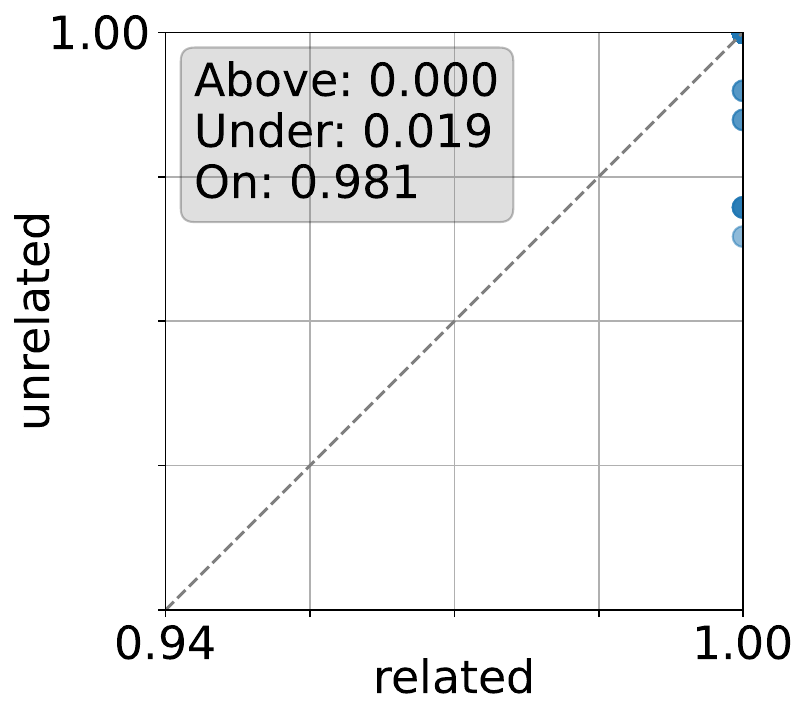} &
\includegraphics[width=0.15\textwidth]{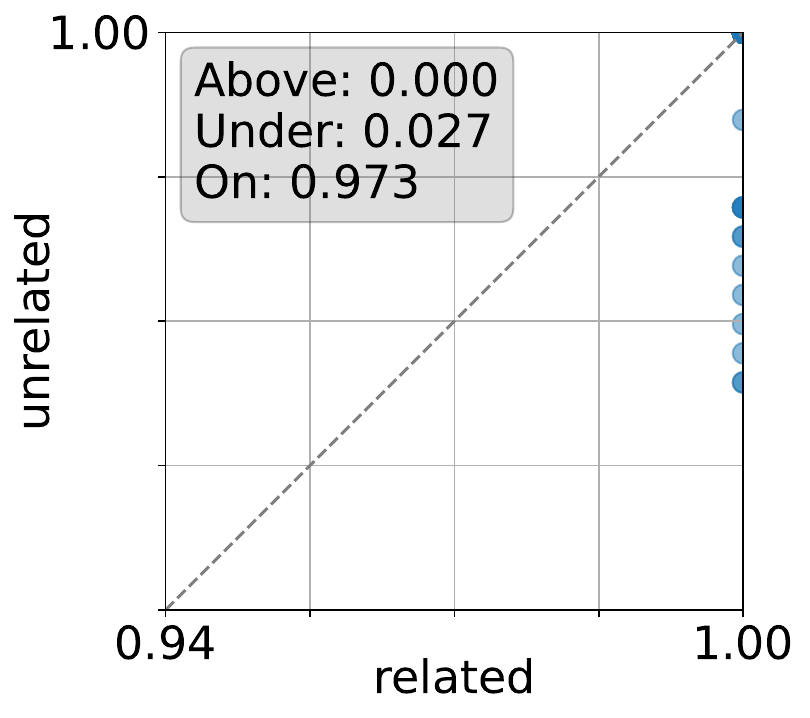} &
\includegraphics[width=0.15\textwidth]{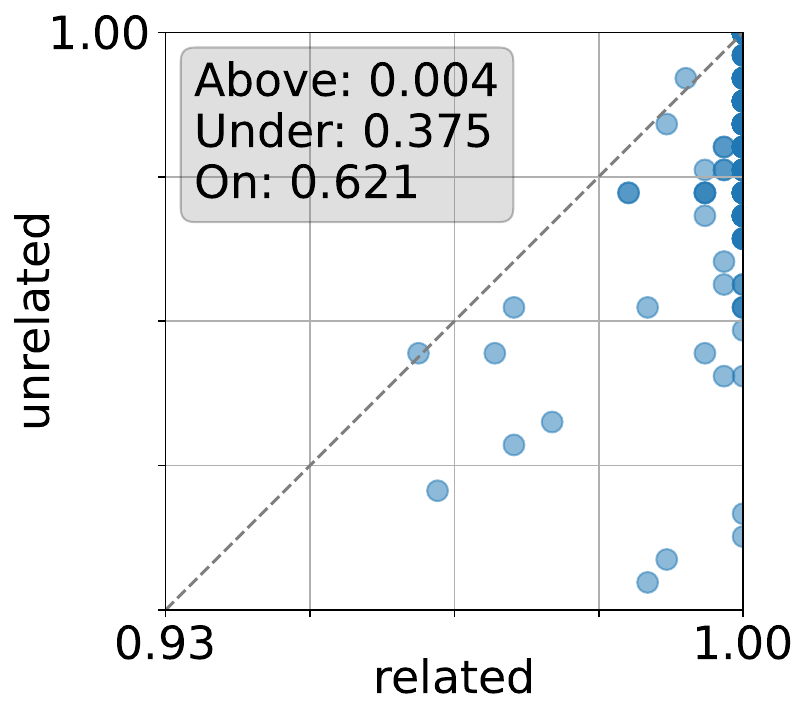} &
\includegraphics[width=0.15\textwidth]{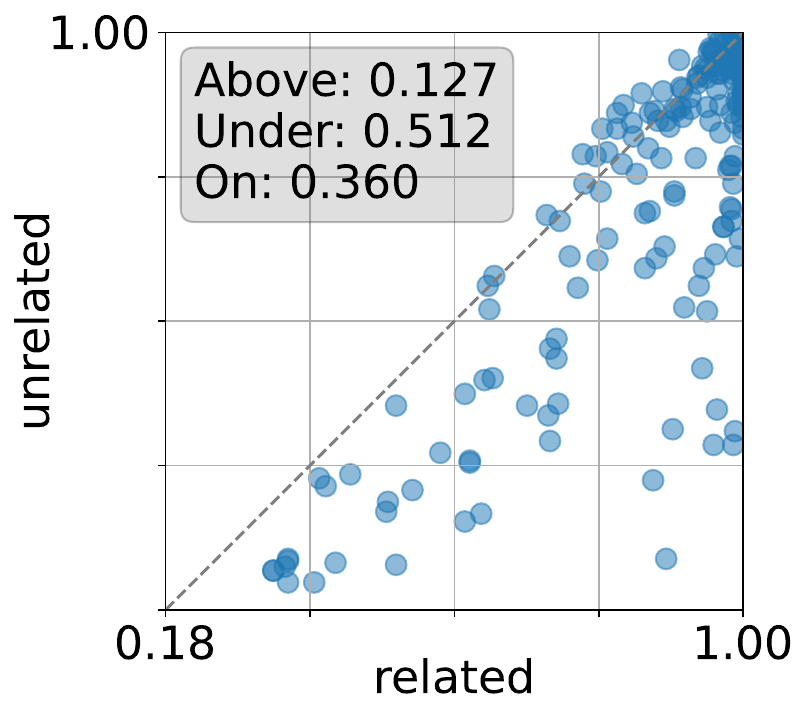} \\
\includegraphics[width=0.15\textwidth]{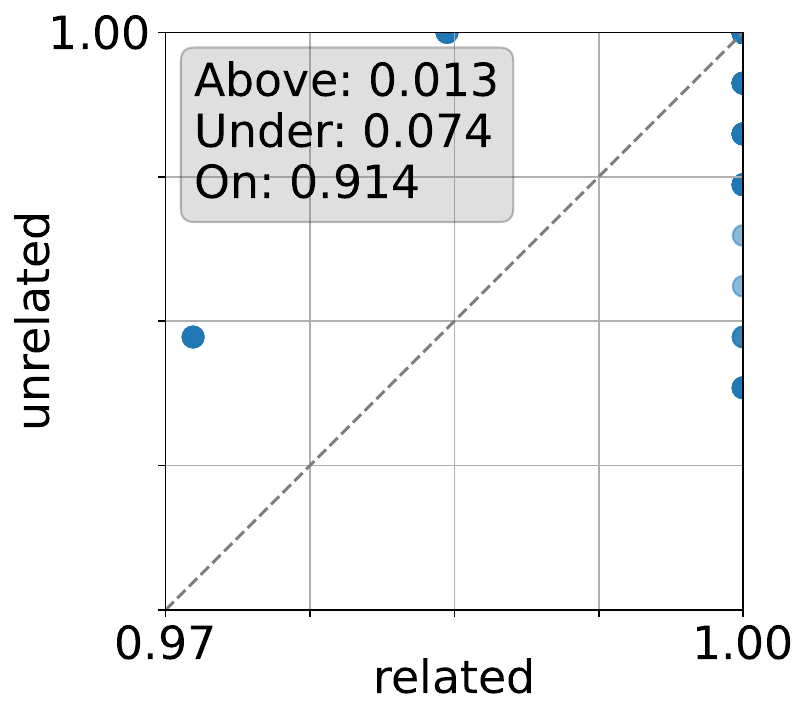} &
\includegraphics[width=0.15\textwidth]{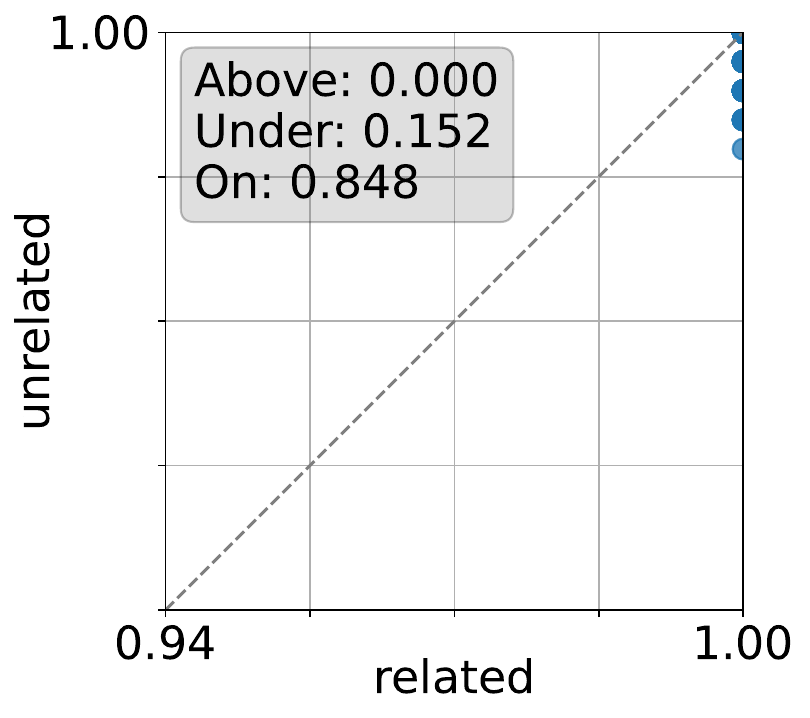} &
\includegraphics[width=0.15\textwidth]{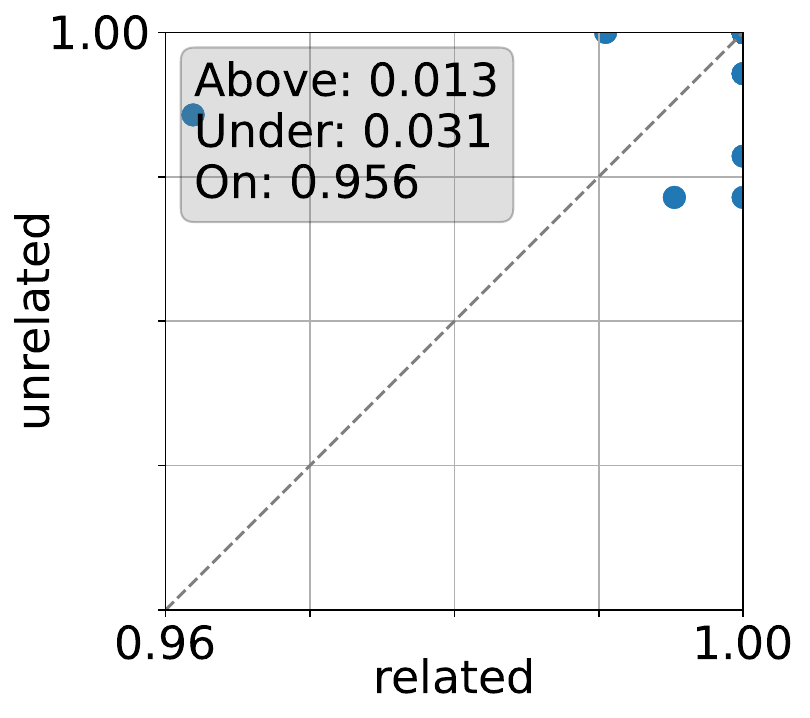} &
\includegraphics[width=0.15\textwidth]{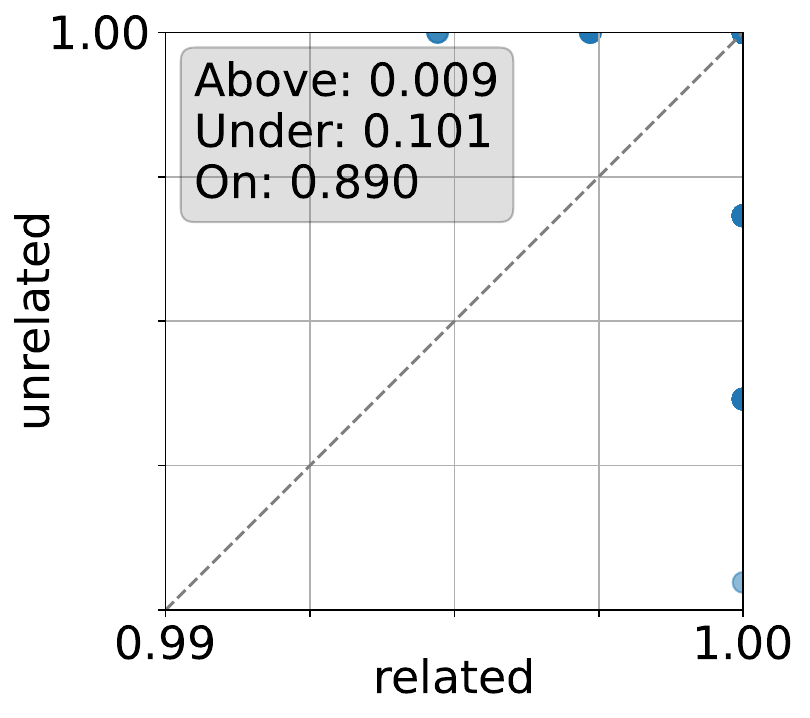} &
\includegraphics[width=0.15\textwidth]{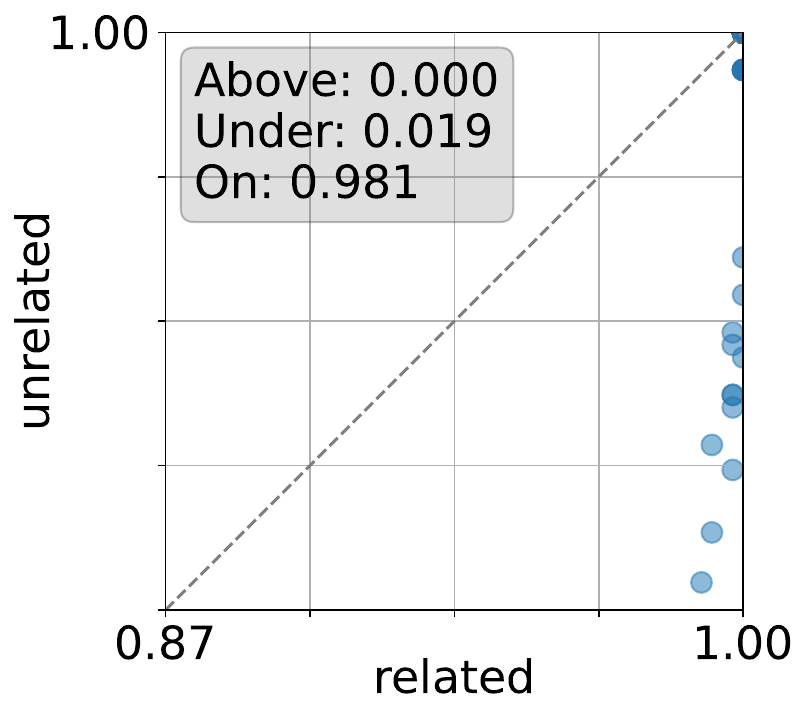} &
\includegraphics[width=0.15\textwidth]{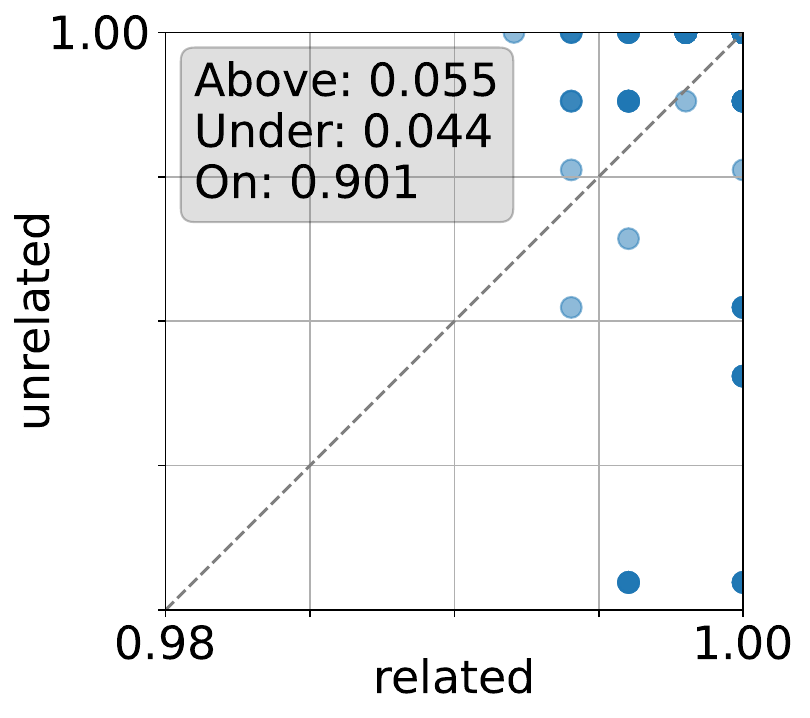} &
\includegraphics[width=0.15\textwidth]{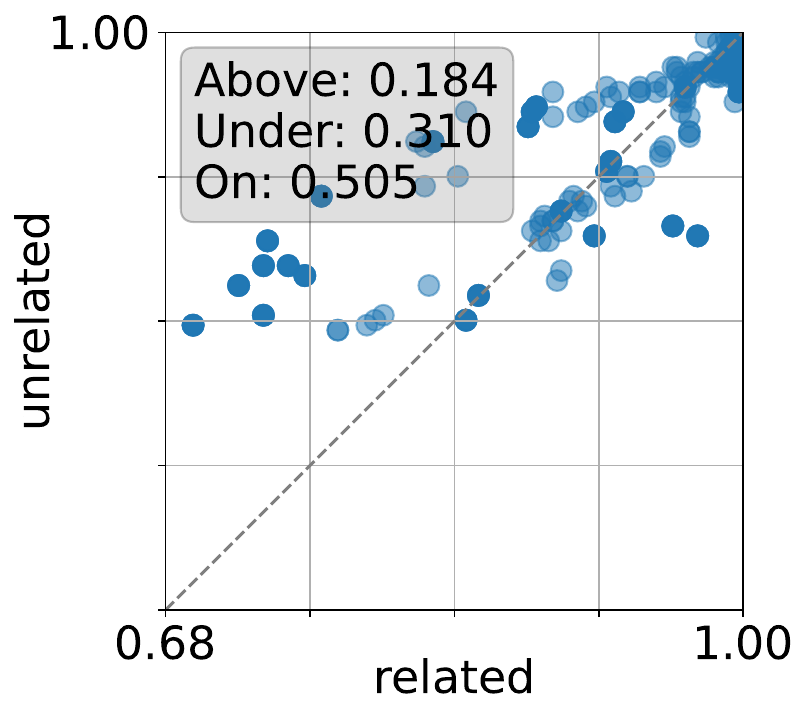} \\
\includegraphics[width=0.15\textwidth]{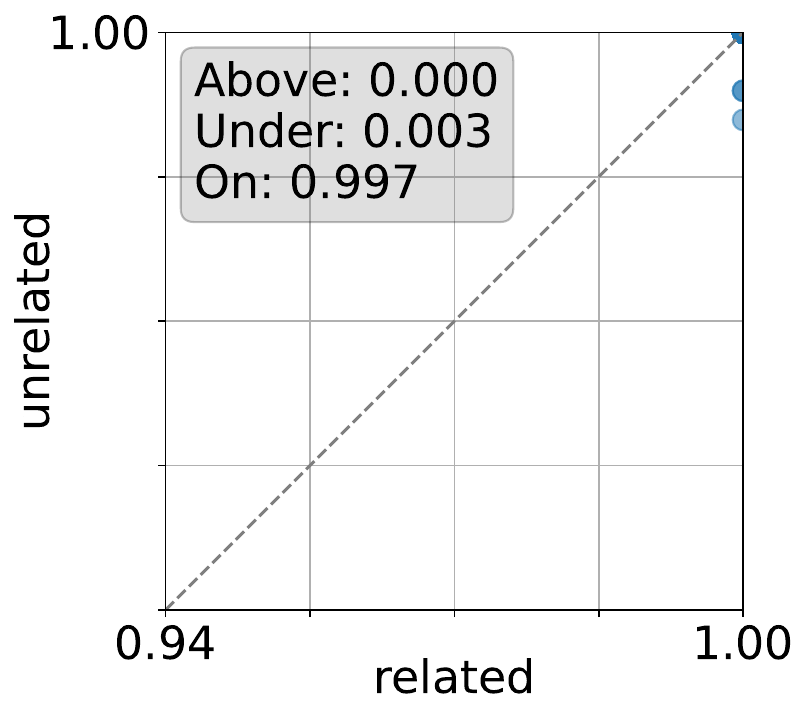} &
\includegraphics[width=0.15\textwidth]{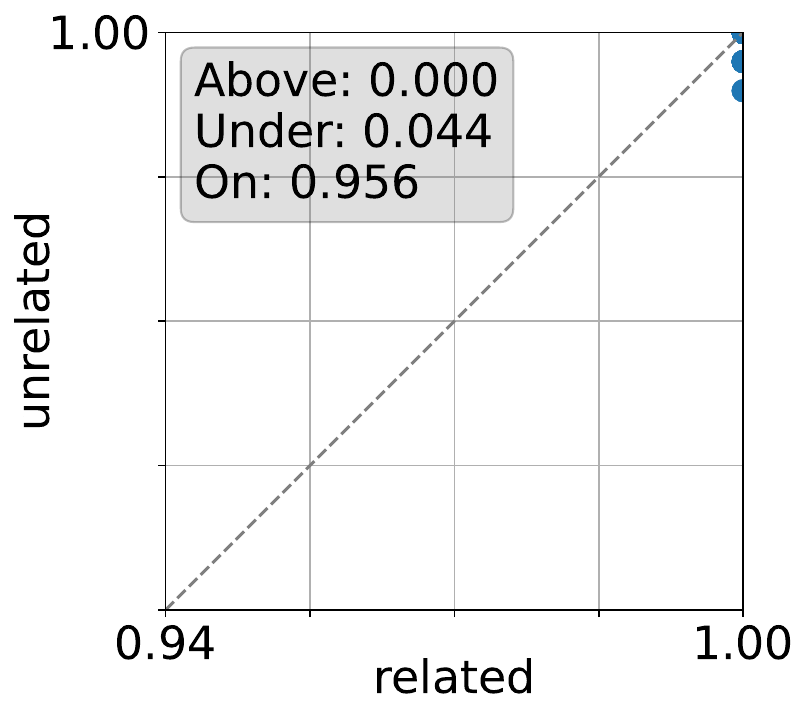} &
\includegraphics[width=0.15\textwidth]{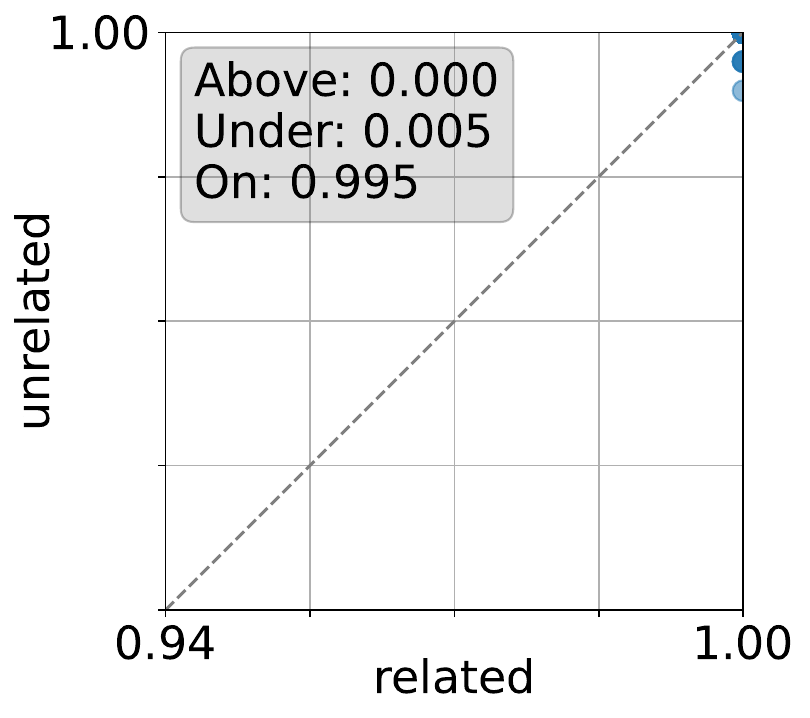} &
\includegraphics[width=0.15\textwidth]{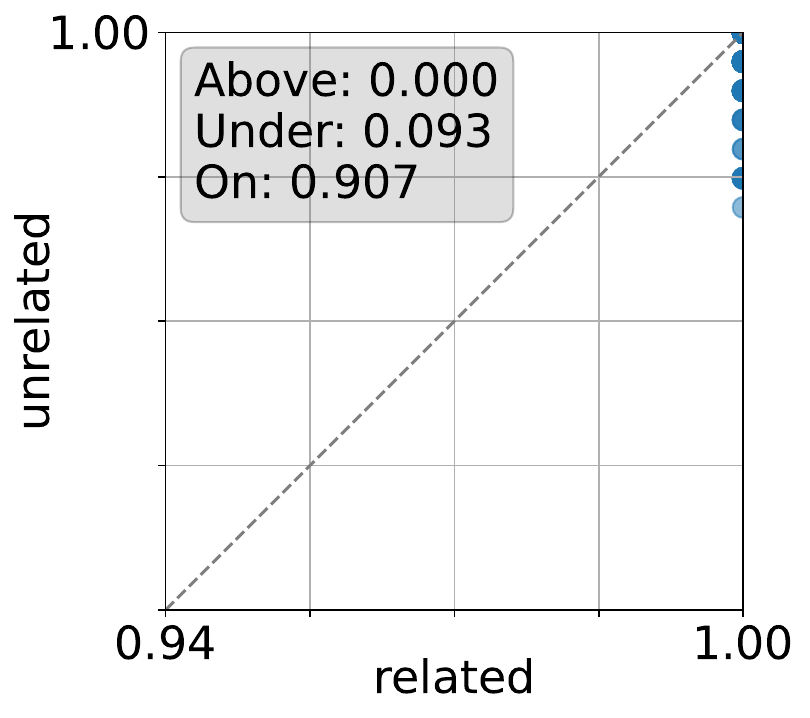} &
\includegraphics[width=0.15\textwidth]{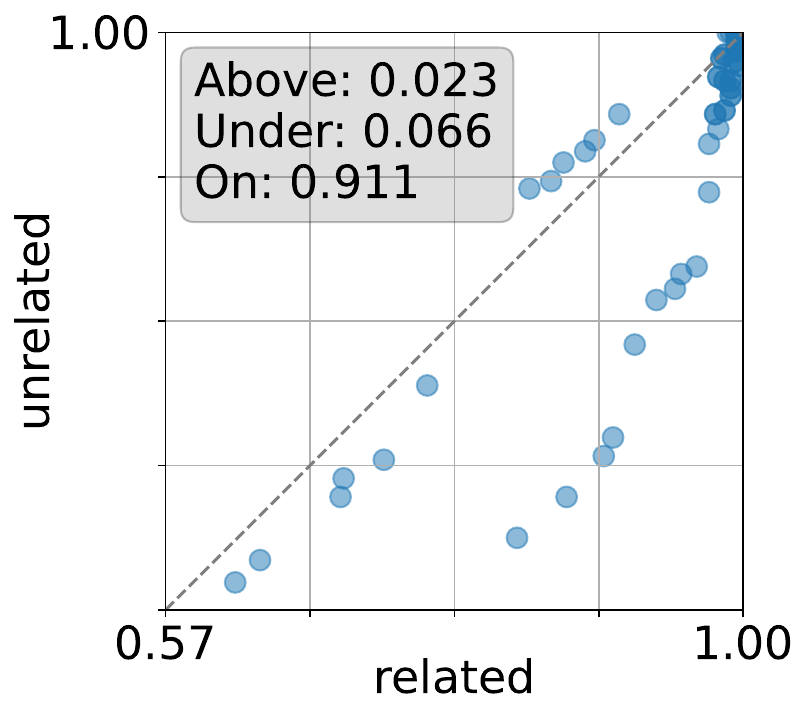} &
\includegraphics[width=0.15\textwidth]{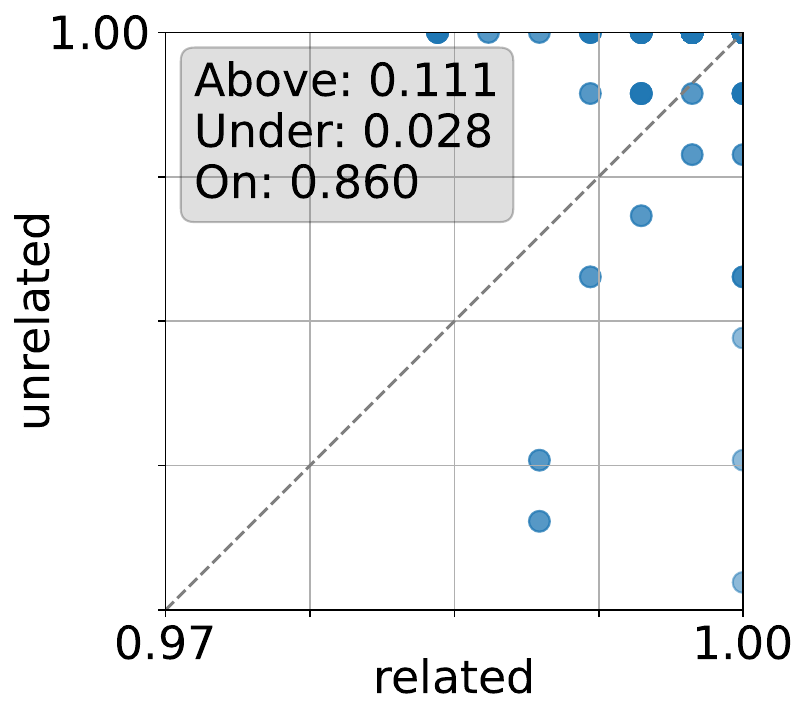} &
\includegraphics[width=0.15\textwidth]{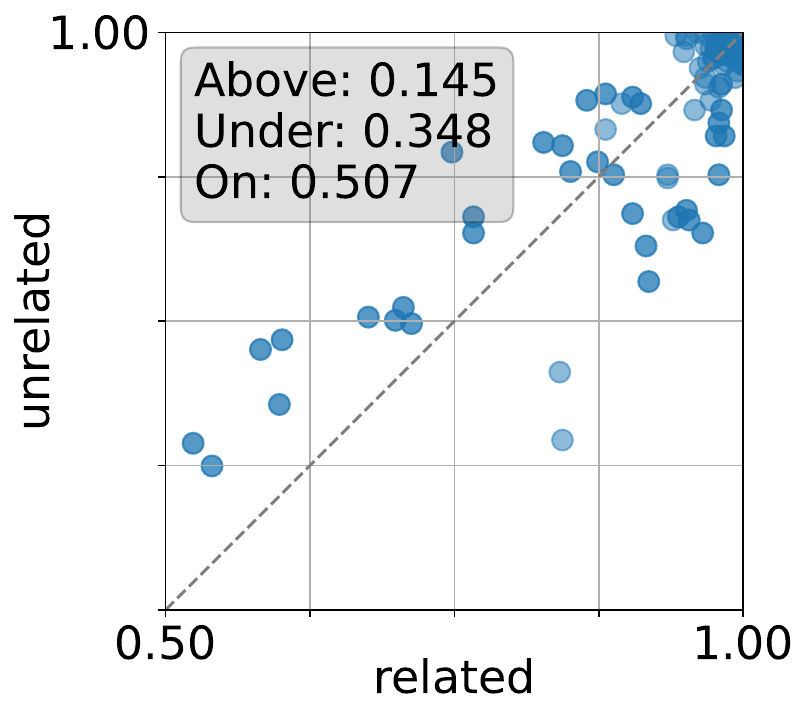} \\
\end{tabular}
\caption{Scatter plots illustrating the relationship of \textbf{unrelated and related word sets} in the NL1, NL2, CS and CA prompt categories (rows from top to bottom).
A point belongs to the same prompt template, the two coordinates are average accuracies over queries with related and unrelated
sets of the given type.}
\label{fig:supp_scatter_related}
\end{figure*}

\begin{figure*}
\centering
\setlength{\tabcolsep}{-2pt}
\begin{tabular}{ccccccc}
\scriptsize Llama3.3-70B &
\scriptsize Llama3.1-70B & 
\scriptsize Phi-3.5-MoE &
\scriptsize Qwen2.5-32B &
\scriptsize Mistral-24B &
\scriptsize Llama-3.1-8B &
\scriptsize Llama-3.2-3B \\
\includegraphics[width=0.15\textwidth]{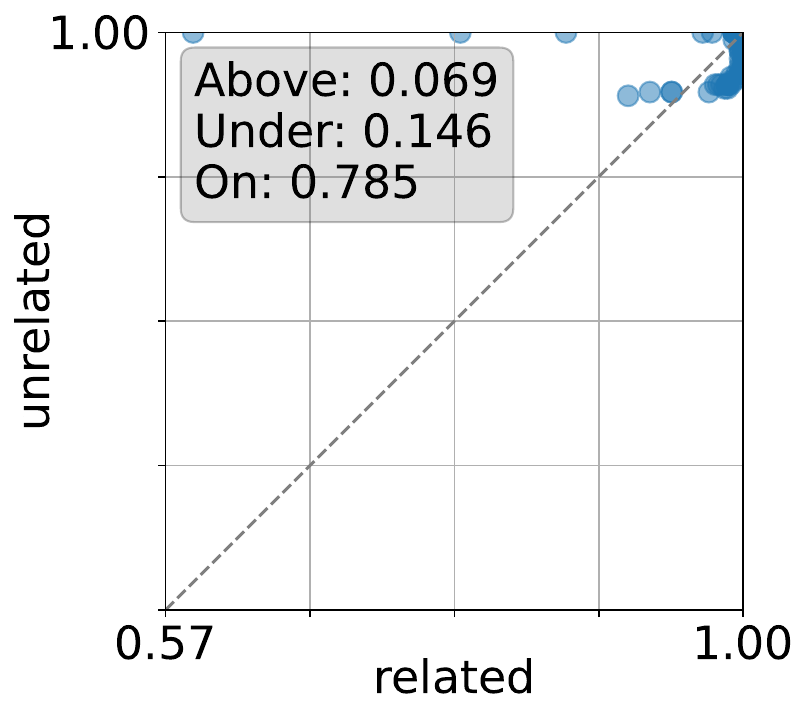} &
\includegraphics[width=0.15\textwidth]{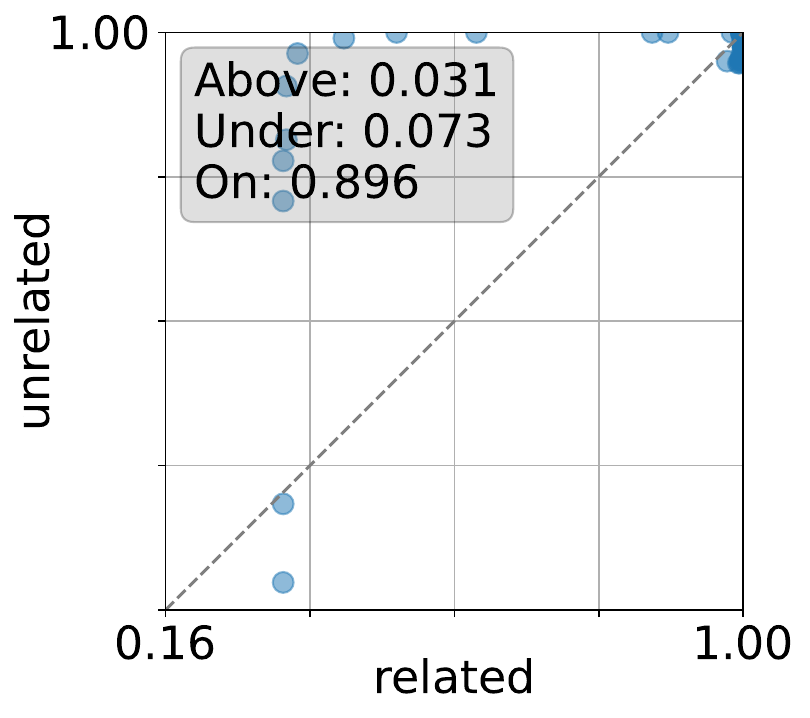} &
\includegraphics[width=0.15\textwidth]{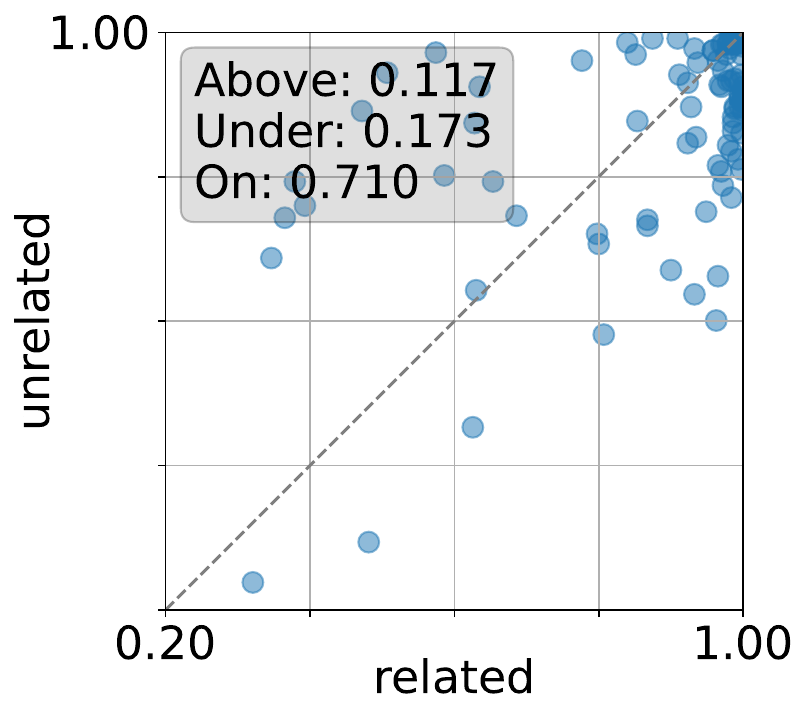} &
\includegraphics[width=0.15\textwidth]{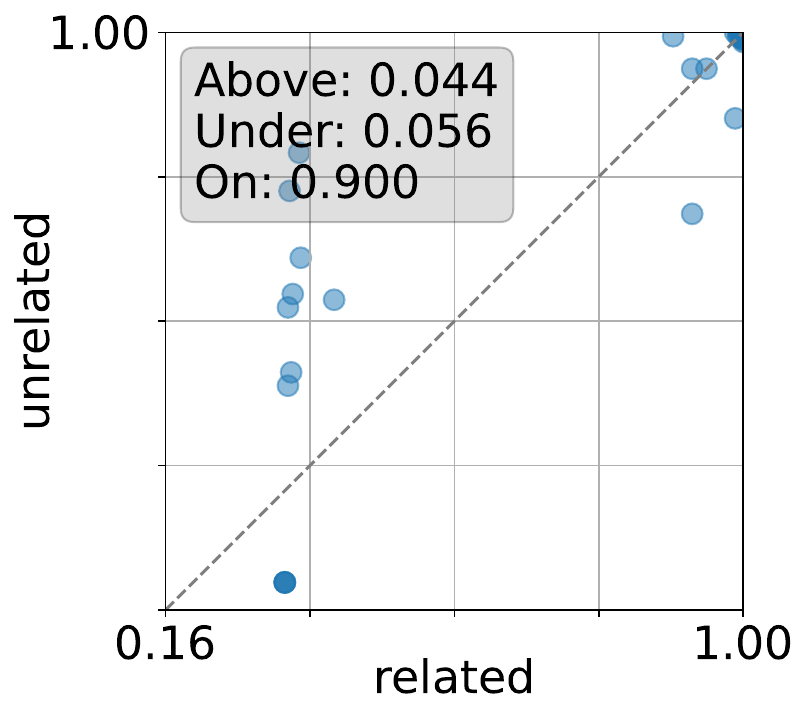} &
\includegraphics[width=0.15\textwidth]{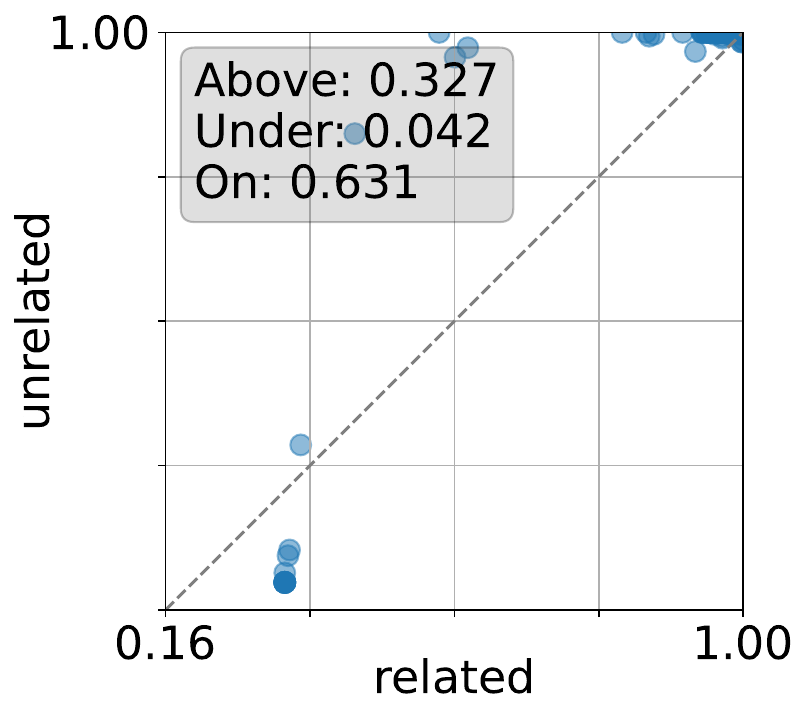} &
\includegraphics[width=0.15\textwidth]{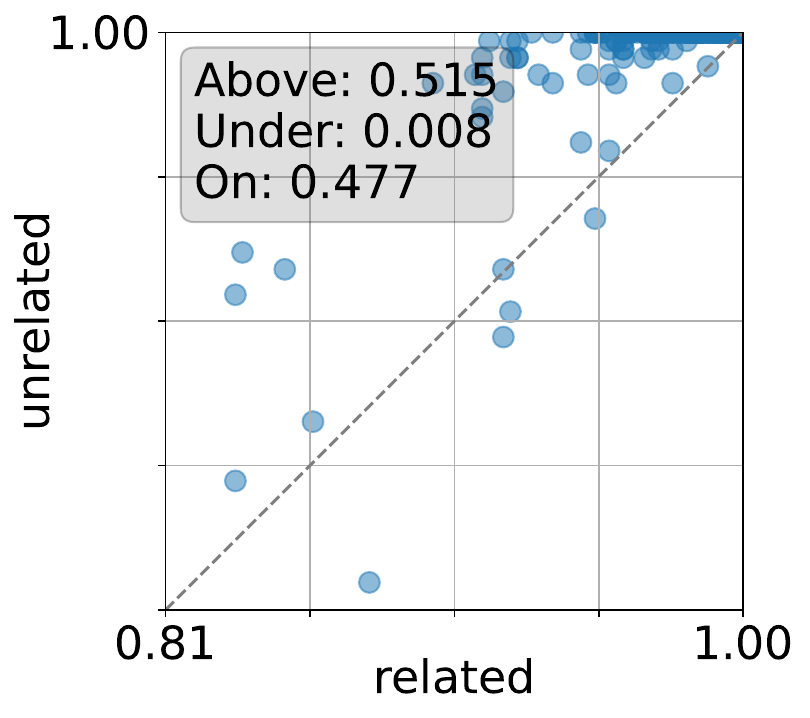} &
\includegraphics[width=0.15\textwidth]{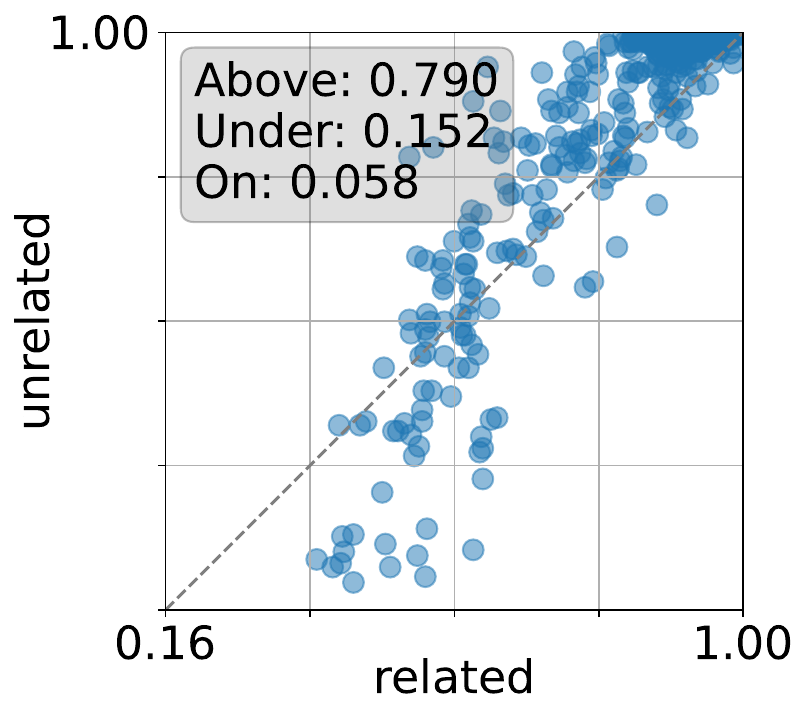} \\
\includegraphics[width=0.15\textwidth]{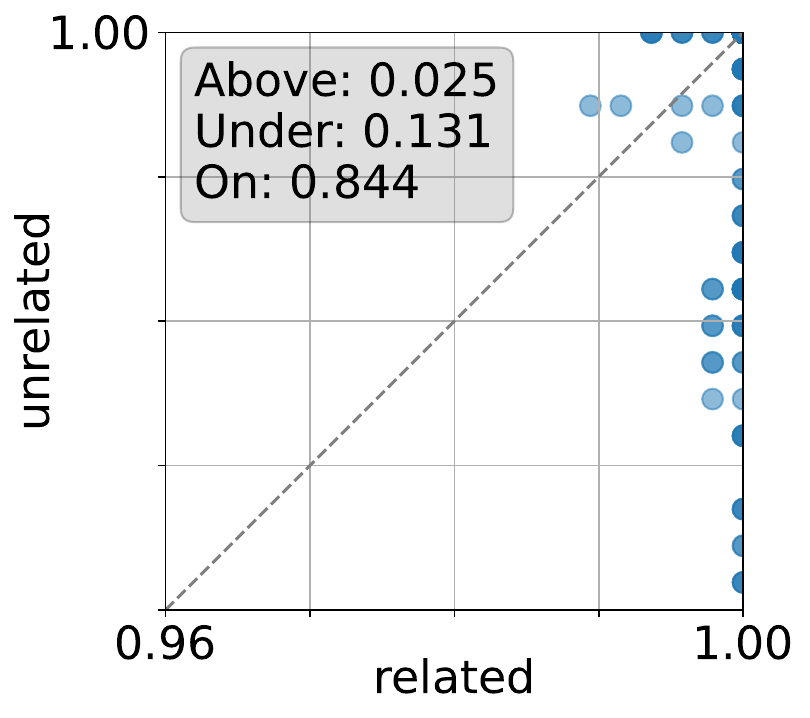} &
\includegraphics[width=0.15\textwidth]{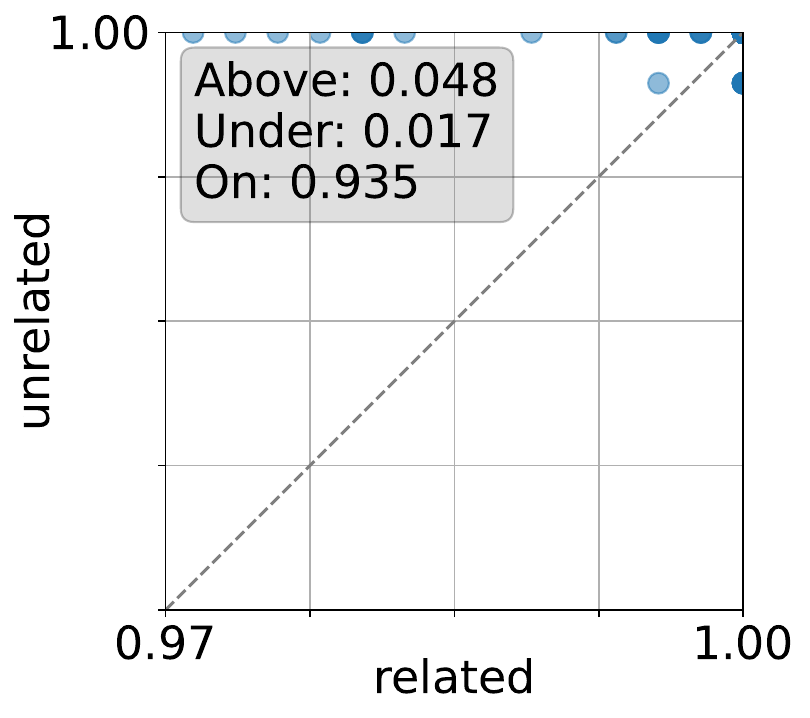} &
\includegraphics[width=0.15\textwidth]{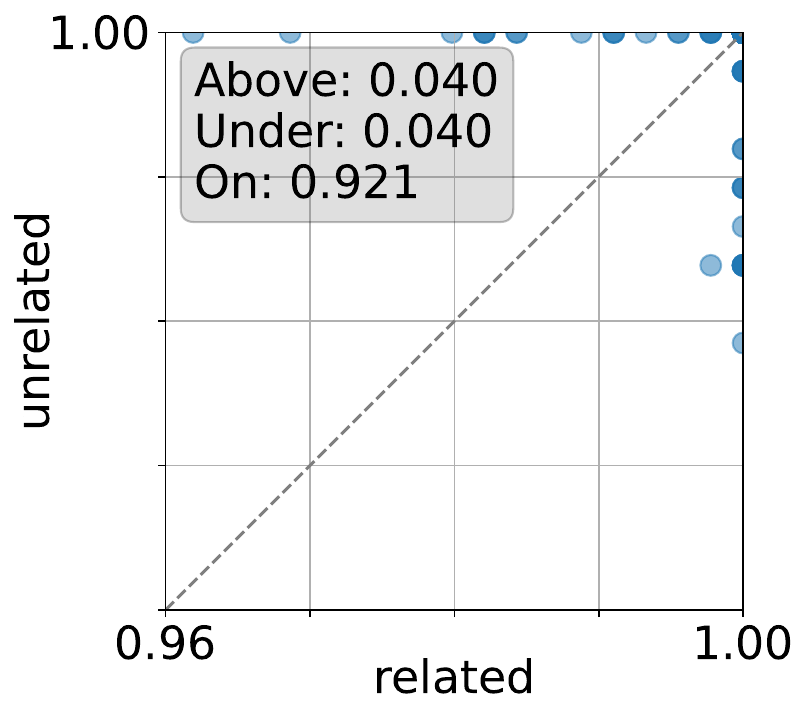} &
\includegraphics[width=0.15\textwidth]{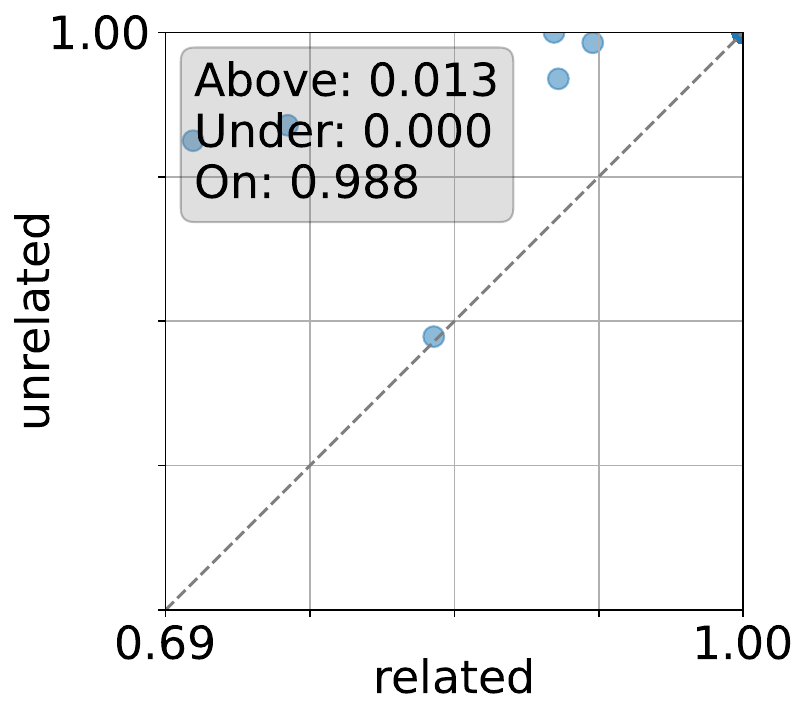} &
\includegraphics[width=0.15\textwidth]{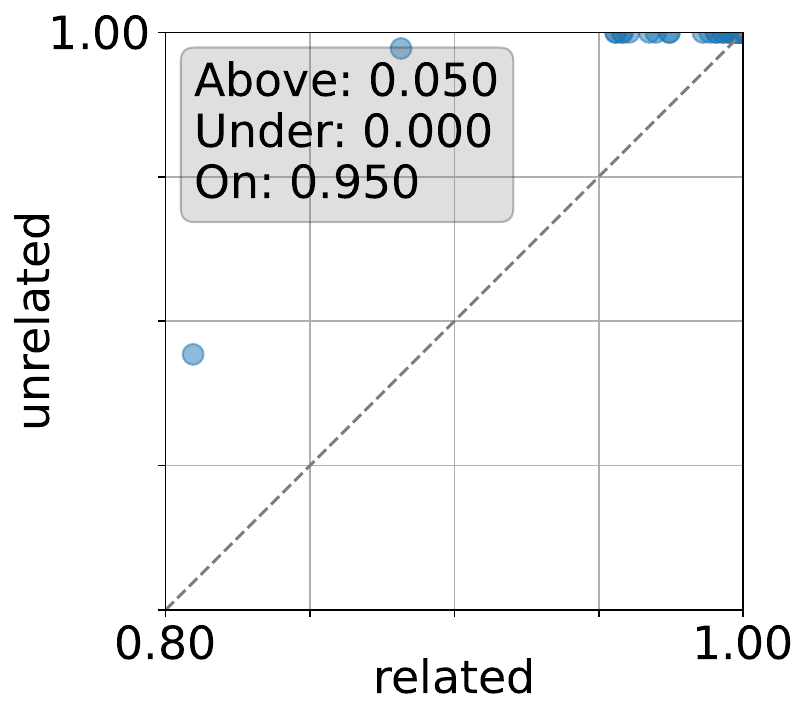} &
\includegraphics[width=0.15\textwidth]{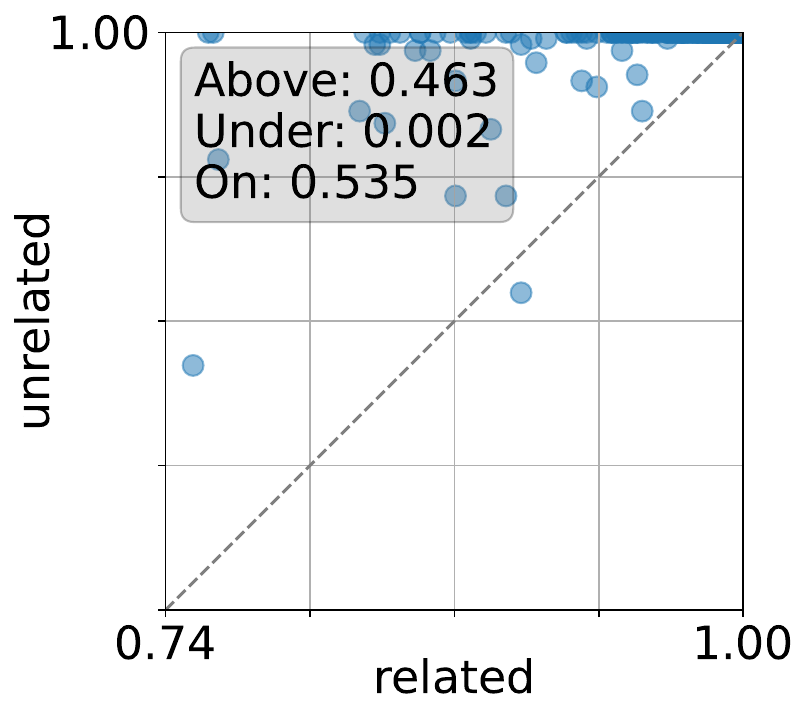} &
\includegraphics[width=0.15\textwidth]{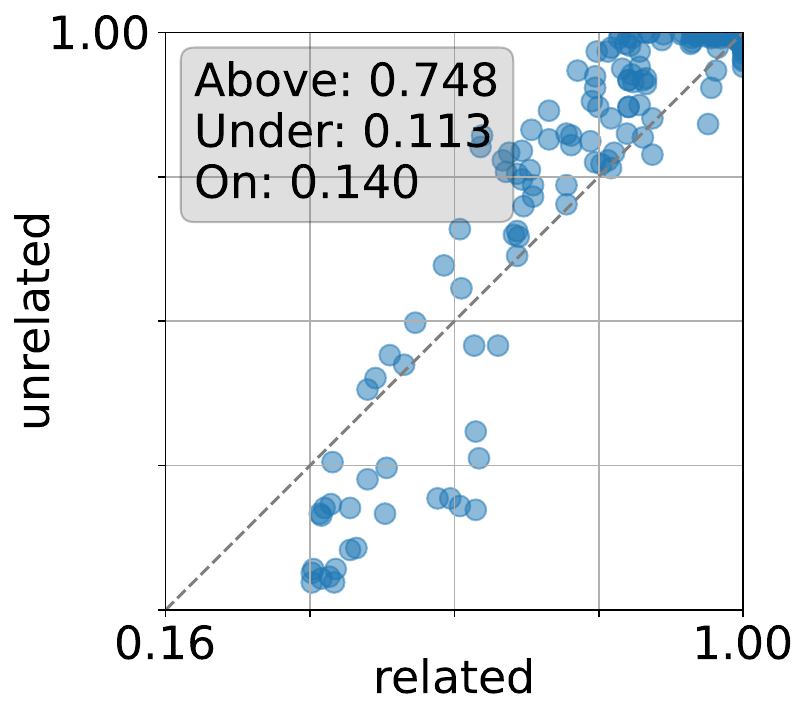} \\
\includegraphics[width=0.15\textwidth]{scatter/CS_llama_70b_3.3_numbers_accuracy} &
\includegraphics[width=0.15\textwidth]{scatter/CS_llama_70b_3.1_numbers_accuracy} &
\includegraphics[width=0.15\textwidth]{scatter/CS_phi_moe_numbers_accuracy} &
\includegraphics[width=0.15\textwidth]{scatter/CS_qwen_32b_numbers_accuracy} &
\includegraphics[width=0.15\textwidth]{scatter/CS_mistral_24b_numbers_accuracy} &
\includegraphics[width=0.15\textwidth]{scatter/CS_llama_8b_numbers_accuracy} &
\includegraphics[width=0.15\textwidth]{scatter/CS_llama_3b_numbers_accuracy} \\
\includegraphics[width=0.15\textwidth]{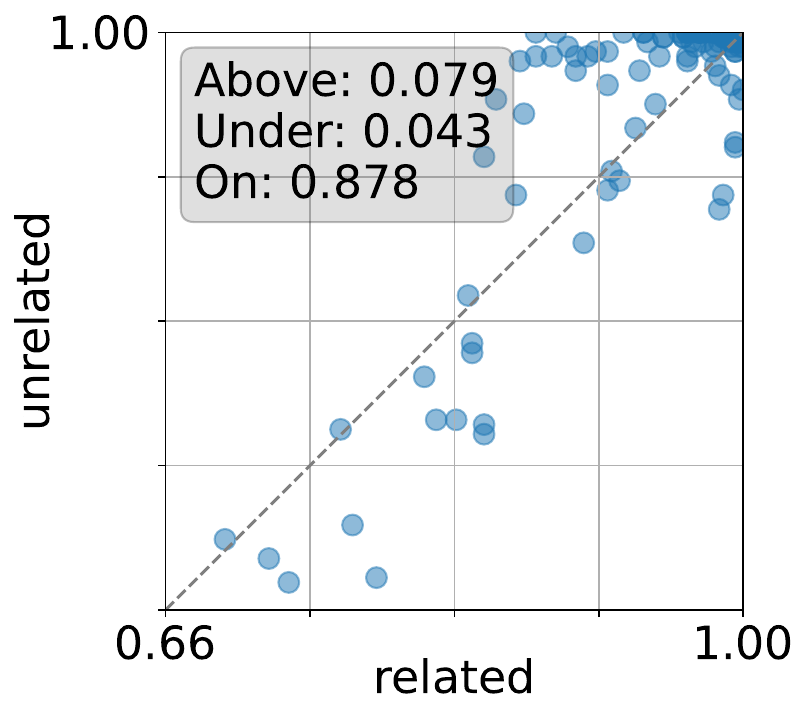} &
\includegraphics[width=0.15\textwidth]{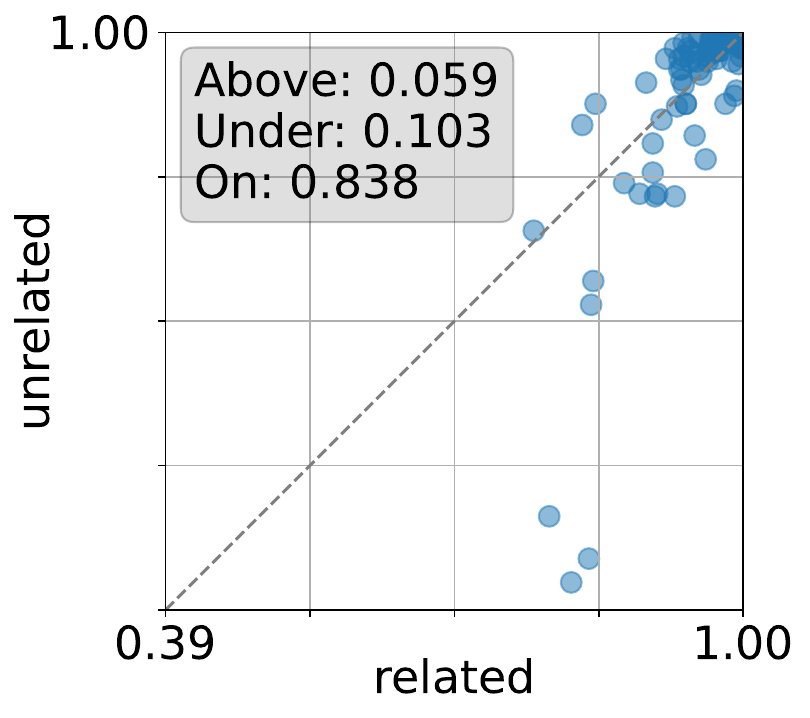} &
\includegraphics[width=0.15\textwidth]{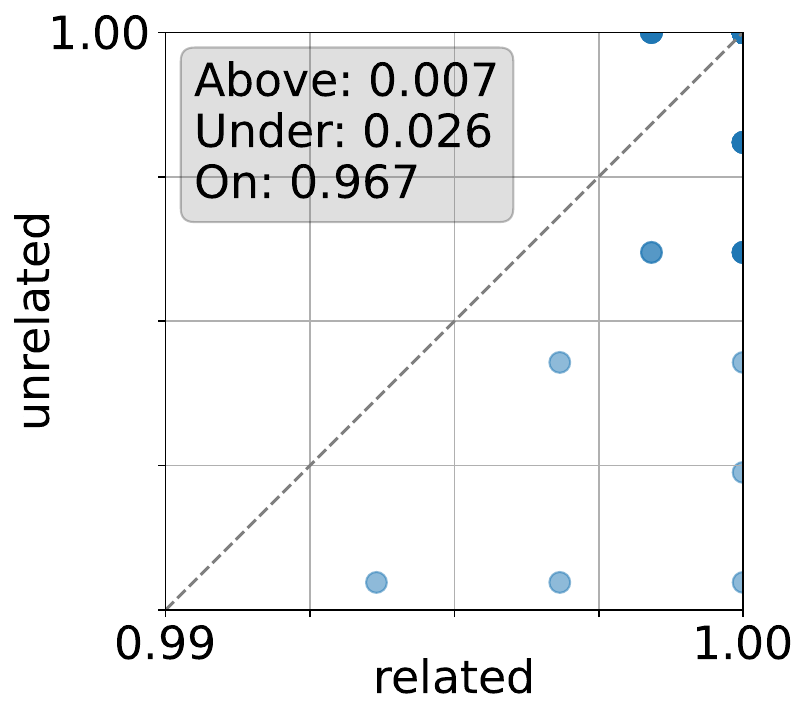} &
\includegraphics[width=0.15\textwidth]{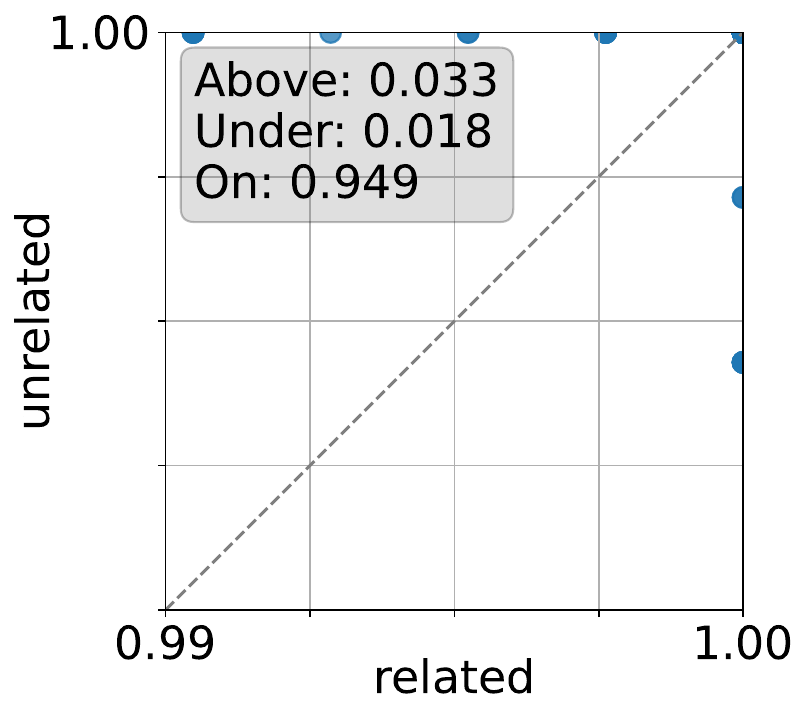} &
\includegraphics[width=0.15\textwidth]{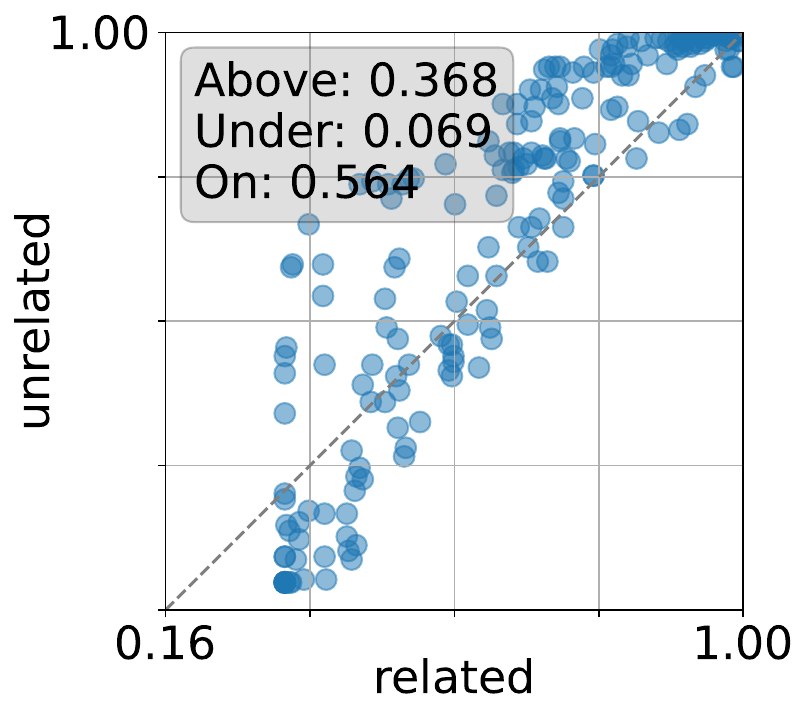} &
\includegraphics[width=0.15\textwidth]{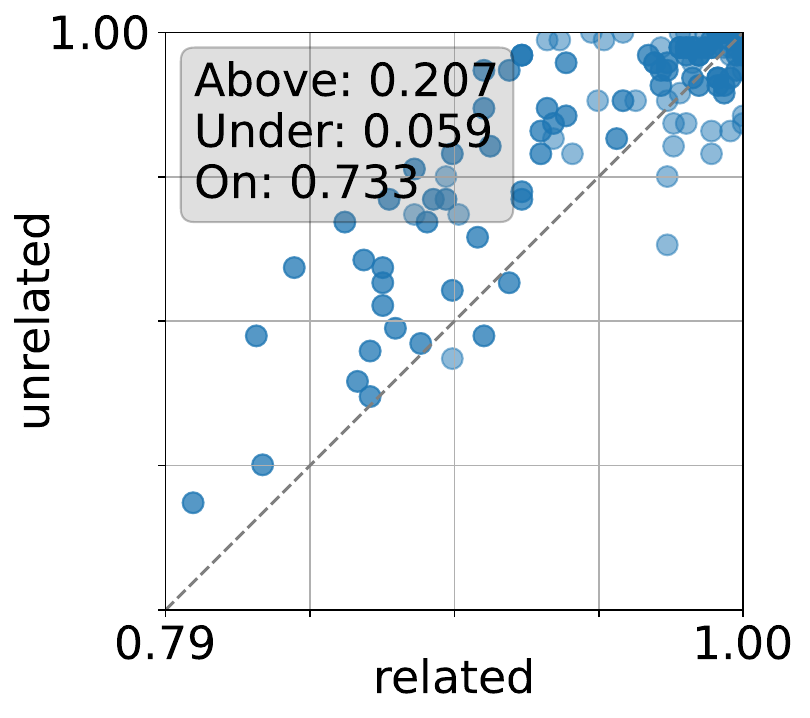} &
\includegraphics[width=0.15\textwidth]{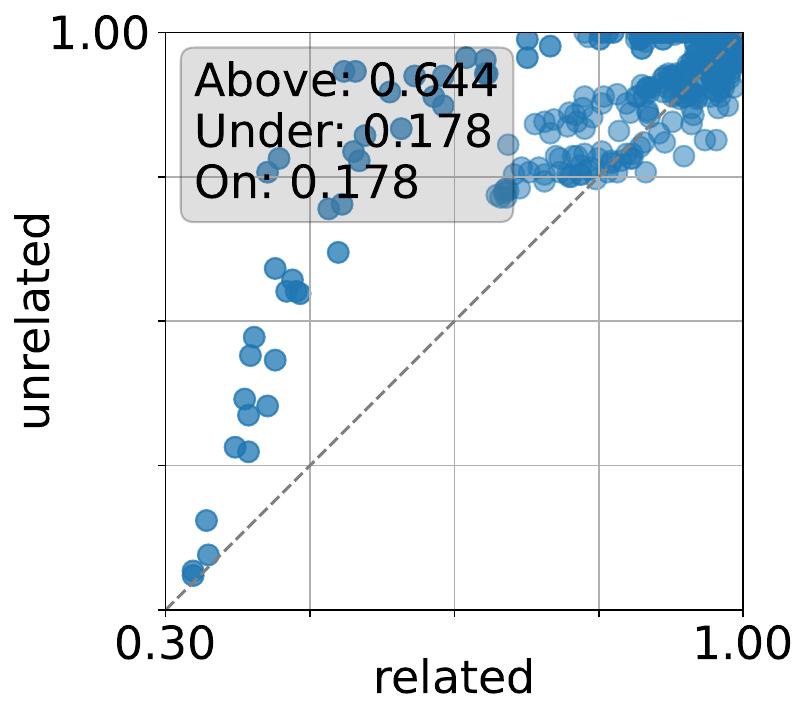} \\
\end{tabular}
\caption{Scatter plots illustrating the relationship of \textbf{unrelated and related numbers sets} in the NL1, NL2, CS and CA prompt categories.
A point belongs to the same prompt template, the two coordinates are average accuracies over queries with related and unrelated
sets of the given type.}
\label{fig:supp_scatter_numbers}
\end{figure*}

\Cref{fig:supp_settype} shows the complete set of results regarding the dependence of
accuracy on set-types, and
\Cref{fig:supp_scatter_complete,fig:supp_scatter_related,fig:supp_scatter_numbers} show the complete
set of scatter plots illustrating the connection between semantic relatedness and unrelatedness
for reference.
The plots give further support to the claims in the paper but also allow one to examine the relationship
between complete and related word sets, or between CS and CA prompts.

\begin{figure*}
\centering
\begin{tabular}{cc}
\scriptsize \phantom{XXXXXX}Llama3.3-70B, NL1 &
\scriptsize \phantom{XXXXXX}Llama3.1-70B, NL1 \\
\includegraphics[width=.45\textwidth]{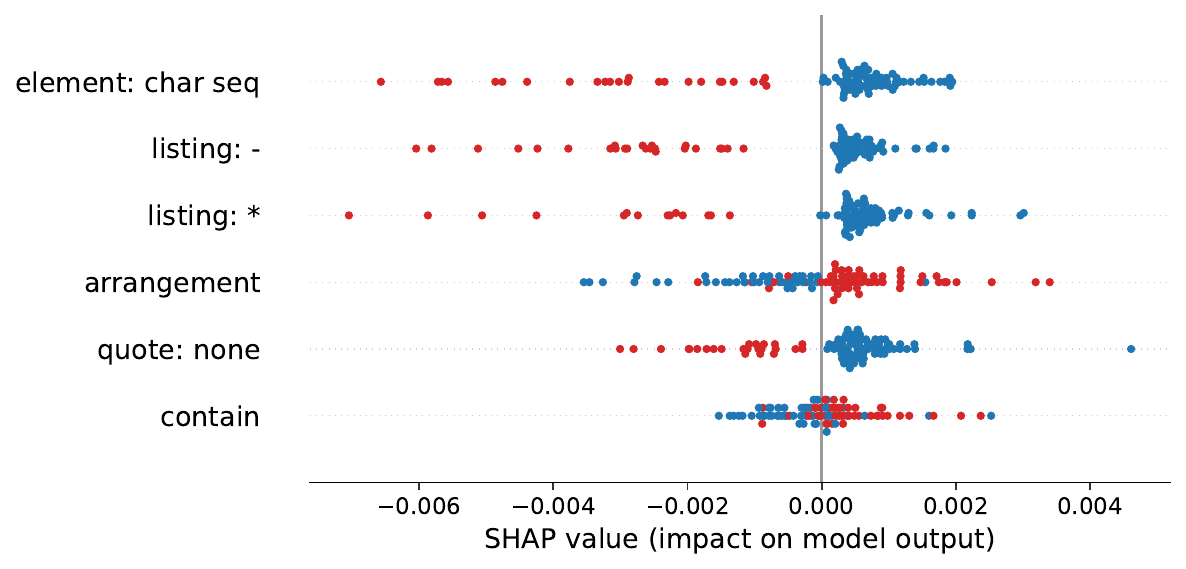} &
\includegraphics[width=.45\textwidth]{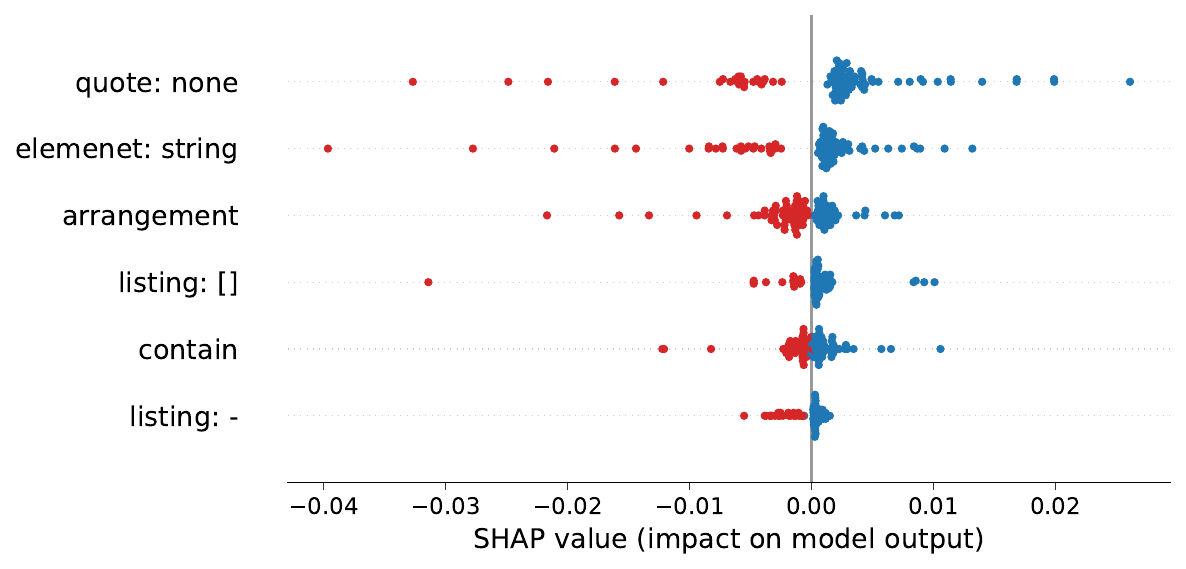}\\[5mm]
\scriptsize \phantom{XXXXXXX}Phi3.5-MoE, NL1 &
\scriptsize \phantom{XXXXXXX}Qwen2.5-32B, NL1 \\
\includegraphics[width=.45\textwidth]{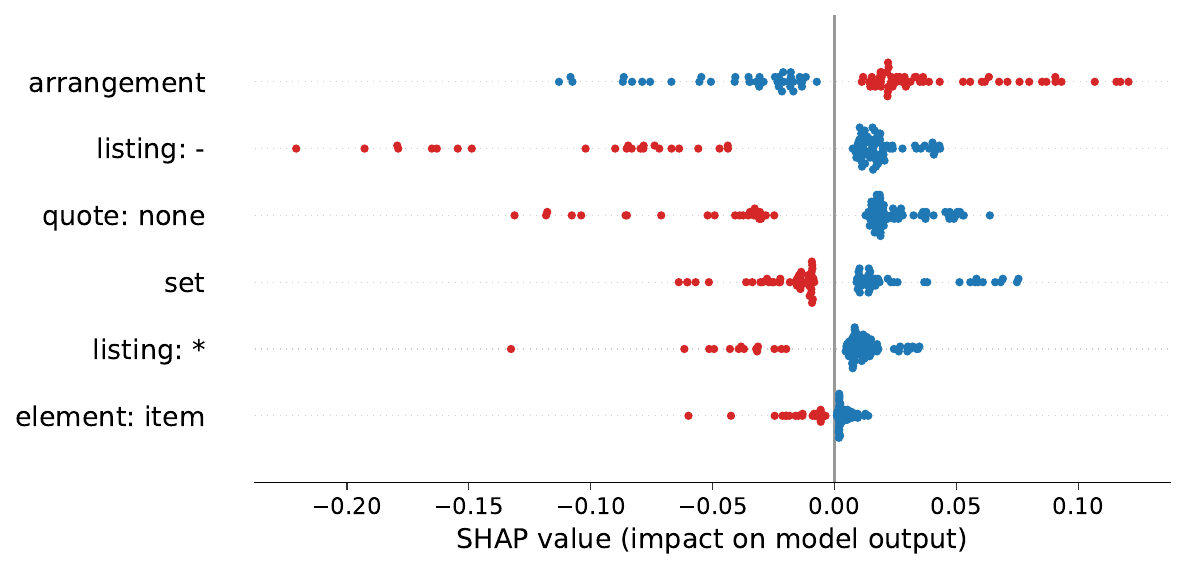} &
\includegraphics[width=.45\textwidth]{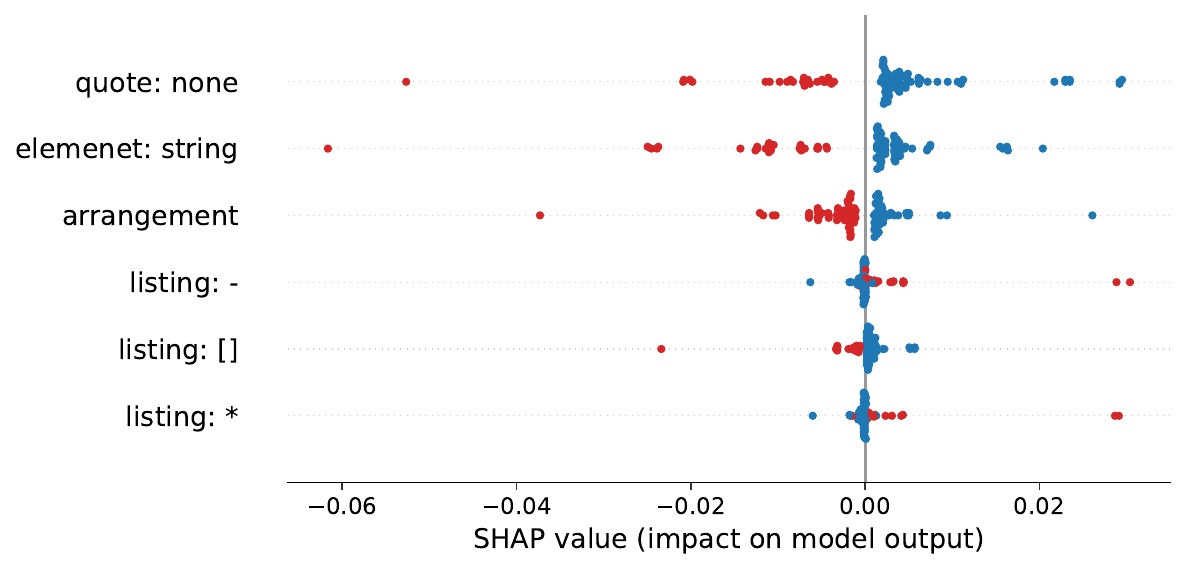}\\[5mm]
\scriptsize \phantom{XXXXXXX}Mistral-24B, NL1 &
\scriptsize \phantom{XXXXXXX}Llama-3.1-8B, NL1 \\
\includegraphics[width=.45\textwidth]{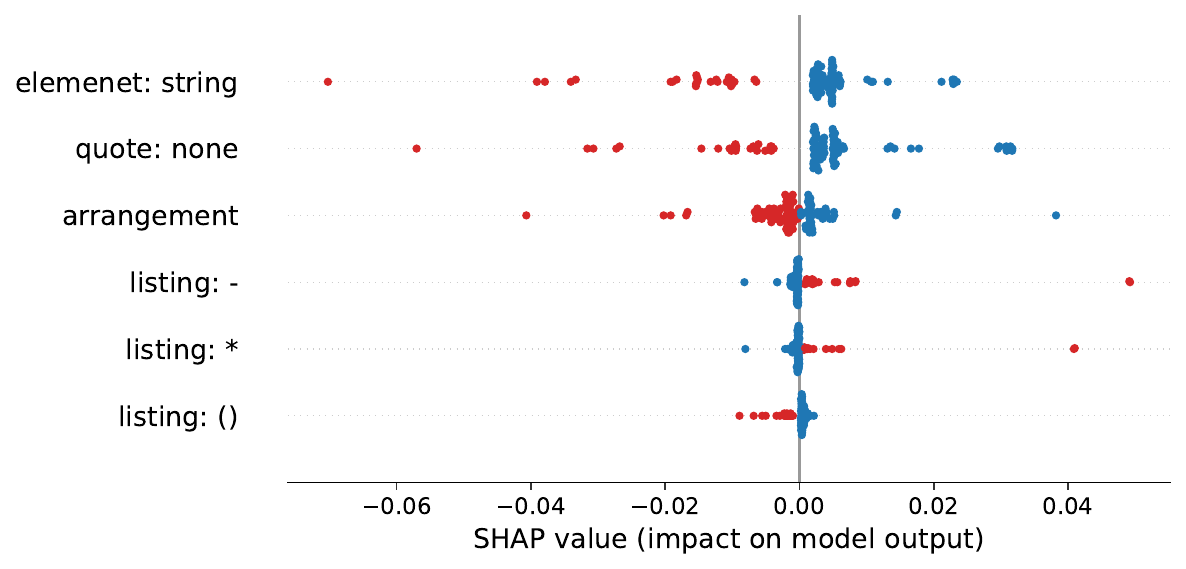} &
\includegraphics[width=.45\textwidth]{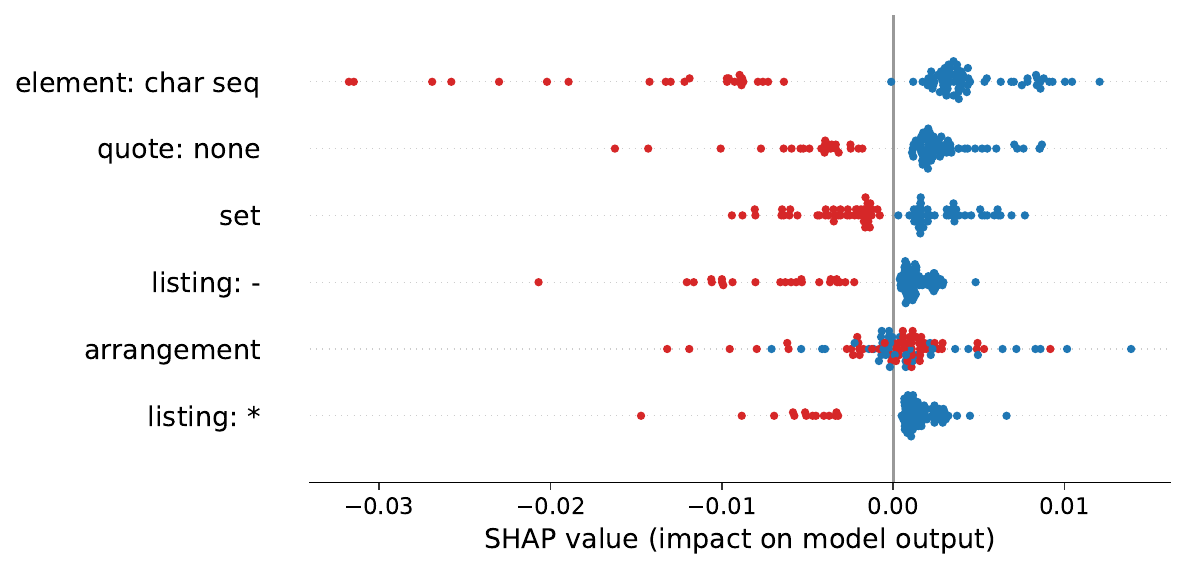}\\[5mm]
\end{tabular}
\begin{tabular}{c}
\scriptsize \phantom{XXXXXXX}Llama-3.2-3B, NL1 \\
\includegraphics[width=.45\textwidth]{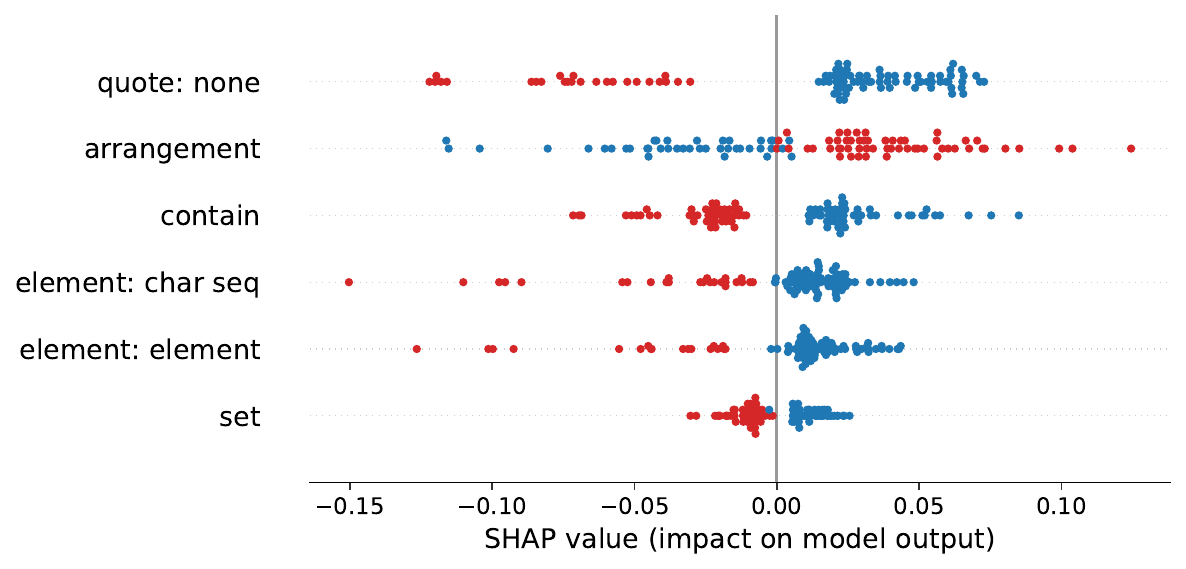}\\
\end{tabular}
\caption{Top 6 Shapley values among binary features of the \textbf{NL1 prompt category}.
Non-binary features are converted to one-hot representation.
A dot corresponds to a prompt template and the color red indicates that the feature is present.}
\label{fig:supp_shap_nl1}
\end{figure*}

\begin{figure*}
\centering
\begin{tabular}{cc}
\scriptsize \phantom{XXXXXX}Llama3.3-70B, NL2 &
\scriptsize \phantom{XXXXXX}Llama3.1-70B, NL2 \\
\includegraphics[width=.45\textwidth]{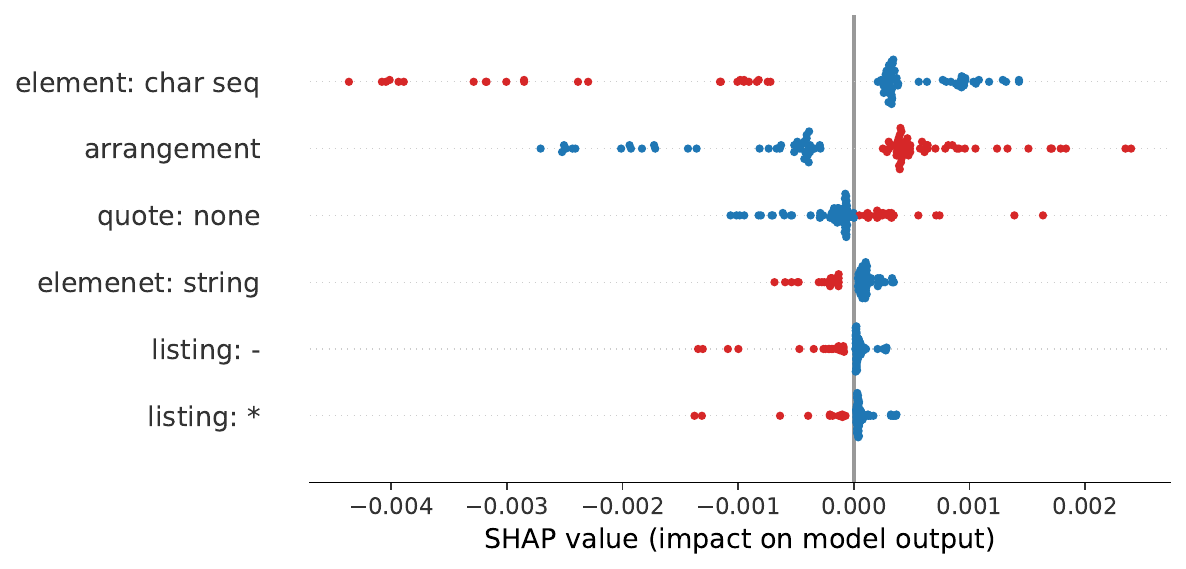} &
\includegraphics[width=.45\textwidth]{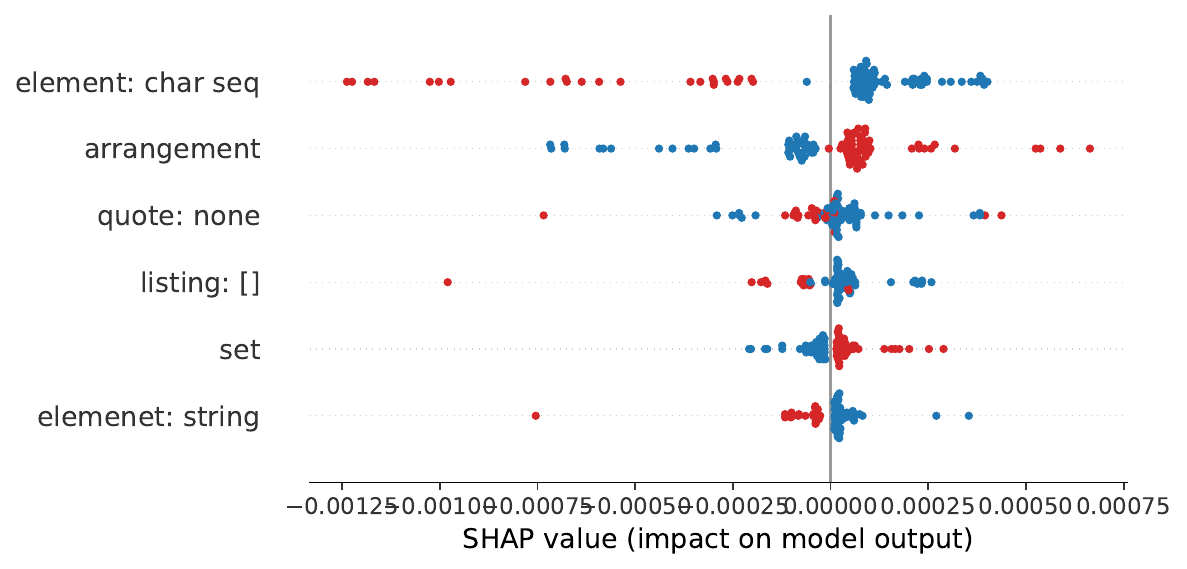}\\[5mm]
\scriptsize \phantom{XXXXXXX}Phi3.5-MoE, NL2 &
\scriptsize \phantom{XXXXXXX}Qwen2.5-32B, NL2 \\
\includegraphics[width=.45\textwidth]{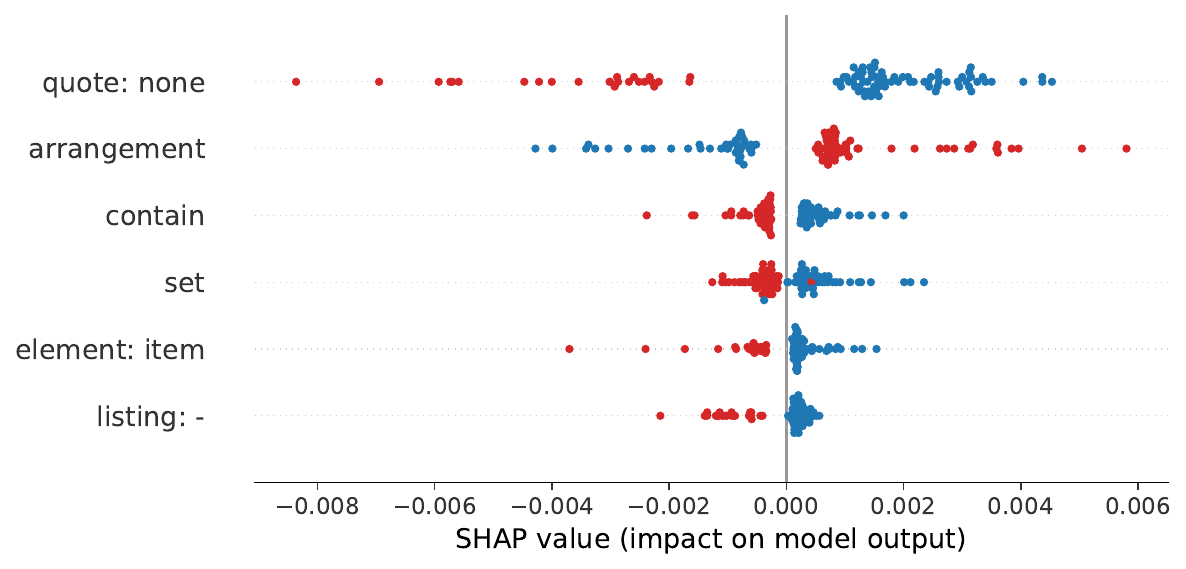} &
\includegraphics[width=.45\textwidth]{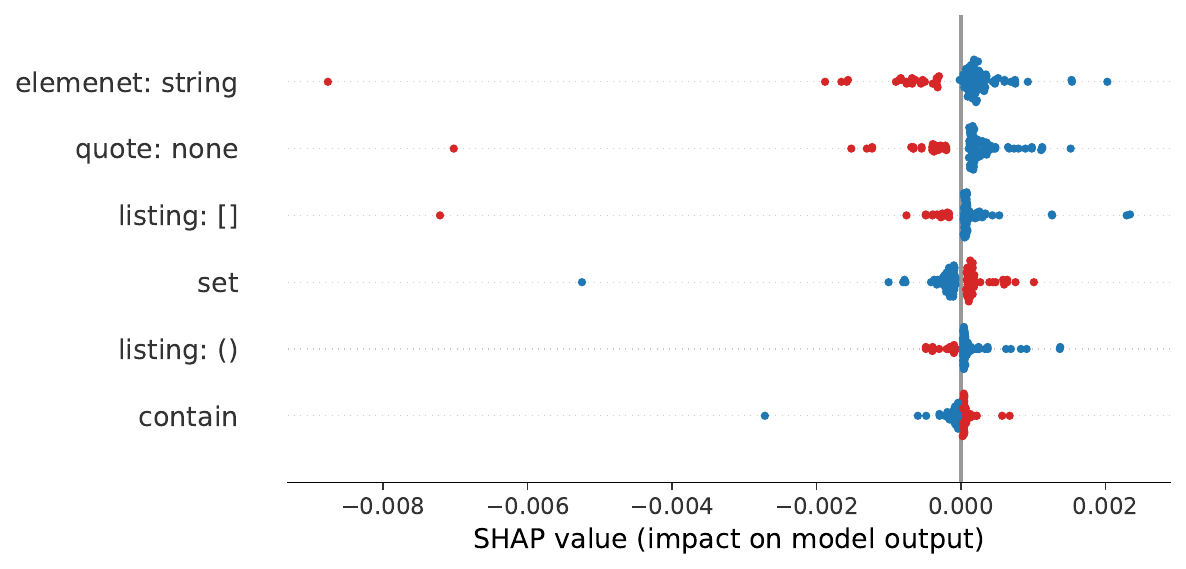}\\[5mm]
\scriptsize \phantom{XXXXXXX}Mistral-24B, NL2 &
\scriptsize \phantom{XXXXXXX}Llama-3.1-8B, NL2 \\
\includegraphics[width=.45\textwidth]{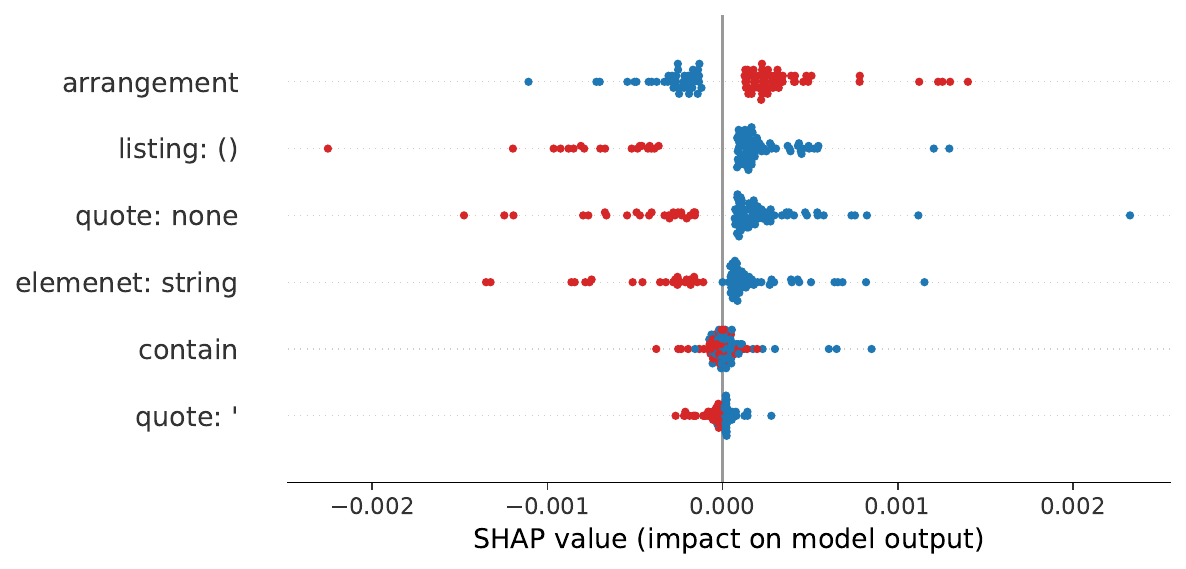} &
\includegraphics[width=.45\textwidth]{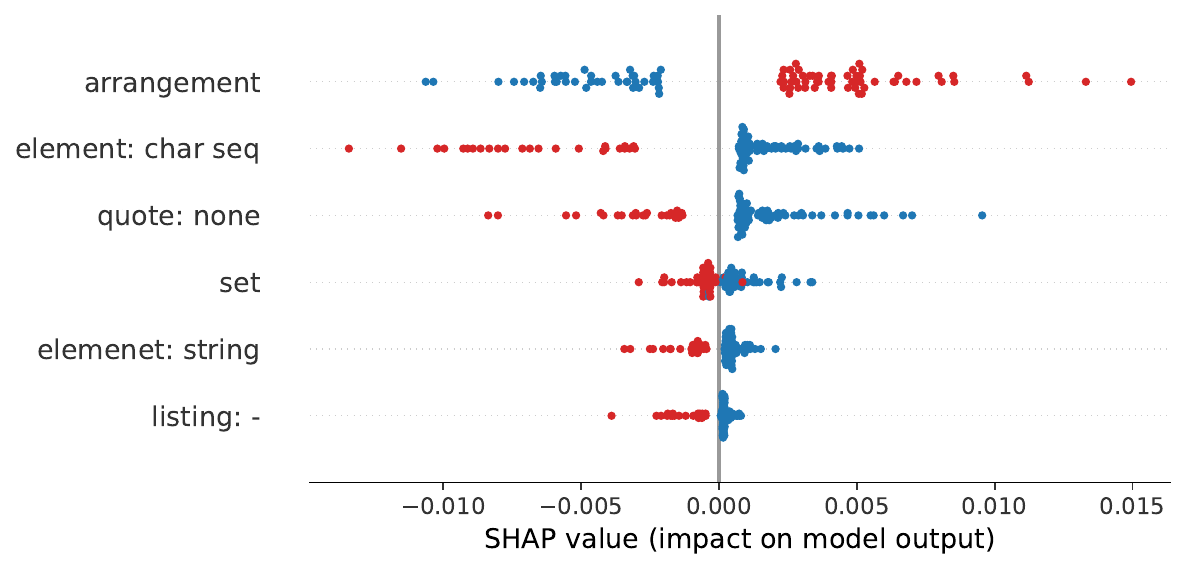}\\[5mm]
\end{tabular}
\begin{tabular}{c}
\scriptsize \phantom{XXXXXXX}Llama-3.2-3B, NL2 \\
\includegraphics[width=.45\textwidth]{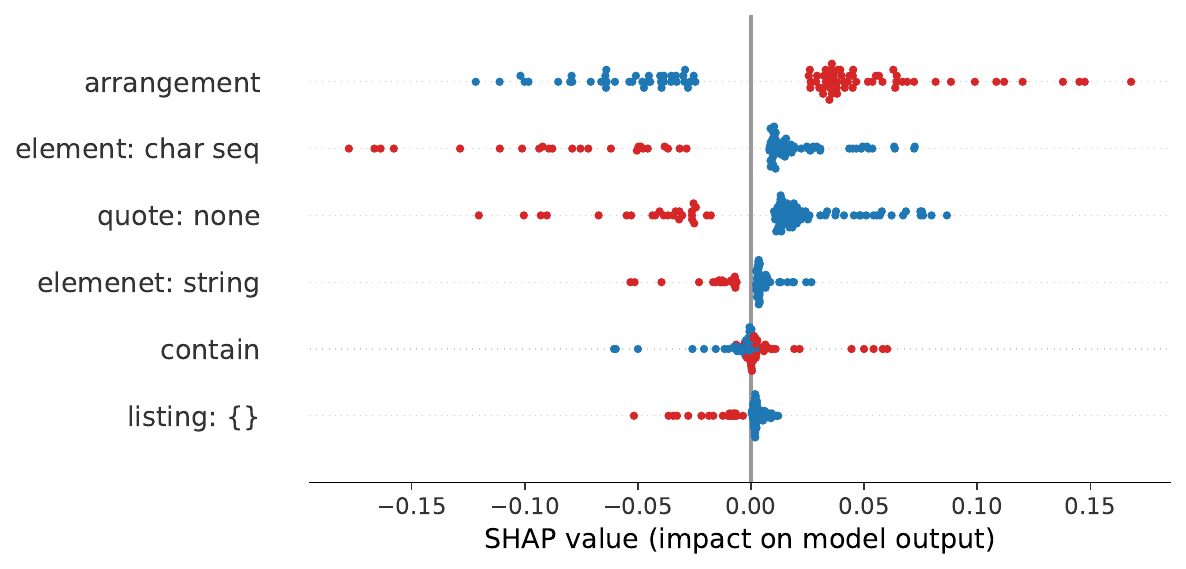}\\
\end{tabular}
\caption{Top 6 Shapley values among binary features of the \textbf{NL2 prompt category}.
Non-binary features are converted to one-hot representation.
A dot corresponds to a prompt template and the color red indicates that the feature is present.}
\label{fig:supp_shap_nl2}
\end{figure*}

\begin{figure*}
\centering
\begin{tabular}{cc}
\scriptsize \phantom{XXXXXX}Llama3.3-70B, CS &
\scriptsize \phantom{XXXXXX}Llama3.1-70B, CS \\
\includegraphics[width=.45\textwidth]{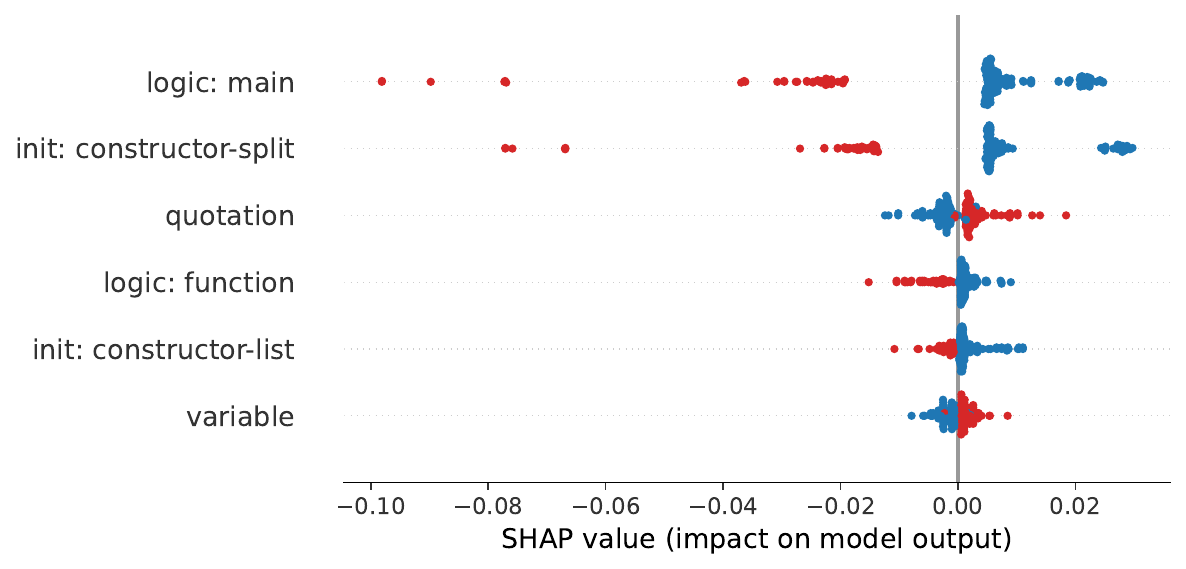} &
\includegraphics[width=.45\textwidth]{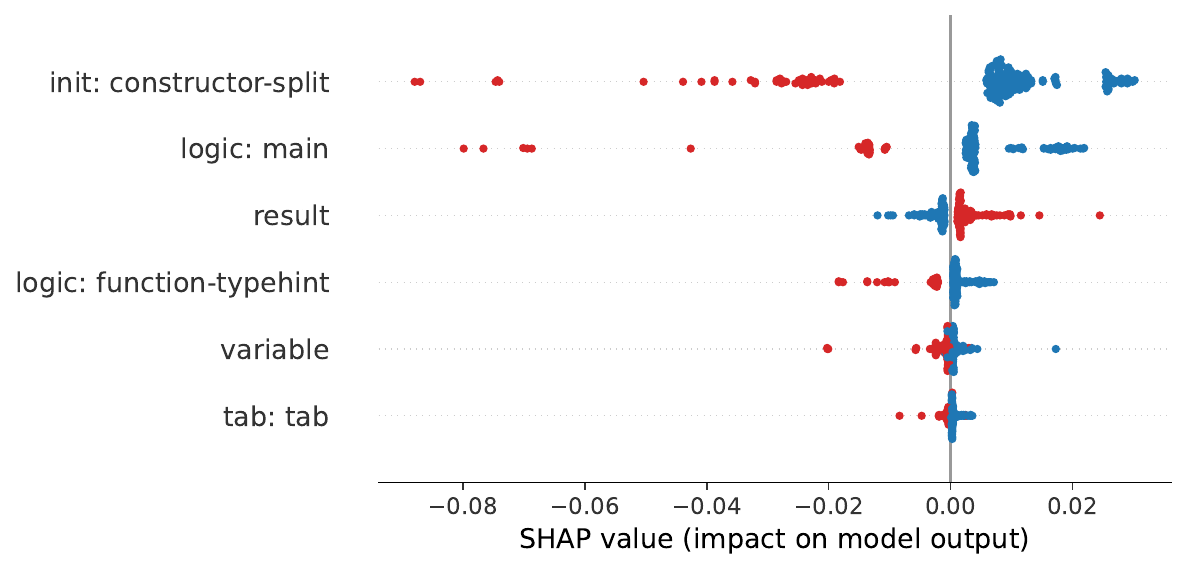}\\[5mm]
\scriptsize \phantom{XXXXXXX}Phi3.5-MoE, CS &
\scriptsize \phantom{XXXXXXX}Qwen2.5-32B, CS \\
\includegraphics[width=.45\textwidth]{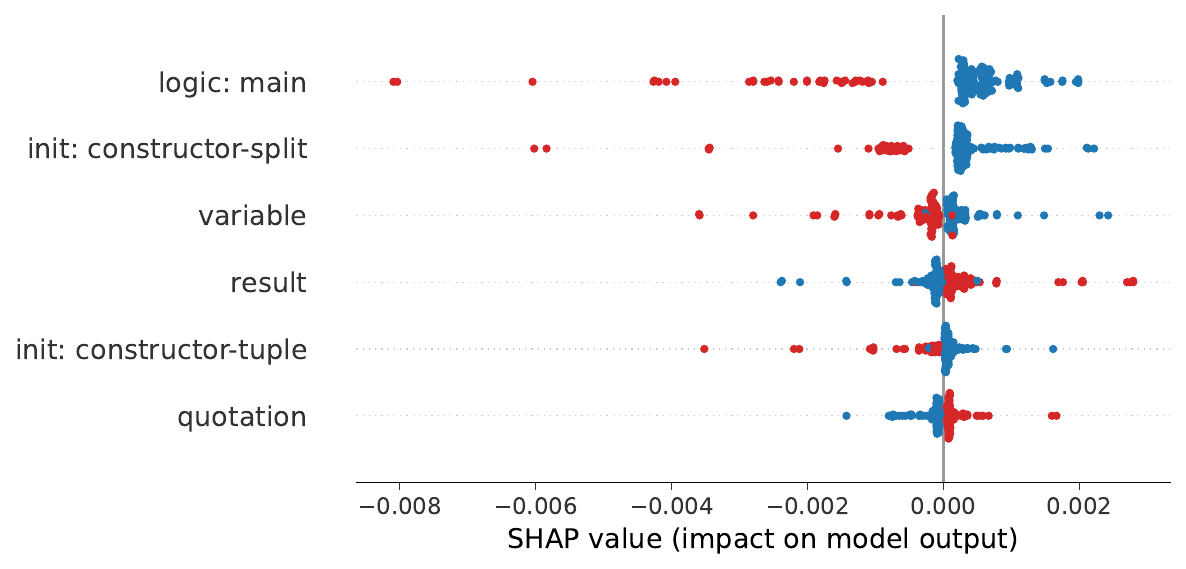} &
\includegraphics[width=.45\textwidth]{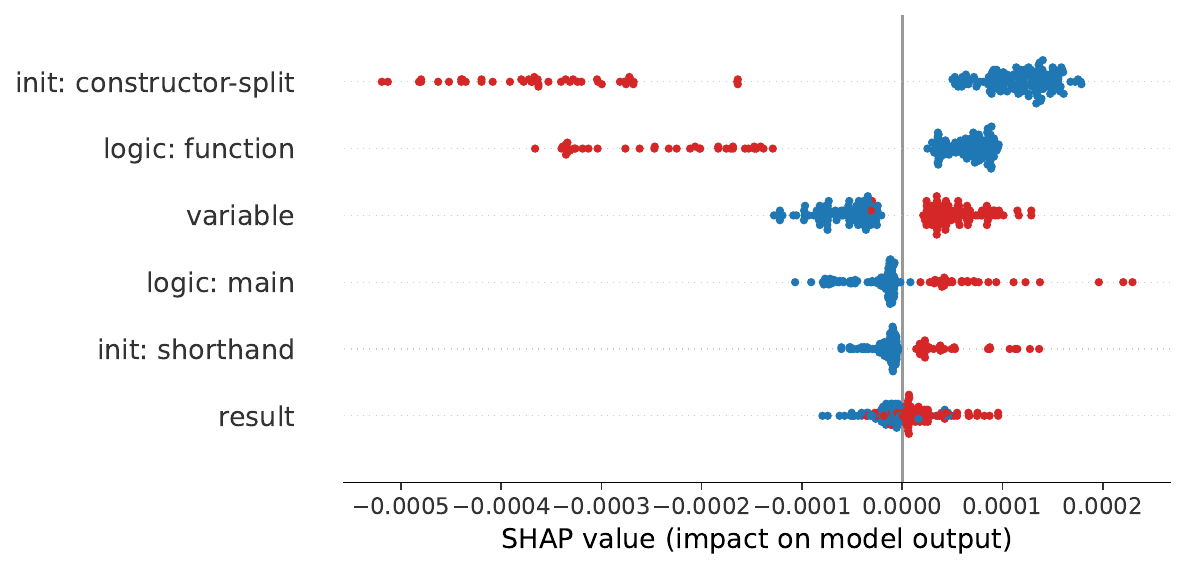}\\[5mm]
\scriptsize \phantom{XXXXXXX}Mistral-24B, CS &
\scriptsize \phantom{XXXXXXX}Llama-3.1-8B, CS \\
\includegraphics[width=.45\textwidth]{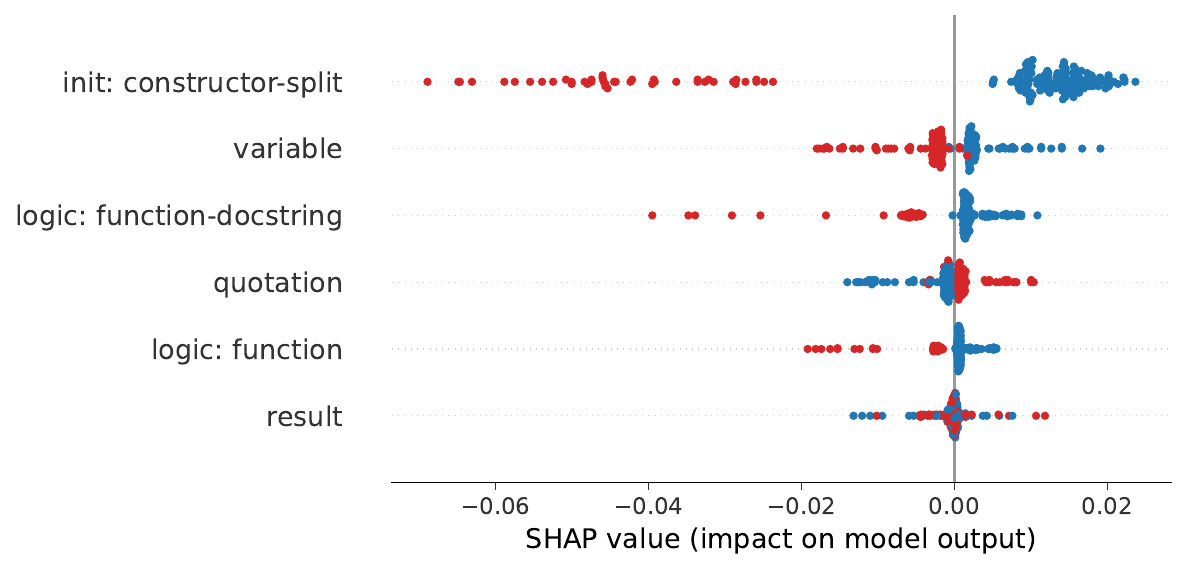} &
\includegraphics[width=.45\textwidth]{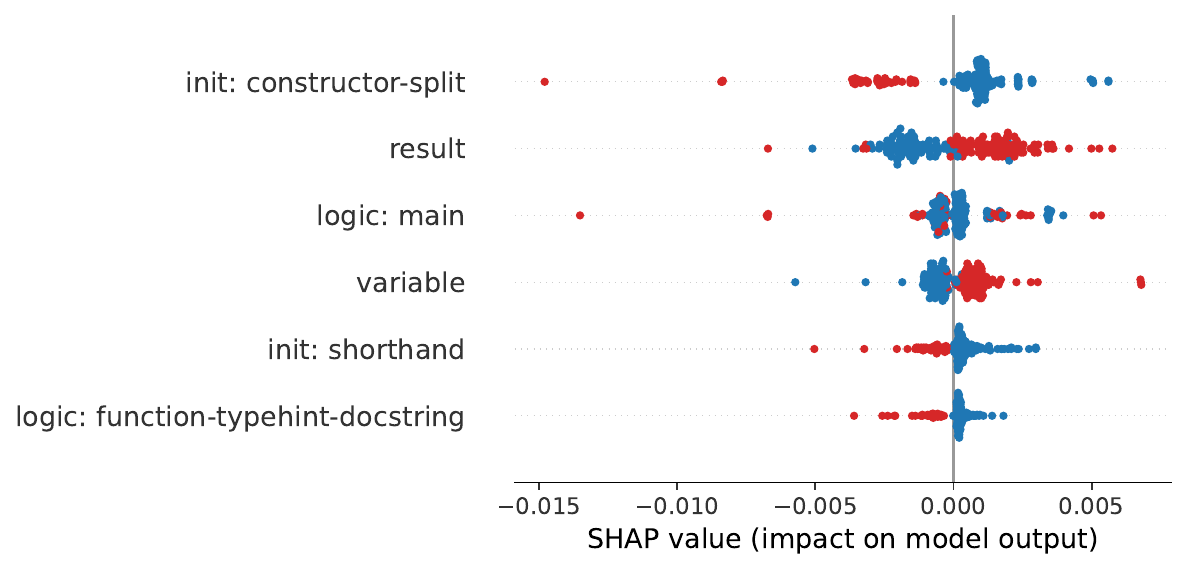}\\[5mm]
\end{tabular}
\begin{tabular}{c}
\scriptsize \phantom{XXXXXXX}Llama-3.2-3B, CS \\
\includegraphics[width=.45\textwidth]{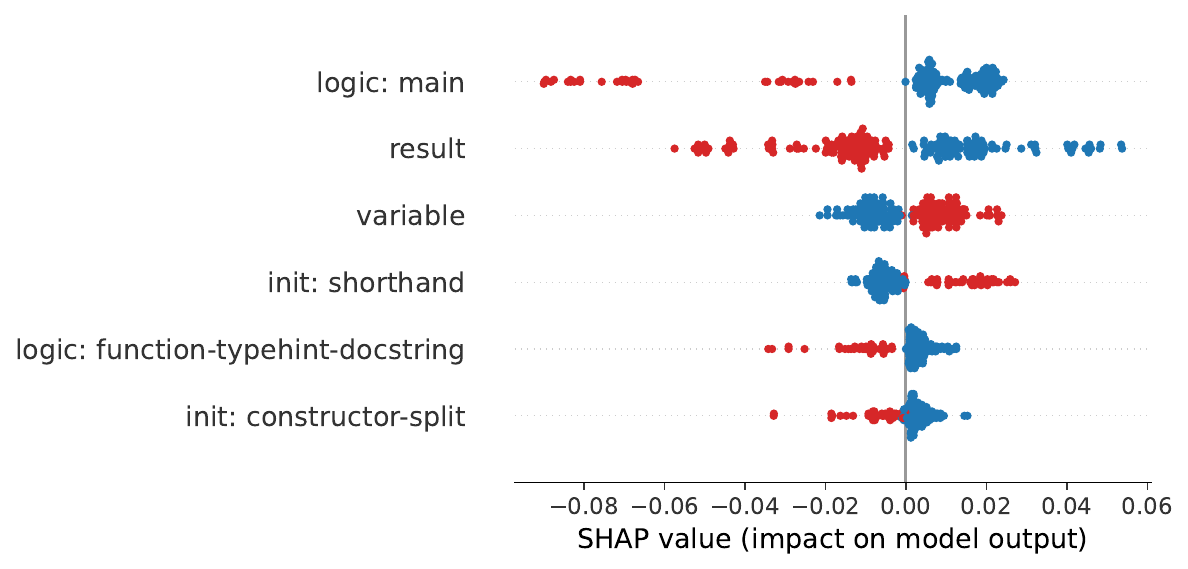}\\
\end{tabular}
\caption{Top 6 Shapley values among binary features of the \textbf{CS prompt category}.
Non-binary features are converted to one-hot representation.
A dot corresponds to a prompt template and the color red indicates that the feature is present.}
\label{fig:supp_shap_cs}
\end{figure*}

\begin{figure*}
\centering
\begin{tabular}{cc}
\scriptsize \phantom{XXXXXX}Llama3.3-70B, CA &
\scriptsize \phantom{XXXXXX}Llama3.1-70B, CA \\
\includegraphics[width=.45\textwidth]{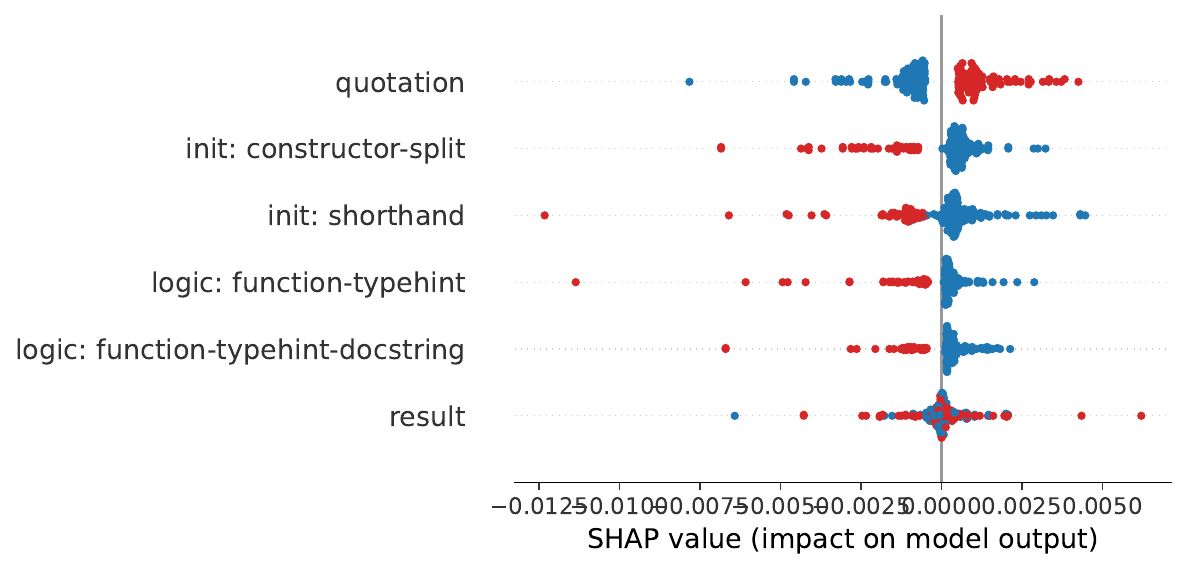} &
\includegraphics[width=.45\textwidth]{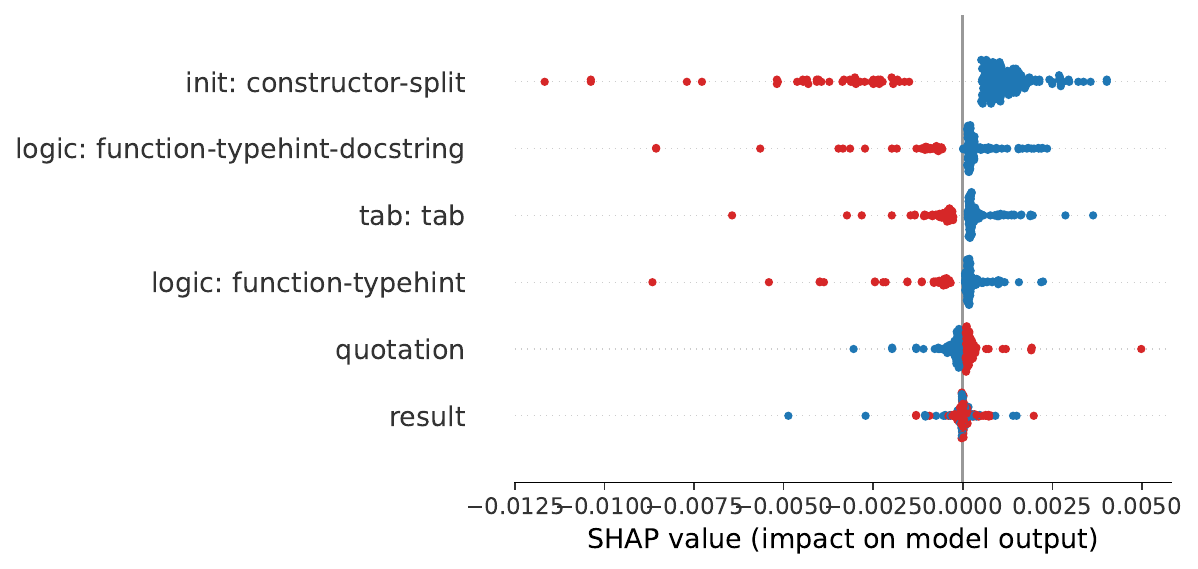}\\[5mm]
\scriptsize \phantom{XXXXXXX}Phi3.5-MoE, CA &
\scriptsize \phantom{XXXXXXX}Qwen2.5-32B, CA \\
\includegraphics[width=.45\textwidth]{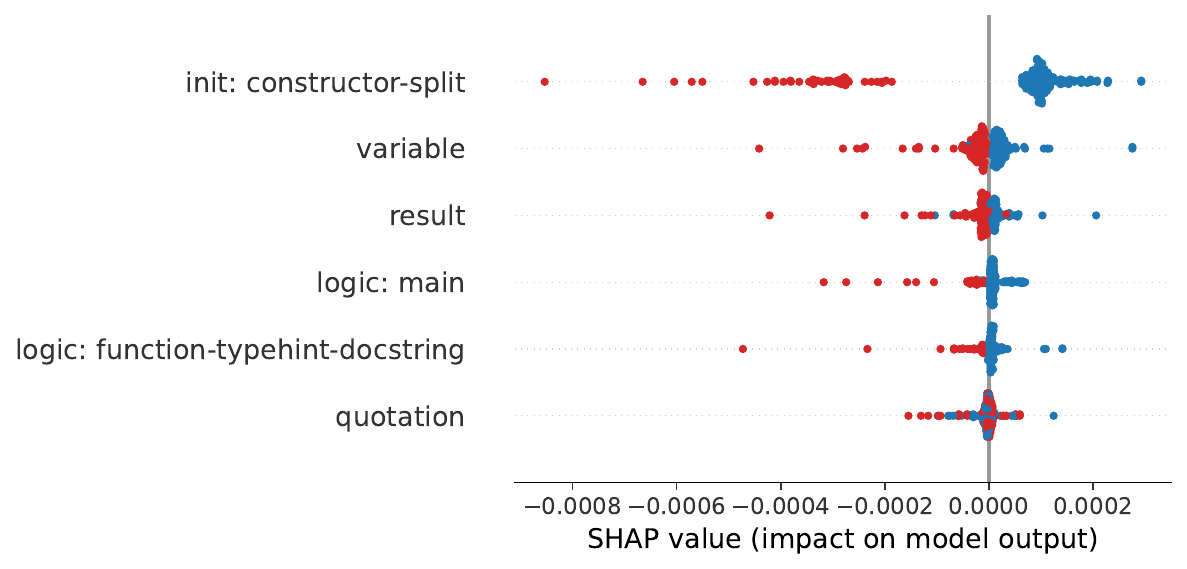} &
\includegraphics[width=.45\textwidth]{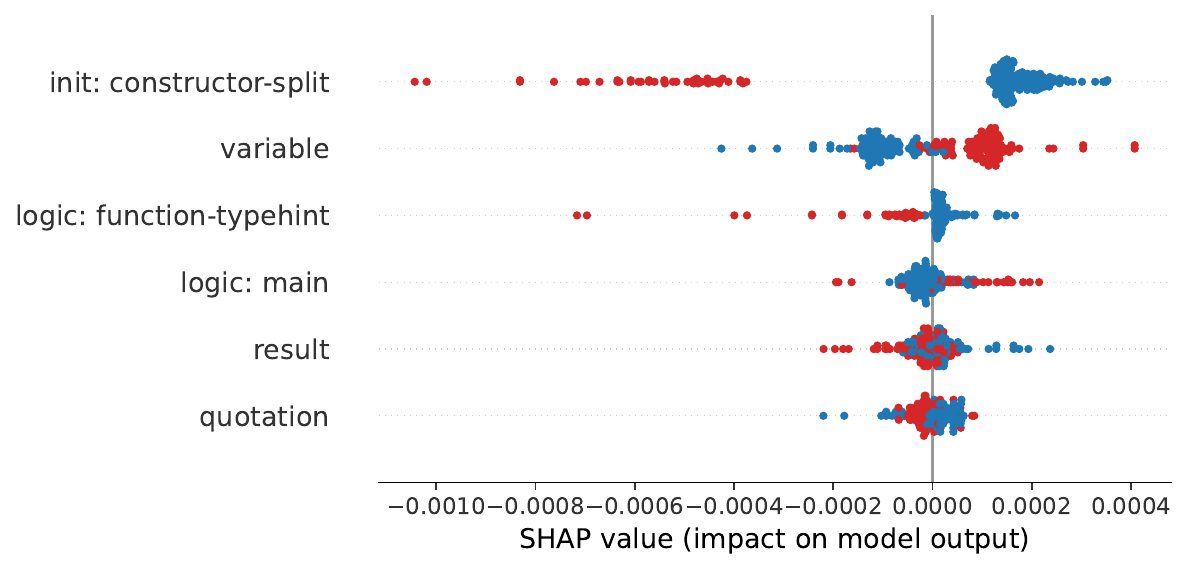}\\[5mm]
\scriptsize \phantom{XXXXXXX}Mistral-24B, CA &
\scriptsize \phantom{XXXXXXX}Llama-3.1-8B, CA \\
\includegraphics[width=.45\textwidth]{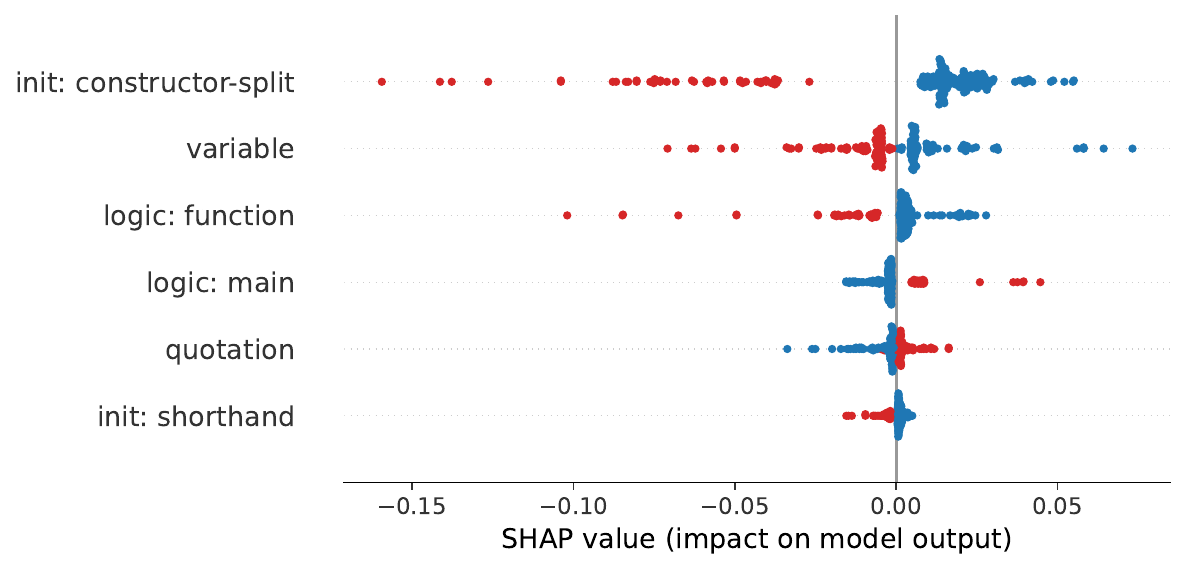} &
\includegraphics[width=.45\textwidth]{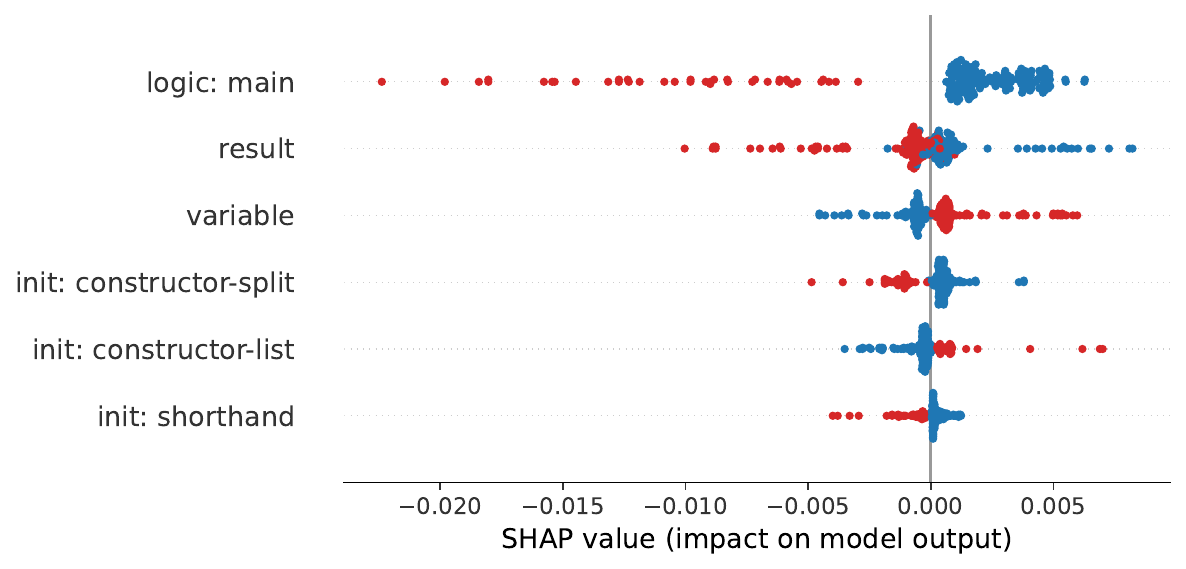}\\[5mm]
\end{tabular}
\begin{tabular}{c}
\scriptsize \phantom{XXXXXXX}Llama-3.2-3B, CA \\
\includegraphics[width=.45\textwidth]{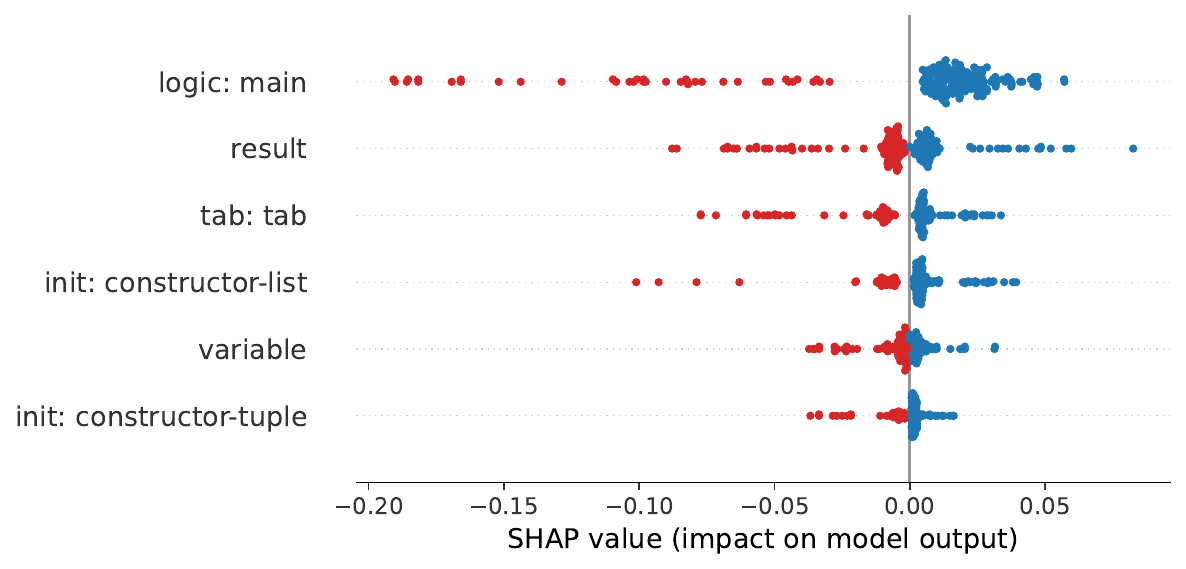}\\
\end{tabular}
\caption{Top 6 Shapley values among binary features of the \textbf{CA prompt category}.
Non-binary features are converted to one-hot representation.
A dot corresponds to a prompt template and the color red indicates that the feature is present.}
\label{fig:supp_shap_ca}
\end{figure*}

\Cref{fig:supp_shap_nl1,fig:supp_shap_nl2,fig:supp_shap_cs,fig:supp_shap_ca} include
the complete results of the \textbf{Shapley analysis}, giving further support to the claims made
in the main paper.

For example, in the case of the Python prompts, the initialization using the split
method is an important feature with a negative effect, except for Llama 3B, where it is not that important.

The feature of not using a helper function (logic: main) mostly results in a strong negative effect, except
some LLMs, like Qwen, where this feature has a slight positive effect.

The most important features are also quite variable. In general, the Python prompts show
very similar brittleness properties to those of the natural language prompts.


\end{document}